\documentclass[runningheads]{llncs}

\usepackage[year=2024]{eccv}

\usepackage{wrapfig}
\usepackage{adjustbox}

\usepackage{epsfig}
\usepackage{graphicx}
\usepackage{tikz}
\usepackage{pgfplots}
    \pgfplotsset{compat=1.18}
\usepackage{caption}
\usepgfplotslibrary{groupplots}
\usetikzlibrary{patterns}

\usepackage{bm}
\usepackage{bbding}

\newcommand\blfootnote[1]{%
  \begingroup
  \renewcommand\thefootnote{}\footnote{#1}%
  \addtocounter{footnote}{-1}%
  \endgroup
}

\def\modelname{\textsc{Meerkat}\xspace}
\def\ourdataset{AVFIT\xspace}   % to be changed
\def\ourdatasetsize{3M\xspace}

\def\taskunificationframework{\textsc{MeerkatBench}\xspace} % to be changed

\def\weakalignmentmodule{AVOpT\xspace} 
\def\strongalignmentmodule{AVACE\xspace}

\newcommand{\customsubsection}[1]{%
  \par
  \pagebreak[2]%
  \refstepcounter{subsection}%
    \everypar={%
      {\setbox0=\lastbox}% Remove the indentation
      \addcontentsline{toc}{subsection}{%
        {\protect\makebox[0.3in][r]{\thesubsubsection.} \hspace*{3pt}#1}}%
      \textbf{\thesubsection\space\space{#1}\space\newline}%
      \everypar={}%
    }%
  \ignorespaces
}

\usepackage{eccvabbrv}

\usepackage{graphicx}
\usepackage{booktabs}
\usepackage{pifont}
\usepackage{pgfplots}
\usepackage{algorithm}
\usepackage[noend]{algpseudocode}

\usepackage{arydshln}

\usepackage[accsupp]{axessibility}  % Improves PDF readability for those with disabilities.

\usepackage{multirow}
\usepackage{graphicx}
\usepackage{colortbl}
\usepackage{color, colortbl}

\usepackage[dvipsnames]{xcolor}

\definecolor{increase}{HTML}{e1e9fc}
\definecolor{decrease}{HTML}{e1e9fc}

\definecolor{task_spatial}{HTML}{FAD9D5}
\definecolor{task_temporal}{HTML}{C7E2F0}
\definecolor{task_mmfact}{HTML}{D0CEE2}
\definecolor{task_caption}{HTML}{E0F2CE}
\definecolor{task_avqa}{HTML}{FFF5B3}

\definecolor{f5e8d0}{HTML}{FAE8C8}
\colorlet{ThemeColor}{f5e8d0}

\definecolor{97C8DB}{HTML}{97C8DB}
\colorlet{TakeAwayColor1}{97C8DB}

\definecolor{AECEE8}{HTML}{AECEE8}
\colorlet{TakeAwayColor2}{AECEE8}

\definecolor{4B9CDE}{HTML}{4B9CDE}
\colorlet{TakeAwayColor3}{4B9CDE}

\definecolor{256bdb}{HTML}{256bdb}
\colorlet{TakeAwayColor4}{256bdb}

\definecolor{D1D8E3}{HTML}{D1D8E3}
\colorlet{TakeAwayColor5}{D1D8E3}

\definecolor{C8C8CC}{HTML}{C8C8CC}
\colorlet{TakeAwayColor6}{C8C8CC}

\definecolor{D4D3DB}{HTML}{D4D3DB}
\colorlet{TakeAwayColor7}{D4D3DB}

\definecolor{F5F5F5}{HTML}{F5F5F5}
\colorlet{TableAlternateColour}{F5F5F5}

\definecolor{44bd8c}{HTML}{44bd8c}
\colorlet{ConferenceColor}{44bd8c}

\definecolor{F08080}{HTML}{F08080}
\colorlet{LinePlot1}{F08080}

\definecolor{D2B48C}{HTML}{D2B48C}
\colorlet{LinePlot2}{D2B48C}

\definecolor{B8D65E}{HTML}{B8D65E}
\colorlet{LinePlot3}{B8D65E}

\definecolor{b5b3b3}{HTML}{b5b3b3}
\colorlet{LinePlot4}{b5b3b3}

\usepackage[pagebackref,breaklinks,colorlinks,citecolor=cyan]{hyperref}
\usepackage{orcidlink}
\usepackage{circledsteps}
\usepackage{pict2e}
\usepackage{wrapfig,lipsum,booktabs}

\usepackage{tcolorbox}
\newtcolorbox{abox}{colback=TakeAwayColor6,colframe=Black}

\definecolor{nicegreen}{rgb}{0.1, 0.6, 0.2}

\begin{document}

\title{~\includegraphics[height=20pt]{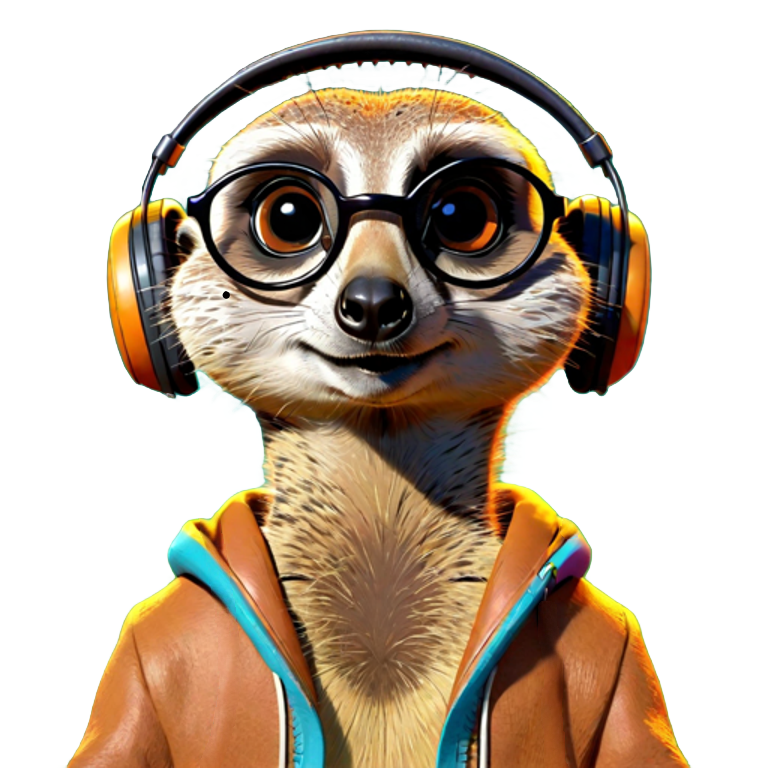}\modelname: Audio-Visual Large Language Model for Grounding in Space and Time}

% Fine-Grained Audio-Visual Multi-task Learning for Large Language Models 

\titlerunning{Meerkat}

\author{Sanjoy Chowdhury*\inst{1} \and
Sayan Nag*\inst{2} \and
Subhrajyoti Dasgupta*\inst{3} \and Jun Chen\inst{4} \and Mohamed Elhoseiny\inst{4}$^\dagger$ \and Ruohan Gao\inst{1}$^\dagger$ \and Dinesh Manocha\inst{1}$^\dagger$}

\authorrunning{Chowdhury et al.}

\institute{University of Maryland, College Park \and
University of Toronto \and Mila and Université de Montréal \and King Abdullah University of Science and Technology (KAUST) \\
Project page -- 
\url{https://github.com/schowdhury671/meerkat} \\
\email{\{sanjoyc,rhgao,dmanocha\}@umd.edu \quad sayan.nag@mail.utoronto.ca \quad subhrajyoti.dasgupta@umontreal.ca \quad \{jun.chen,mohamed.elhoseiny\}@kaust.edu.sa}}

\maketitle

% \vspace{-1cm}
\begin{figure}[h]
    \centering
    \vspace{-0.2in}
    \includegraphics[width=\columnwidth]{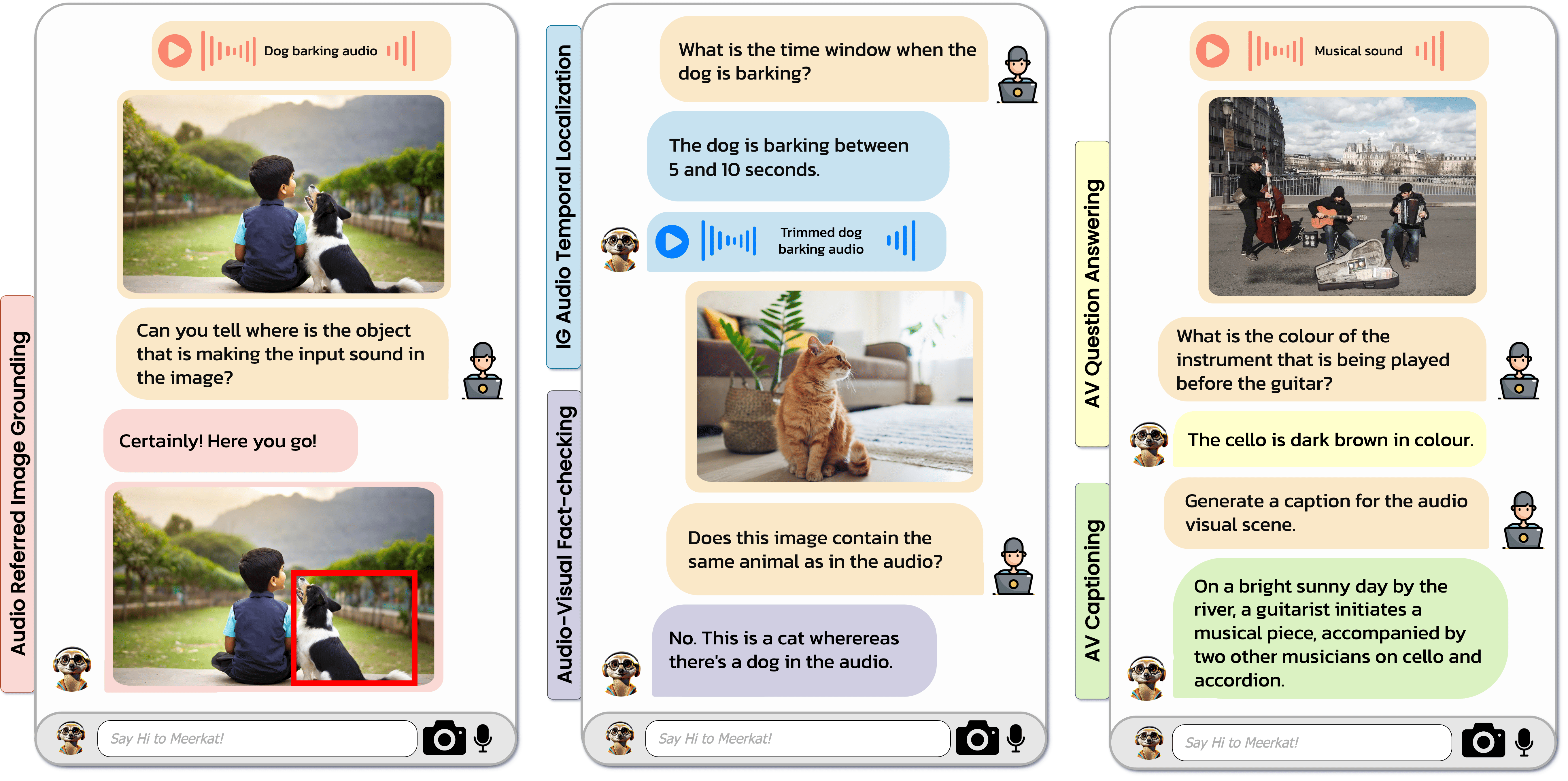}
    \vspace{-0.2in}
    \caption{We present \textbf{\modelname, an audio-visual LLM that can effectively ground both spatially and temporally in image and audio}.  
    %equipped with strong audio-visual understanding capabilities.
    Our model is adept in tasks that require fine-grained understanding such as  \colorbox{task_spatial}{\makebox(130,4){\textcolor{black}{Audio Referred Image Grounding}}}, \colorbox{task_temporal}{\makebox(192,4){Image Guided (IG) Audio Temporal Localization}} \& \colorbox{task_mmfact}{\makebox(128,4){Audio-Visual (AV) Fact-checking}}. It can also be extended to perform coarse-grained tasks like \colorbox{task_avqa}{\makebox(20,4){AVQA}} \& \colorbox{task_caption}{\makebox(55,4){AV Captioning}}.}
    \label{fig:teaser}
\end{figure}
\vspace{-12mm}

\blfootnote{$^*$Equal contribution.}
\blfootnote{$^\dagger$Equal advising.}

\begin{abstract}
    Leveraging Large Language Models’ remarkable proficiency in text-based tasks, recent works on Multi-modal LLMs (MLLMs) extend them to other modalities like vision and audio. However, the progress in these directions has been mostly focused on tasks that only require a coarse-grained understanding of the audio-visual semantics. We present \textit{\modelname}, an audio-visual LLM equipped with a fine-grained understanding of image and audio both spatially and temporally. With a new modality alignment module based on optimal transport and a cross-attention module that enforces audio-visual consistency, \textit{\modelname} can tackle challenging tasks such as audio referred image grounding, image guided audio temporal localization, and audio-visual fact-checking. Moreover, we carefully curate a large dataset \textit{\ourdataset} that comprises 3M instruction tuning samples collected from open-source datasets, and introduce \textit{\taskunificationframework}  that unifies \textit{five} challenging audio-visual tasks. We achieve state-of-the-art performance on all these downstream tasks with a relative improvement of up to \textit{37.12\%}.

\end{abstract}

\vspace{-1cm}
\section{Introduction}
\label{sec:intro}
Large Language Models (LLMs) \cite{gpt3, llama, vicuna, palm, raffel2020exploring} have demonstrated remarkable performance in various natural language processing tasks, achieving human-level accuracies in comprehension and reasoning abilities. Furthermore, powered by the emergent instruction fine-tuning paradigm \cite{instructgpt, chung2022scaling, peng2023instruction}, these language models can be equipped to follow open-ended natural language instructions, or even combined with other modalities, especially vision 
%To harness the potential of LLMs beyond language, recent studies 
\cite{flamingo, minigpt4, xllm, videollama, llava, otter, llamaadapter, pandagpt, videochatgpt, valley, macawllm, listenthinkunderstand}.
%combine these language models with other modalities (e.g., image and video). 
Audio, though often complementary to the associated visual scene, remains largely under-explored in the context of LLMs. 
%The audio modality often presents complementary information naturally associated with a visual scene. Despite its usefulness, the audio-visual modality in the context of MLLMs is relatively under-explored. 
%Effective utilization of this multi-modal information 
Building Multi-modal LLMs (MLLMs) that can \emph{listen} may enable new applications in multimedia content analysis, multi-modal virtual assistants, %healthcare, 
education and training, etc.     

Limited prior works (refer to Tab. \ref{tab:comparison_overview}) have incorporated audio in MLLMs \cite{listenthinkunderstand, xinstructblip, avllm}. However, they mostly focus on coarse-grained tasks such as captioning and question-answering,
%that only involve high-level understanding,
which is comparatively straightforward to be subsumed into an LLM interface \cite{pandagpt, macawllm, avllm, videollama}. 
%On the other hand, fine-grained tasks like detection, and segmentation necessitate alignment of spatial information and semantics. 
% \jun{I would not group these two tasks as coarse-grained tasks and fine-grained tasks. It is more like a multi-modal LLM, one w/o visual/audio grounding ability and one w/ grounding ability.}
Although there have been some recent advancements in leveraging MLLMs for grounding \cite{visionllm, bubogpt, cogvlm, ferret, minigptv2, shikra, vistallm}, they either only focus on the visual modality \cite{ferret, minigptv2, shikra, vistallm, lisa}, or struggles to capture fine-grained details occurring within audio-visual events due to insufficient joint modeling of the two modalities \cite{macawllm, pandagpt, videollama}. 

% \RGnote{Please double check the references I put here are correct.}
%in the context of MLLMs to
%perform visual-only grounding \cite{ferret, minigptv2, shikra, kosmos, unitab, vistallm}, 
%and audio understanding \cite{listenthinkunderstand, bubogpt, xinstructblip, avllm}, 
%the overarching goal of empowering LLMs to comprehend fine-grained AV semantics remains largely unexplored.

% \jun{my first impression for audio-visual spatio-temporal understanding is foucing the Video understanding} 

% There are a handful of works leveraging audio in LLMs \cite{listenthinkunderstand, bubogpt, xinstructblip, avllm} but their focus so far has been restricted to coarse-level understanding such as QA, captioning etc.

% \jun{there are no any prior works focusing on the audio-visual LLM?} 
% We attempt to address the above gap by finding answers to the following key questions: \textit{(i)} How to unify audio-visual spatio-temporal tasks in one framework, and will they benefit each other? \textit{(ii)} How to make spatio-temporal localization open-vocabulary, instruction following, general purpose? \textit{(iii)} Can such a fine-grained framework be extended to tackle coarse-grained tasks?

%Motivated by this research gap,
Our goal is to %
harness the power of LLMs for fine-grained audio-visual understanding. 
%build an audio-visual LLM with fine-grained understanding that can effectively ground both spatially and temporally in images and audios. 
This is challenging mainly because:
%However, there are considerable challenges to this including the 
\textit{(i)} there is a disparity of input and output formats across different tasks (e.g., image grounding from an audio query, image-guided audio temporal localization), 
% However, the problem becomes more challenging under a multi-modal setting due to an inherent disparity of input and output formats across different tasks. \jun{unclear about this inherent disparity of input and output formats, maybe give more concrate examples}. 
%These challenges propagate even further when the task demands learning fine-grained reasoning abilities across modalities. The challenges are two-fold: 
\textit{(ii)} no large-scale datasets exist for training audio-visual LLMs with grounding capabilities. 
%grounding datasets in the audio-visual domain which leads to an absence of audio-visual LLM-based foundational models with grounding capabilities, 
Existing audio-visual LLMs \cite{macawllm,pandagpt,avllm} are restricted to coarse-grained tasks and do not incorporate cross-modality fusion, which is a crucial component for achieving fine-grained understanding and reasoning capabilities, as shown in \cite{fiber,glip}.
 % It is inherently difficult to perform cross-modality fusion that can achieve fine-grained understanding and reasoning capabilities, as shown in \cite{fiber, glip}. 
%while there exist few models with audio-visual understanding abilities, they do not incorporate cross-modality fusion, a critical component in achieving fine-grained understanding and reasoning capabilities
%Consequently, no existing model can perform both image grounding and temporal localization simultaneously. 
Although there exist individual models capable of handling image grounding (BuboGPT \cite{bubogpt}) and temporal localization (TimeChat \cite{timechat}) separately, they are either not suitable for open-domain audio (TimeChat) or are not trained in an end-to-end fashion (BuboGPT) (refer to Tab. \ref{tab:comparison_overview}).

In light of these challenges, we present \modelname\footnote{Meerkats are known for their strong spotting and listening abilities.} (ref Fig. \ref{fig:teaser}), the first \textit{unified} audio-visual LLM framework that can effectively ground both spatially and temporally in image and audio, respectively. 
%an audio-visual LLM equipped with fine-grained understanding fuelled by 
%modality alignment and audio-visual consistency enforcement modules \weakalignmentmodule and \strongalignmentmodule respectively. 
It has two crucial modules that are key to its strong capability in fine-grained understanding: a modality alignment module 
%In the absence of grounding information (bounding boxes) in the annotations, the \weakalignmentmodule module 
that learns the cross-modal alignment between image and audio patches in a weakly-supervised manner based on optimal transport, and a cross-modal attention module that is 
%-inspired alignment strategy.
% \jun{still not clear how you do this. expand this a little bit to discuss the core idea}. 
%We empirically show this is superior to contrastive alignment approaches \cite{infonce, nce, clip} which are non-interpretable and operate on global embedding levels. 
% However, \weakalignmentmodule does not incorporate inter-modality interaction which can be enabled through a fusion strategy. Therefore, 
%o induce fine-grained supervision between audio-visual modality, we propose \strongalignmentmodule module,
capable of enforcing consistency in the cross-attention heatmaps. Together, these two modules enable learning better joint audio-visual representations that subsequently enhance downstream tasks.
%and subsequently improve the prediction performances in all the downstream tasks. 
%This is done by restricting the cross-modal attention within the bounding box annotation. Note that such enforcement is only performed in cases when the bounding box information is available during training. 
% We choose three downstream tasks that promote a fine-grained understanding of audio-visual information. 
% \jun{It is more like one is non-grounded alignment and the one is visual-grounding guided alignment. I would not say the second one as the strong supervision. strong is a bit suspicious. }
% \RGnote{I also commented out a lot in this para as many are too implementation details. Please double check.}

%%%% our contributions add
% In light of these challenges, we propose a novel model design choice by introducing weak and strong supervision modules \weakalignmentmodule and \strongalignmentmodule respectively. We empirically show having a weak alignment module boosts the performance of \strongalignmentmodule.

% \usepackage{multirow}
% \usepackage{graphicx}
\begin{table}[!t]
\centering
\renewcommand{\arraystretch}{1.0}
\resizebox{\columnwidth}{!}{%
\begin{tabular}{l|cc|c|c|c|ccc}
\toprule
\multicolumn{1}{c|}{} &
  \multicolumn{2}{c|}{\textbf{Audio Types}} &
   &
   &
   &
  \multicolumn{3}{c}{\textbf{Data Features}} \\
  % \cline{2-3} \cline{7-9} 
\multicolumn{1}{c|}{\multirow{-2}{*}{\textbf{Model}}} &
  \textbf{Speech} &
  \textbf{Open-domain} &
  \multirow{-2}{*}{\textbf{\begin{tabular}[c]{@{}c@{}}Output Image \\ Grounding\end{tabular}}} &
  \multirow{-2}{*}{\textbf{\begin{tabular}[c]{@{}c@{}}Output Audio \\ Grounding\end{tabular}}} &
  \multirow{-2}{*}{\textbf{End-to-end}} &
  \textbf{Convention} \hspace{2pt} &
  \textbf{GPT-Prompted} \hspace{2pt} &
  \textbf{Robustness} \\ \midrule
% LTU \cite{listenthinkunderstand} &
%   \textcolor{ForestGreen}{\ding{51}} &
%   \textcolor{ForestGreen}{\ding{51}} &
%   \textcolor{OrangeRed}{\ding{55}} &
%   \textcolor{ForestGreen}{\ding{51}} &
%   \textcolor{ForestGreen}{\ding{51}} &
%   \textcolor{ForestGreen}{\ding{51}} &
%   \textcolor{ForestGreen}{\ding{51}} &
%   \textcolor{OrangeRed}{\ding{55}} \\  
% VideoChat \cite{videochat} &
%   \textcolor{OrangeRed}{\ding{55}} &
%   \textcolor{OrangeRed}{\ding{55}} &
%   \textcolor{OrangeRed}{\ding{55}} &
%   \textcolor{OrangeRed}{\ding{55}} &
%   \textcolor{ForestGreen}{\ding{51}} &
%   \textcolor{ForestGreen}{\ding{51}} &
%   \textcolor{ForestGreen}{\ding{51}} &
%   \textcolor{OrangeRed}{\ding{55}} \\
% % \rowcolor{TableAlternateColour}
% VideoChatGPT \cite{videochatgpt} &
%   \textcolor{OrangeRed}{\ding{55}} &
%   \textcolor{OrangeRed}{\ding{55}} &
%   \textcolor{OrangeRed}{\ding{55}} &
%   \textcolor{OrangeRed}{\ding{55}} &
%   \textcolor{ForestGreen}{\ding{51}} &
%   \textcolor{ForestGreen}{\ding{51}} &
%   \textcolor{ForestGreen}{\ding{51}} &
%   \textcolor{OrangeRed}{\ding{55}} \\
% \rowcolor{TableAlternateColour}
VideoLlama \cite{videollama} &
  \textcolor{ForestGreen}{\ding{51}} &
  \textcolor{ForestGreen}{\ding{51}} &
  \textcolor{OrangeRed}{\ding{55}} &
  \textcolor{OrangeRed}{\ding{55}} &
  \textcolor{ForestGreen}{\ding{51}} &
  \textcolor{ForestGreen}{\ding{51}} &
  \textcolor{OrangeRed}{\ding{55}} &
  \textcolor{OrangeRed}{\ding{55}} \\
Macaw-LLM \cite{macawllm} &
  \textcolor{ForestGreen}{\ding{51}} &
  \textcolor{OrangeRed}{\ding{55}} &
  \textcolor{OrangeRed}{\ding{55}} &
  \textcolor{OrangeRed}{\ding{55}} &
  \textcolor{ForestGreen}{\ding{51}} &
  \textcolor{ForestGreen}{\ding{51}} &
  \textcolor{ForestGreen}{\ding{51}} &
  \textcolor{OrangeRed}{\ding{55}} \\
% \rowcolor{TableAlternateColour}
PandaGPT \cite{pandagpt} &
  \textcolor{ForestGreen}{\ding{51}} &
  \textcolor{ForestGreen}{\ding{51}} &
  \textcolor{OrangeRed}{\ding{55}} &
  \textcolor{OrangeRed}{\ding{55}} &
  \textcolor{ForestGreen}{\ding{51}} &
  \textcolor{ForestGreen}{\ding{51}} &
  \textcolor{OrangeRed}{\ding{55}} &
  \textcolor{OrangeRed}{\ding{55}} \\
AV LLM \cite{avllm} &
  \textcolor{ForestGreen}{\ding{51}} &
  \textcolor{ForestGreen}{\ding{51}} &
  \textcolor{OrangeRed}{\ding{55}} &
  \textcolor{OrangeRed}{\ding{55}} &
  \textcolor{ForestGreen}{\ding{51}} &
  \textcolor{ForestGreen}{\ding{51}} &
  \textcolor{ForestGreen}{\ding{51}} &
  \textcolor{OrangeRed}{\ding{55}} \\
% \rowcolor{TableAlternateColour}
X-InstructBLIP \cite{xinstructblip} &
  \textcolor{OrangeRed}{\ding{55}} &
  \textcolor{ForestGreen}{\ding{51}} &
  \textcolor{ForestGreen}{\ding{51}} &
  \textcolor{OrangeRed}{\ding{55}} &
  \textcolor{ForestGreen}{\ding{51}} &
  \textcolor{ForestGreen}{\ding{51}} &
  \textcolor{OrangeRed}{\ding{55}} &
  \textcolor{OrangeRed}{\ding{55}} \\
TimeChat \cite{timechat} &
  \textcolor{ForestGreen}{\ding{51}} &
  \textcolor{OrangeRed}{\ding{55}} &
  \textcolor{OrangeRed}{\ding{55}} &
  \textcolor{ForestGreen}{\ding{51}} &
  \textcolor{ForestGreen}{\ding{51}} &
  \textcolor{ForestGreen}{\ding{51}} &
  \textcolor{ForestGreen}{\ding{51}} &
  \textcolor{OrangeRed}{\ding{55}} \\
BuboGPT \cite{bubogpt} &
  \textcolor{OrangeRed}{\ding{55}} &
  \textcolor{ForestGreen}{\ding{51}} &
  \textcolor{ForestGreen}{\ding{51}} &
  \textcolor{OrangeRed}{\ding{55}} &
  \textcolor{OrangeRed}{\ding{55}} &
  \textcolor{ForestGreen}{\ding{51}} &
  \textcolor{OrangeRed}{\ding{55}} &
  \textcolor{OrangeRed}{\ding{55}} \\
\rowcolor{ThemeColor} 
\textbf{\modelname\ (ours)} &
  \textcolor{ForestGreen}{\ding{51}} &
  \textcolor{ForestGreen}{\ding{51}} &
  \textcolor{ForestGreen}{\ding{51}} &
  \textcolor{ForestGreen}{\ding{51}} &
  \cellcolor[HTML]{FAE8C8}\textcolor{ForestGreen}{\ding{51}} &
  \textcolor{ForestGreen}{\ding{51}} &
  \textcolor{ForestGreen}{\ding{51}} &
  \textcolor{ForestGreen}{\ding{51}} \\ \bottomrule
\end{tabular}%
}
\vspace{0.05in}
\caption{\textbf{Comparison of \modelname with recent Audio-Visual LLMs}. `Convention' refers to a collection of publicly available data that has been transformed using templates, `GPT-Prompted' signifies if the generated instructions are obtained/refined employing GPT, and `Robustness' is the model's ability to tackle negative samples. We compare our method against these approaches in Sec. \ref{sec:experiments_and_results}.}
\label{tab:comparison_overview}
\vspace{-10mm}
\end{table}

To support \modelname, we further introduce \taskunificationframework that unifies \textit{five} different audio-visual tasks (shown in Tab. \ref{tab:dataset_table}), including audio referred image grounding, image-guided audio temporal localization, audio-visual fact checking, audio-visual question answering, and audio-visual captioning (see Fig. \ref{fig:teaser} for examples). To enable the training of these five tasks, we also curate a large dataset AVFIT, which contains 3M instruction tuning samples with various degrees of difficulties for learning fine-grained audio-visual semantics. Extensive experiments on these tasks demonstrate the effectiveness of our proposed model.%, and we achieve state-of-the-art results on all of them, setting new baselines for future research in this direction.

In summary, we make the following main contributions:
\vspace{-0.1in}
\begin{itemize}
    \item We present \modelname, the first audio-visual LLM equipped with fine-grained spatio-temporal understanding that can ground in image and audio.
    \item We introduce \taskunificationframework that unifies five audio-visual learning tasks, and a new large instruction-tuning dataset \ourdataset to enable learning fine-grained audio-visual semantics.
    \item Evaluating on these five benchmark tasks, we set new state-of-the-art results on all of them with a relative improvement up to \textit{37.12\%}.
\end{itemize}

\vspace{-5mm}
\section{Related Works}
\label{sec:related_works}
% \noindent{\textbf{Audio-Visual Learning.}}  
\vspace{-1mm}
\noindent{\textbf{Multi-modal Large Language Models.}} 
Inspired by the success of instruction following capabilities of large language models \cite{instructgpt, vicuna, alpaca}, the community has recently started to leverage LLMs for understanding multi-modal contents. Powered by high-quality multi-modal instructional data, recent methods \cite{minigpt4, llava, otter, pandagpt, xllm, kosmos, shikra, flamingo} extend LLMs for multi-modal learning. While some approaches such as MiniGPT4 \cite{minigpt4}, X-LLM \cite{xllm}, and Video-ChatGPT \cite{videochatgpt} perform latent alignment between the pre-trained LLM and other modalities via learned visual encoder. Other methods like Otter \cite{otter}, and LLaMA-Adapter \cite{llamaadapter} learn cross-attention layers into the LLM to infuse multi-modal information. Prior works in the realm of LLMs predominantly focus on either visual-only inputs \cite{otter, llava, minigpt4, mplugowl} or tackle coarse-grained tasks \cite{videochat, videochatgpt} leaving room for fine-grained audio-visual understanding. 
% Although, more recently PG-Video-LLaVA \cite{pgvideollava} integrates audio for comprehensive video understanding, it restricts itself to human voice impulses. 
Unlike prior approaches, in this work, we focus on equipping LLMs with strong audio-visual comprehension abilities.

% \usepackage{multirow}
% \usepackage{graphicx}
\begin{table}[t]
\centering
\resizebox{\columnwidth}{!}{%
\begin{tabular}{c|c|c|c|c|c|c|c|c}
\hline
\textbf{\begin{tabular}[c]{@{}c@{}}Task\\ Granularity\end{tabular}} &
  \textbf{Task Name} &
  \textbf{Dataset} &
  \textbf{Train} &
  \textbf{Test} &
  \textbf{\begin{tabular}[c]{@{}c@{}}Spatial\\ Bounding Box\end{tabular}} &
  \textbf{\begin{tabular}[c]{@{}c@{}}Time \\ Interval\end{tabular}} &
  \textbf{\begin{tabular}[c]{@{}c@{}}\# Samples\\ Train / Test\end{tabular}} &
  \textbf{Metrics} \\ \hline
 &
  \cellcolor[HTML]{FAD9D5}{\color[HTML]{212529} } &
  Openimages-AudioSet &
  \textcolor{ForestGreen}{\ding{51}} &
  \textcolor{OrangeRed}{\ding{55}} &
  \textcolor{ForestGreen}{\ding{51}} &
  \textcolor{OrangeRed}{\ding{55}} &
  1.07M / -- &
  -- \\
 &
  \cellcolor[HTML]{FAD9D5}{\color[HTML]{212529} } &
  Openimages-VGGSound &
  \textcolor{ForestGreen}{\ding{51}} &
  \textcolor{OrangeRed}{\ding{55}} &
  \textcolor{ForestGreen}{\ding{51}} &
  \textcolor{OrangeRed}{\ding{55}} &
  180K / -- &
  -- \\
  &
  \cellcolor[HTML]{FAD9D5}{\color[HTML]{212529} } &
  AVSBench$^\dagger$ &
  \textcolor{ForestGreen}{\ding{51}} &
  \textcolor{ForestGreen}{\ding{51}} &
  \textcolor{ForestGreen}{\ding{51}} &
  \textcolor{OrangeRed}{\ding{55}} &
  2.30K / 0.49K &
  cIOU, AUC \\
 &
  \cellcolor[HTML]{FAD9D5}{\color[HTML]{212529} } &
  VGGSS &
  \textcolor{OrangeRed}{\ding{55}} &
  \textcolor{ForestGreen}{\ding{51}} &
  \textcolor{ForestGreen}{\ding{51}} &
  \textcolor{OrangeRed}{\ding{55}} &
  -- / 4.38K &
  cIOU, AUC \\
 &
  \cellcolor[HTML]{FAD9D5}{\color[HTML]{212529} } &
  PASCAL Sound &
  \textcolor{OrangeRed}{\ding{55}} &
  \textcolor{ForestGreen}{\ding{51}} &
  \textcolor{ForestGreen}{\ding{51}} &
  \textcolor{OrangeRed}{\ding{55}} &
  -- / 0.56K &
  cIOU, AUC \\
 &
  \multirow{-6}{*}{\cellcolor[HTML]{FAD9D5}{\color[HTML]{212529} \begin{tabular}[c]{@{}c@{}}Audio Referred\\ Image Grounding\end{tabular}}} &
  Flickr-Soundnet &
  \textcolor{OrangeRed}{\ding{55}} &
  \textcolor{ForestGreen}{\ding{51}} &
  \textcolor{ForestGreen}{\ding{51}} &
  \textcolor{OrangeRed}{\ding{55}} &
  -- / 2.78K &
  cIOU, AUC \\ \cline{2-9} 
 &
  \cellcolor[HTML]{C7E2F0} &
  Openimages-AudioSet Strong &
  \textcolor{ForestGreen}{\ding{51}} &
  \textcolor{ForestGreen}{\ding{51}} &
  \textcolor{OrangeRed}{\ding{55}} &
  \textcolor{ForestGreen}{\ding{51}} &
  96.5K / 24.1K &
  F1-score \\
 &
  \multirow{-2}{*}{\cellcolor[HTML]{C7E2F0}\begin{tabular}[c]{@{}c@{}}Image Guided Audio\\ Temporal Localization\end{tabular}} &
  LLP &
  \textcolor{OrangeRed}{\ding{55}} &
  \textcolor{ForestGreen}{\ding{51}} &
  \textcolor{OrangeRed}{\ding{55}} &
  \textcolor{ForestGreen}{\ding{51}} &
  -- / 2.32K &
  F1-score \\ \cline{2-9} 
\multirow{-9}{*}{Fine} &
  \cellcolor[HTML]{D0CEE2}Audio-Visual Fact-checking &
  Openimages-AudioSet &
  \textcolor{ForestGreen}{\ding{51}} &
  \textcolor{ForestGreen}{\ding{51}} &
  \textcolor{OrangeRed}{\ding{55}} &
  \textcolor{OrangeRed}{\ding{55}} &
  1.18M / 321K &
  F1-score \\ \hline
 &
  \cellcolor[HTML]{E0F2CE} &
  AVQA &
  \textcolor{ForestGreen}{\ding{51}} &
  \textcolor{ForestGreen}{\ding{51}} &
  \textcolor{OrangeRed}{\ding{55}} &
  \textcolor{OrangeRed}{\ding{55}} &
  40.4K / 16.9K &
  Accuracy \\
 &
  \multirow{-2}{*}{\cellcolor[HTML]{E0F2CE}\begin{tabular}[c]{@{}c@{}}AV Question\\ Answering\end{tabular}} &
  Music AVQA &
  \textcolor{ForestGreen}{\ding{51}} &
  \textcolor{ForestGreen}{\ding{51}} &
  \textcolor{OrangeRed}{\ding{55}} &
  \textcolor{OrangeRed}{\ding{55}} &
  25.7K / 7.36K &
  Accuracy \\ \cline{2-9} 
\multirow{-3}{*}{Coarse} &
  \cellcolor[HTML]{FFF5B3}AV Captioning &
  VALOR &
  \textcolor{ForestGreen}{\ding{51}} &
  \textcolor{ForestGreen}{\ding{51}} &
  \textcolor{OrangeRed}{\ding{55}} &
  \textcolor{OrangeRed}{\ding{55}} &
  25.0K / 3.50K &
  B@4, M, R, C \\ \hline
\end{tabular}%
}
\vspace{0.05in}
\caption{\textbf{Task-wise dataset distribution, dataset details, and metrics}. We collect \ourdataset, which is a collection of 12 datasets. We denote dataset-wise train/test usage. The visual grounding datasets contain spatial bounding box annotations while the audio temporal localization contains time-interval annotations. We consider audio-visual fact-checking as a fine-grained task as it requires an understanding of spatio-temporal grounding information (refer to Sec. \ref{sec:mmfact} for more details). Here B@4: BLUE@4, M: METEOR, R: ROUGE, C: CIDEr. For all our experiments we consider F1@0.5. $^\dagger$ We obtain the bounding box from the segmentation maps.}
\label{tab:dataset_table}
\vspace{-7mm}
\end{table}

\vspace{0.05in}

\noindent{\textbf{Fine-grained Multi-modal Understanding.}}  
Of late, general-purpose multi-modal large language models have demonstrated their effectiveness in unifying a versatile array of vision-language or video-understanding tasks. These models, powered by LLMs \cite{llama, llama2, flan, bloom, opt, galactica, palm} have superior reasoning and understanding capabilities. As a natural extension, MLLMs have been leveraged to unify region-based grounding tasks \cite{kosmos, shikra, minigptv2, ferret, cogvlm, bubogpt, lisa, gpt4roi, visionllm}. Despite significant strides, these models are still limited to fine-grained comprehension within a single modality. In this work, we propose \modelname to precisely address this research gap under in-the-wild audio-visual event settings. To this end, we present a novel audio-visual task unification framework which promotes strong multi-modal reasoning and understanding capabilities. 

\vspace{0.05in}

\noindent{\textbf{LLM guided Task Unification.}}
% talk about unified i/o, x-instructblip, anygpt, minigptv2,  llava, 
LLMs as an interface of task unification framework have seen massive advancements in recent times. Fuelled by the success of language models \cite{unitab, ofa, unifiedio}, the community has started to explore ways to unify generative and reasoning tasks under the sphere of language models leveraging its ease of accessibility. Various approaches \cite{ferret, minigptv2, xinstructblip, videochat} present alternative ways to integrate new tasks within the scope of LLMs. Inspired by the success of these approaches, we present, to the best of our knowledge, the first approach to unifying fine-grained audio-visual tasks. 

\vspace{0.05in}

\noindent{\textbf{Audio-Visual Learning.}} Benefiting from the natural synchrony between the visual and the auditory modalities, audio-visual learning has opened up abundant applications including audio-visual sound source localization \cite{mo2022closer, mo2023audio, sun2023learning, huang2023egocentric}, audio-visual sound separation \cite{chen2023iquery, tan2023language, majumder2022active}, audio-visual segmentation \cite{mao2023multimodal, avsbench, liu2024annotation}, audio-visual question answering \cite{avqadataset, yun2021pano, musicavqadataset}, audio-visual captioning \cite{vast, valor, tian2018attempt}. Different from these lines of work that focus on a single task, we aim to harness the power of LLM to propose a multi-task learning setting by unifying \textit{five} different audio-visual tasks with the LLM serving as a common interface.

% \RGnote{Should have a concise section reviewing Audio-Visual Learning, especially those relevant tasks studied in the paper. Something like: Exciting prior work on audio-visual learning have tackles a wide variety of tasks, including sound source location in video frames[cite], av-separation [cite], av-segmentation[cite], av captioning / question answering[cite], etc..... Different from the bove work that focuses on a single task, our work aims to harness the power of LLM to unify ....}

% \jun{For the figure: some suggestions for the figure. 
%   1) You can just mention CLIP and CLAP instead of using abbreviations. 
%   2) in the AVOpT module sub figure B, The image patches are after CLIP vision encoder, so they are the implicit features instead of the original pixel as you demonstrated. this also happens to the audio patches. You need to make this more clearly illustrated.
%   3) if $A^{c}$ is the just cross attention, you can directly write it as cross-attention since it is very well-known. 
%   4) You input the bbox mask directly though the AVACE module instead of from the CLIP? if this is the case, you also need to make it visible in the figure.
%   5) you don't have to mention the tokenizer. Tokenizer can be a part of LLaMA2 language model. this can simplify the diagram.}

\begin{figure}[t]
    \centering
    \includegraphics[width=\textwidth]{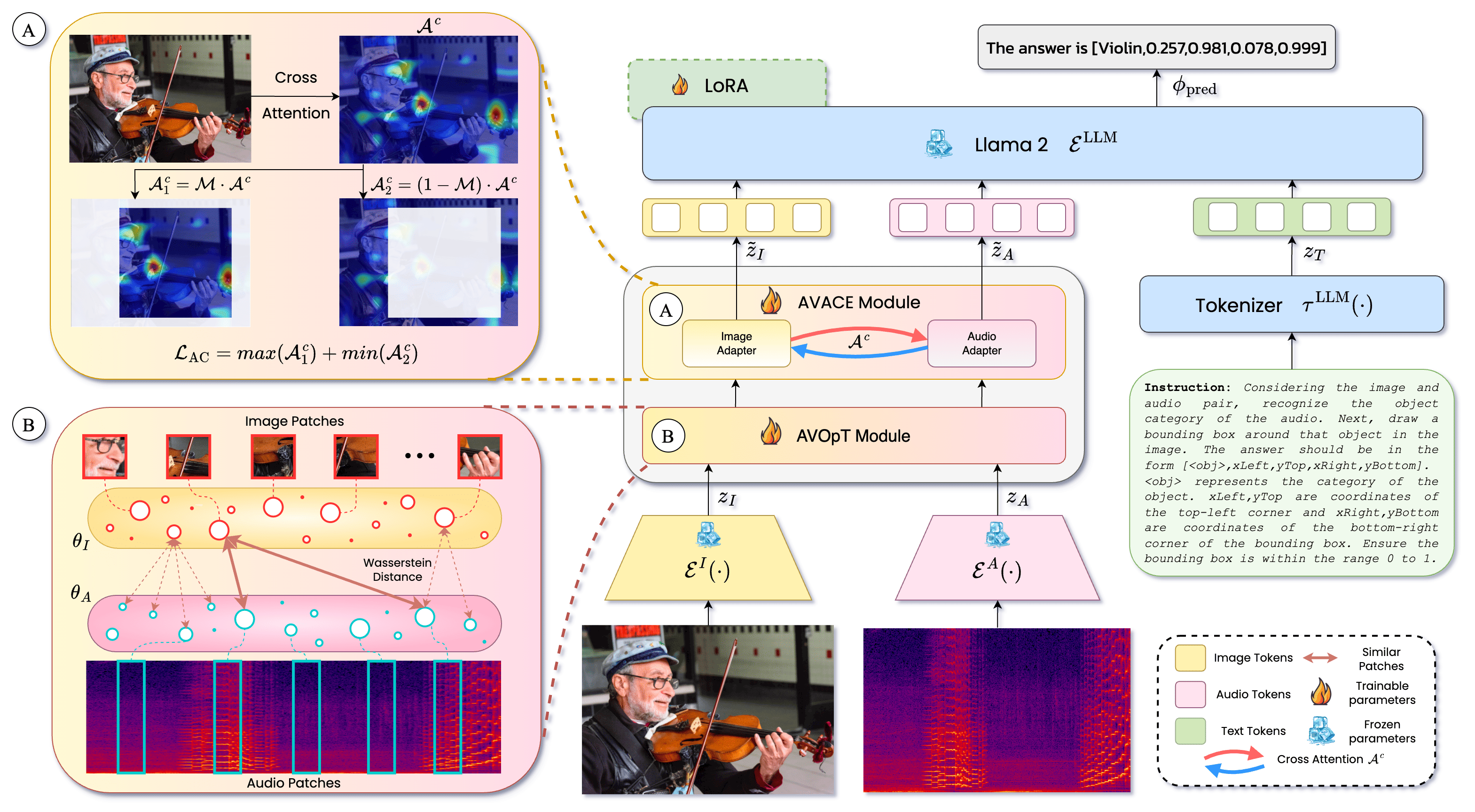}
    \vspace{-0.2in}
    \caption{\textbf{Overview of \modelname}. Our model is equipped with fine-grained audio-visual comprehension abilities. When fed with image \textit{I}, audio \textit{A} pairs, the Audio-Visual  Optimal Transport alignment (\weakalignmentmodule) module \Circled{B} learns the patch-wise image-audio association to facilitate weak alignment between the two modalities by minimizing the patch-level Wasserstein distance. Subsequently, the Audio-Visual Attention Consistency Enforcement (\strongalignmentmodule) module \Circled{A} maximizes the region-level alignment by confining the cross-modal attention maps around the objects of interest and minimizing the association with the background. After tokenizing the text instruction \textit{T}, the modality-specific latents ($\tilde{z}_{I}, \tilde{z}_{A}, z_{T}$) are passed to the instruction tuned Llama 2 model which serves as a unified interface for the downstream tasks. We employ a LoRA-based fine-tuning of the LLM.}
    \label{fig:network_diagram}
    \vspace{-5mm}
\end{figure}

% \vspace{-15mm}

% \RGnote{Provide an overview of this section / the whole model first before diving into details of each module. Also I am tr trying to think whether there might be a better structure for this section. It now reads a bit like a laundry list of different components. One potential way for you to consider: Sec. 3.1 Multimodal Feature Extraction (where you discuss the Image Encoder, Audio Encoder, and LLM); Sec. 3.2 Audio-Visual Feature Alignment (where you discuss the alignment module with Optimal Transport); Sec. 3.3 Audio-Visual Attention for Grounding in Space and Time (or a better name, where maybe you can discuss the attention module, as well as the numerical representations of box location and time segment; Sec. 4.4 Training Pipeline and Objective (where you present the complete Algorithm, and the training objective.}
\vspace{-4mm}

% \vspace{-2mm}
\section{Methodology}
\label{sec:methodology}
%\vspace{0.05in}

In this section, we introduce \modelname. Fig. \ref{fig:network_diagram} provides an overview of our approach. We first discuss the multi-modal feature extraction in Sec. \ref{multi-modal feature extraction}. In Sec. \ref{audio visual feature alignment modules} we introduce our novel audio-visual feature alignment modules. In Sec. \ref{overall training objective} we add the overall training objective followed by Sec. \ref{box and time representation} where we elaborate the numerical representations of the visual bounding box and time intervals. 

\vspace{0.05in}
\customsubsection{Multi-modal Feature Extraction}
\label{multi-modal feature extraction}
\vspace{0.05in}

% \subsubsection{Input: } 
% refer to Ferret sec 3.2

% add dimension and other details of each modality and how the overall input is fed into the pipeline

% \jun{You may put the discussion of Image encoder, audio encoder and LLM in the later part or even supplementary. You may prioritize discussing our model design direct to the point.}

\noindent{\textbf{Image Encoder.}} Given a batch of $k$ input images $\mathbf{I} = \{I_i\}^{k}_{i=1}: I_i \in \mathbb{R}^{H \times W \times C}$ where $H$, $W$, $C$ represent the height, width and channels respectively, we employ a pretrained CLIP-ViT-B/16 \cite{clip} encoder $\mathcal{E}^{I}(\cdot)$ to extract the image embeddings. Where $i^{\text{th}}$ image embedding can be represented as $z_I \in \mathbb{R}^{\mathcal{S}_I \times \mathcal{D}_I}$, where $\mathcal{S}_I$ and $\mathcal{D}_I$ denote the number of image tokens and hidden dimension respectively.
% \textcolor{red}{<add a concluding line>} 
% Notably, our image encoder exploits a significantly wider range of supervision by directly learning from unprocessed textual data related to images.

\noindent{\textbf{Audio Encoder.}} The audio encoder transforms the raw audio input into an audio embedding. We use the audio transformer backbone from CLAP \cite{clap} as our audio encoder due to its success in diverse audio tasks owing to its superior multi-modal alignment. We leverage this powerful pre-trained encoder ($\mathcal{E}^{A}(\cdot)$) to extract meaningful audio representations. For a batch of $k$ processed audio inputs $ \textbf{A} = \{A_i\}^{k}_{i=1}$: $A_{i} \in \mathbb{R}^{F \times T}$ where $F$ is the number of spectral components (e.g. Mel bins) and $T$ is the number of time bins. Each $i^{\text{th}}$ audio embedding is denoted as $z_{A} \in \mathbb{R}^{\mathcal{S}_A \times \mathcal{D}_A}$, $\mathcal{S}_A$ and $\mathcal{D}_A$ are the number of audio tokens and hidden dimension respectively. 
% \jun{is it because that this model CLAP model well aligning the audio with the language data?} 

% $z_{A}=\mathcal{E}^{A}\left(A_{i}\right)$

% \jun{the description for CLAP is not so clear}

\noindent{\textbf{LLM.}} \modelname adopts the open sourced Llama 2-Chat (7B) \cite{llama} as the large language model backbone. Pre-trained LLMs tokenizer projects the text sequence \textit{T} into embeddings $z_T \in \mathbb{R}^{\mathcal{S}_T \times \mathcal{D}_T}$, where $\mathcal{S}_T$ and $\mathcal{D}_T$ refer to token length and hidden dimension respectively. Before passing the image and audio embeddings into the LLM, they undergo transformations via additional linear layers to ensure the embedding dimensions across different modalities remain consistent. Since the LLM serve as the unified interface for audio-visual inputs, we rely on the language tokens to carry out the individual tasks.

% For the visual grounding tasks requiring the identification of spatial locations, we instruct the language model to generate textual descriptions of bounding boxes, indicating their spatial placements directly.  \jun{you may discuss the bounding box representation seperately. For this part, you can only discuss LLM itself}

\vspace{0.05in}
\customsubsection{Audio-Visual Feature Alignment}
\vspace{0.1in}

Inspired by the success of recent pre-training frameworks in grounding tasks \cite{fiber, minigptv2, glip}, we equip our model with two different levels of supervision: weak supervision through modality alignment module (\weakalignmentmodule) and strong supervision through audio-visual consistency enforcement module (\strongalignmentmodule). We follow a single-stage training strategy and empirically show our method achieves similar performance compared to two-stage training (more details in the appendix).   

\vspace{0.05in}
\noindent\textbf{Audio-Visual Optimal Transport Alignment Module (\weakalignmentmodule).}\label{audio visual feature alignment modules}
% The pretrained image and audio encoders generate respective feature representations which need to be aligned. Although typical alignment methods using contrastive losses \cite{infonce,nce} provide reasonable semantic supervision, they are assessed on the global feature level, and therefore imitate non-interpretable cross-modal alignment. On the contrary, bounding boxes and/or segmentation masks may be used for superior fine-grained cross-modal alignment \cite{glip,unitab,ofa}. However, such box-level or mask-level annotations are often not available. Therefore, we resort to an alternative strategy of aligning image and audio embeddings at a local-feature level in a weakly-supervised fashion. 
% % To the best of our knowledge, this is the first of its kind application of such a method in the audio-visual domain.
% In the recent times Earth Mover Distance (EMD) based algorithms \cite{} involving Optimal Transport (OT) methods \cite{} have been recently utilized for patch-alignment between the query and the support images in a siamese network \cite{deepemd}. Furthermore, in the vision-language context, OT based algorithms have been employed for patch-word alignment \cite{uniter}.
% Although fine-grained supervision can often facilitate learning strong associations across different streams of inputs by learning modality-agnostic features, obtaining large-scale annotated data is challenging. 
Weak supervision as a precursor to fine-grained supervision has been proven to be an effective training strategy in various tasks \cite{fiber, albef}. Earth Mover Distance based algorithms \cite{deepemd} involving Optimal Transport (OT) methods \cite{graphoptimaltransport} have been recently leveraged for patch-level alignment between the query and the support images in a siamese network \cite{deepemd}. Furthermore, in the context of vision-language models, OT-based algorithms have been employed for patch-word alignment \cite{uniter}. As the image (CLIP) and audio (CLAP) encoders are trained separately their learned embeddings are in a different semantic space. Our intuition is that such a patch-level alignment can improve vision and audio semantic consistency\cite{georgescu2023audiovisual}. We experimentally demonstrate that this patch-level weak guidance is superior to contrastive loss-based \cite{nce, infonce} global supervision (more details in appendix).
   
From a given image $I$ and audio $A$ pair, we obtain patch-level (local) feature embeddings $z_I$ and $z_A$ where, $z_I = \mathcal{E}^{I}(I); z_A = \mathcal{E}^{A}(A)$. For modeling cross-modal relations by utilizing the inherent rich semantic structures in these feature representations, we generate two discrete distributions, represented by $\theta_{I} \in \mathbf{P}(\mathbb{Z}_{I})$ and $\theta_{A} \in \mathbf{P}(\mathbb{Z}_{A})$, for image and audio respectively:
\begin{equation}
\vspace{-0.05in}
    \theta_{I}=\sum_{k=1}^{M}u_{I}(k)\delta_{z_I}(k); \theta_{A}=\sum_{l=1}^{N}u_{A}(l)\delta_{z_A}(l)
\vspace{-0.05in}
\end{equation}
where, $\sum_{k=1}^{M}u_{I}(k)=\sum_{l=1}^{N}u_{A}(l)=1$, $u_{I}$ and $u_{A}$ being the respective weight vectors for the probability distributions $\theta_{I}$ and $\theta_{A}$. $\delta_{z}$ is the Dirac delta function placed at support point $z$ in the embedding space \cite{plot}. The goal is to discern the \textit{optimal} transport plan while matching these two distributions. Therefore, we compute the Wasserstein Distance (WD) between these probability distributions $\theta_I$ and $\theta_A$ while preserving the topological information during the cross-domain alignment process, mathematically given as follows:
\begin{equation}
\vspace{-0.05in}
	\mathcal{L}_{\text{OT}} = \mathcal{D}_\mathrm{Wasserstein}(\theta_I,\theta_A) = \min_{\mathbf{\Omega}\in \mathrm{\Psi}(u_I,u_A)}\sum_{k} \sum_{l} \mathbf{\Omega}_{kl} \cdot \phi(z_I(k),z_A(l))
 \vspace{-0.05in}
\end{equation}
Here, $\mathrm{\Psi}(u_I,u_A) = \{ \mathbf{\Omega} \in \mathbb{R}_+^{M\times N} | \mathbf{\Omega}\mathbf{1}_N=u_I, \mathbf{\Omega}^\top\mathbf{1}_M=u_A \}$, $\phi(z_I(k),z_A(l))$ is the function computing the cosine distance between the cross-modal embedding pair, and $\mathbf{\Omega}$ is the transport plan, imitating the amount of \textit{mass} shifted from the distribution $\theta_I$ to the distribution $\theta_A$. An exact solution to the above expression leads to a sparse representation of the transport plan $\mathbf{\Omega}$ which at most $(2\cdot\text{max}(M,N) - 1)$ non-zero elements, ensuing an explainable and robust cross-modal alignment. We defer additional details to the appendix.

% \vspace{0.02in}
\noindent{\textbf{Audio-Visual Attention Consistency Enforcement Module (AVACE).}}
% \RGnote{In this subsection, we should briefly comment on its relation to prior work leveraging attention, cross-modal attention, and cite some representative work.}
% \jun{I think the cross attention is very common operation for the multi-modality alignment and you don't have to spend too many words on describing this.  The most important part is the region-level alignment, this region-level will encourage efficient localization. And it is better to discuss this motivation at the start. } \RGnote{Exactly. The motivation to very briefly discuss this and cite some work (maybe jsut one sentence) is to avoid overclaiming. We didn't invent attention, and I think it is good to acknowledge that. Something like: Using attention [CITE] in cross-modal learning has been widely explored in prior work [CITE]. In similar spirit but differently, here we use .....  }
Cross-modal interaction is essential for aligning the audio and visual modalities. Moreover, region-level supervision can encourage efficient localization. Inspired by the success of recent methods \cite{fiber, apollo, senocak2023sound}, we employ an adapter-based cross-attention strategy for efficient sound source localization.  
% plays a crucial role in designing a system equipped with superior spatial localization abilities.   
The modality-specific features in \weakalignmentmodule lack awareness \cite{huang2023sdif} of information from alternative modalities which can be infused through cross-modal attention.   
% does not involve any multi-modal interaction \jun{why you mention that the weakly supervised alignment phase does not involve any multi-modal interaction? minimizing the Wasserstein distance is also a way to build the interaction between these two interaction, is that true? }, and therefore not suitable for complex cross-modal feature representations. 
Therefore, to enable the audio-visual cross-modal reciprocity, we propose the \strongalignmentmodule module.

% The image ($z_I$) and audio ($z_A$) feature representations are respectively passed through the AVACE module to generate feature embeddings which are informed of each other's modality. This cross-domain synergy is established by the insertion of inter-modal attention fusion into the modality specific adapter layers.
% \begin{equation}
%     \tilde{z}_I = \text{SA}(z_I) + \text{CA}(Z_I,z_A); \tilde{z}_A = \text{SA}(z_A) + \text{CA}(z_A,z_I)
% \end{equation}
% where SA represents Self-Attention, CA represents Cross-Attention, $\tilde{z}_I$ and $\tilde{z}_A$ are the audio-informed image feature and image-informed audio feature respectively. 

Although in a multi-modal context, feature fusion through a cross-attention scheme is effective in attending to relevant objects in the image, inconsistencies may arise such as attended regions being dispersed throughout the image including background objects. The reasons can be attributed to the quality of interplay between the feature embeddings. Considering CLAP audio encoder pre-trained with examples such as \textit{`a man playing the violin'} (refer Fig. \ref{fig:network_diagram}) paired with audio of a violin, the cross-modal knowledge of audio representations encourages it to focus on both the man and the violin in the image. Therefore, to ensure superior region-level alignment we confine the cross-modality attention map ($\mathcal{A}^c$) within the boundaries of the object of interest, denoted by the ground-truth bounding box. Considering a bounding box represented as $[x_{\text{Left}}, y_{\text{Top}}, x_{\text{Right}}, y_{\text{Bottom}}]$, we define a mask $\mathcal{M}$ such that $\mathcal{M}(y_{\text{Top}}:y_{\text{Bottom}}, x_{\text{Left}}:x_{\text{Right}}) = 1 \text{, otherwise 0}$. Our goal is to maximize the attention within this bounding box and minimize it elsewhere. Therefore, we mathematically formulate the attention consistency objective  $\mathcal{L}_{\text{AC}}$ as follows:
\begin{equation}
\vspace{-0.05in}
    \mathcal{L}_{\text{AC}} = \lambda_1\left(1 - \frac{\sum_{i,j}{\mathcal{M}(i,j) \mathcal{A}^c(i,j)}}{\sum_{i,j}{\mathcal{M}(i,j)} + \epsilon_{1}}\right) + \lambda_2\left(\frac{\sum_{i,j}{\left(1 - \mathcal{M}(i,j)\right) \mathcal{A}^c(i,j)}}{\sum_{i,j}{\left(1 - \mathcal{M}(i,j)\right)} + \epsilon_{2}}\right)
% \vspace{-0.02in}
\end{equation}
Here, $\mathcal{A}^c$ denotes the audio-visual cross-modality attention, $(i,j)$ represents the pixel location, $\lambda_1$, $\lambda_2$ are the loss hyper-parameters (we keep $\lambda_1=\lambda_2=0.5$), and $\epsilon_1$, $\epsilon_2$ are the stability factors respectively. In Sec. \ref{sec:ablation}, we demonstrate that $\mathcal{L}_{\text{AC}}$ encourages efficient localization and audio-visual alignment of the cross-attention maps, eventually leading to improved fine-grained cross-modal representations for downstream tasks.

\vspace{0.02in}
\customsubsection{Overall training objective}
\label{overall training objective}
\vspace{0.05in}

Our overall training objective comprises a combination of three sub-objectives: cross-entropy loss ($\mathcal{L}_{\text{CE}}$), weak AV alignment loss ($\mathcal{L}_{\text{OT}}$), and attention consistency loss ($\mathcal{L}_{\text{AC}}$). These losses are added together to obtain the final training loss for \modelname\ given as:
\begin{equation}\label{eqn:total_loss}
    % \vspace{-5mm}
    \mathcal{L}_{\modelname} = \mathcal{L}_{\text{CE}} + \lambda_{\text{OT}} \cdot \mathcal{L}_{\text{OT}} + \lambda_{\text{AC}} \cdot \mathcal{L}_{\text{AC}}
    % \vspace{-3mm}
\end{equation}
Here, $\lambda_{\text{OT}}$ and $\lambda_{\text{AC}}$ are the loss weighting factors. We provide Algorithm \ref{algo:training} outlining the overall training procedure.

%%%% TAKEWAY  %%%%%%%%

% % \vspace{-1mm}
% \begin{quote}
%     \makebox[\linewidth]{%
%         \colorbox{ThemeColor}{%
%             \hspace*{0mm} % Adjust left alignment
%             \begin{minipage}{\dimexpr\linewidth+10\fboxsep\relax} % Adjust width
%                 % \fontsize{9pt}{10pt}\selectfont % Font settings
%                 \textbf{Takeaway:} Put here.
%             \end{minipage}%
%         }%
%     }
% \end{quote}
% % \vspace{-5mm}

\begin{algorithm}[!t]
\small
\caption{\modelname: Training}
\label{algo:training}
% \begin{abox}
\begin{algorithmic}[1]
\Require{Image: $I$; Audio: $A$; Textual Instruction: $T$; 
Pre-trained LLM: $\mathcal{E}^{\text{LLM}}(\cdot)$; LLM Tokenizer: $\tau^{\text{LLM}}(\cdot)$; Pre-trained Image Encoder: $\mathcal{E}^I(\cdot)$; Pre-trained Audio Encoder: $\mathcal{E}^A(\cdot)$; AVACE Module: $\text{AVACE}(\cdot,\cdot)$; Masks from GT Bounding-Boxes: $\mathcal{M}$; Loss Hyperparameters: $\lambda_{\text{OT}}, \lambda_{\text{AC}}$; GT Tokens: $\phi_{\text{GT}}$.}
\Ensure{Fine-tuned LLM: $\mathcal{E}^T(\cdot)$; Trained AVACE Module: $\text{AVACE}(\cdot,\cdot)$; Predicted Tokens: $\phi_{\text{pred}}$.}
\State $z_{I} \leftarrow \mathcal{E}^I(I); z_{A} \leftarrow \mathcal{E}^A(A)$ \Comment{\textit{Obtain Visual and Audio Embeddings.}}
\State $z_{T} \leftarrow \tau^{\text{LLM}}(T)$ \Comment{\textit{Tokenize and Obtain Textual Encodings.}}
\State $\tilde{z}_{I},\tilde{z}_{A},\mathcal{A}^c \leftarrow \text{AVACE}(z_I,z_A)$ \Comment{\textit{Obtain Audio-Visual Projections, Cross-Attn Map.}}
\State $z_{AVT} \leftarrow (\tilde{z}_I \parallel \tilde{z}_A \parallel z_T)$ \Comment{\textit{Concatenate Embeddings.}}
\State $\phi_{\text{pred}} \leftarrow \mathcal{E}^{\text{LLM}}(z_{AVT})$ \Comment{\textit{LLM Output.}}
\State $\mathcal{L}_{\modelname} \leftarrow \mathcal{L}_{\text{CE}}(\phi_{\text{pred}},\phi_{\text{GT}}) + \lambda_{\text{OT}} \cdot \mathcal{L}_{\text{OT}}(z_I, z_A) + \lambda_{\text{AC}} \cdot \mathcal{L}_{\text{AC}}(\mathcal{A}^c, \mathcal{M})$
\State Optimize model parameters to reduce $\mathcal{L}_{\modelname}$ until convergence.
\State \Return $\mathcal{E}^T(\cdot)$, $\text{\strongalignmentmodule}(\cdot,\cdot)$, $\phi_{\text{pred}}$.
\end{algorithmic}
% \end{abox}
\end{algorithm}

\vspace{0.05in}
\customsubsection{Numerical Representation of Box Location and Time Segment}\label{box and time representation}
\vspace{0.05in}

\noindent \textbf{Representation of Box Location.} We embed the location of bounding boxes with numerical values in the natural language sequence. A box is represented intuitively by its top-left and bottom-right corners, i.e., [$x_{\text{Left}}$, $y_{\text{Top}}$, $x_{\text{Right}}$, $y_{\text{Bottom}}$]. Notably, these values are normalized whose factors are determined by the size of the respective image to which the bbox belongs. These coordinates may appear in either the input or the output sequences depending on the task. For instance, in \textit{Audio Referred Image Grounding} task, \modelname\ predicts the bounding box of the object of interest, whereas, for \textit{Audio-Visual Fact-checking} task, the text input to \modelname\ might contain the box coordinates.

\vspace{1mm}
\noindent \textbf{Representation of Time Segment.} We embed the time interval information using numerical figures in the natural language expression. A time segment is intuitively represented by its start and end times, i.e., [\texttt{tStart}, \texttt{tEnd}], designating the onset of an event or an activity. Similar to boxes, these representations may appear in either the input or the output sequences depending on the task. For instance, in \textit{Image Guided Audio Temporal Localization} task, the model predicts the time interval within which the query might have occurred, while for \textit{Audio-Visual Fact-checking}, the input sequence might contain a reference time window. We add more details on the instruction preparation formats in the appendix.

% \jun{How you make the model to be time-aware? Maybe also briefly discuss this}

% \RGnote{Till now, we haven't talked about tasks. I saw it's currently discussed in the experiments section, which might be too late. How about making Sec. 4: MeerkatBench: A Unified Benchmark Suite for Fine-grained Audio-Visual Understanding. Then Sec 4.1 talks about the dataset (used for training). Sec. 4.1 provides an overview of the five tasks (it's also important to reiterate the value of these tasks, and emphasize the benefit of unifying them here). Then Sec. 5 talks about the actual results for the five tasks, baselines, etc.}

\vspace{-4mm}
% \newpage
\section{\taskunificationframework: A Unified Benchmark Suite for Fine-grained Audio-Visual Understanding} \label{task_specifications}

\vspace{-2mm}
\customsubsection{Task Overview} 
\vspace{0.05in}

Multi-modal conversation as an emergent ability is gaining prominence in the context of MLLMs. Although a line of research \cite{ferret, vistallm, minigptv2} addresses vision-language tasks, extension to other modalities such as audio is relatively underexplored. The task's difficulty escalates further when an intricate understanding of the modality-specific information is necessitated. To add to this, there doesn't exist any publicly available dataset that particularly facilitates such tasks. One of our primary contributions is to introduce a novel audio-visual fine-grained task unification benchmark. To this end, we present \taskunificationframework comprising three fine-grained tasks: \textit{(i)} audio referred image grounding, \textit{(ii)} image guided audio temporal localization, \textit{(iii)} audio-visual fact-checking, and two coarse-grained tasks: \textit{(iv)} audio-visual question answering, \textit{(v)} audio-visual captioning. %While our method is primarily aimed at tackling fine-grained tasks, we observe that our approach can be easily extended to coarse-grained tasks. 
% by training on limited data. We finetune our model with only 25K instruction tuning samples especially designed for this.

\vspace{0.05in}
\customsubsection{\ourdataset-3M: \underline{A}udio \underline{V}isual \underline{F}inegrained \underline{I}nstruction \underline{T}uning Dataset} 
\label{sec:datasets} In this section, we present \ourdataset, an AV instruction tuning dataset comprising \ourdatasetsize multi-modal dialogues for model training. \ourdataset consists of samples collected in the following ways: \textit{(i)} suitable adaptation of public datasets and \textit{(ii)} instruction-tuning data generation via prompting GPT-3.5 \cite{gpt3}. Next, we discuss the data curation procedure: \label{data_curation}

\noindent{\textbf{Adaptation of Public Datasets.}} Depending on the task and availability of datasets, we either collect the image-audio pairs directly from the publicly available datasets (VGG-SS \cite{vggss}, AVSBench \cite{avsbench}, Flickr-SoundNet \cite{flickrsoundnetarda}, LLP \cite{llpdataset}, AVQA \cite{avqadataset}, MUSIC-AVQA \cite{musicavqadataset}, VALOR \cite{valor}) or follow a semi-automated strategy to prepare the pairs by forming matching image-audio pairs from large-scale datasets having visual grounding annotation such as Openimages \cite{openimages}, PASCAL \cite{pascal} and audio event datasets like AudioSet/AudioSet Strong \cite{audioset}, VGG-Sound \cite{vggsound}. We retain the original category labels (\textit{`Existential', `Temporal', etc.}) from the MUSIC-AVQA. To get similar insights in the AVQA dataset, we categorise every sample into one of the \textit{`Existential', `Temporal', `Localisation', `Count'} and \textit{`World Knowledge'} categories. During the direct collection of pairs, we augment the audio snippet with a carefully chosen representative frame from the associated video. On the other hand, while forming pairs ourselves, we refer to a lookup table which we prepare beforehand by matching the corresponding class labels from the image and the audio datasets (more details in the appendix). We associate each image sample with its counterpart from the audio dataset. Finally, we supplement the image-audio pairs with the generated instructions as explained next. Details on the task-wise dataset details can be found in Tab. \ref{tab:dataset_table}.  

% create a lookup to associate the corresponding categories between the visual and audio datasets to form image-audio pairs which
% we supplement further with instruction tuning as explained next. Details on the task-wise dataset details can be found in Fig. \ref{fig:dataset_division}\\

\noindent{\textbf{GPT-Assisted Instruction Generation.}} Instruction tuning datasets \cite{llava, scienceqa, flickr30k, unnaturalinstruction} have primarily focused on coarse-grained details like global image descriptions in the form of captioning or question answering without explicitly capturing fine-grained details. In this work, we aim to bridge this gap by introducing \ourdataset that promotes region-level and time-sensitive understanding in the following ways: \textit{(i)} \ourdataset includes spatial coordinates of objects of interest (bounding box) along with corresponding audio snippets which leverage the synergy between audio-visual data. \textit{(ii)} The designed dialogues audio time intervals either in input or output or both. \textit{(iii)} To generate high-quality instructions we manually write a few example descriptions of each task and resort to GPT-3.5 \cite{gpt3} to create different variations. For further refinement of the generated dialogues we re-prompt GPT-4 \cite{gpt4} to ensure quality by reducing its context size. During
training, we randomly pick one instruction for each sample. Fig. \ref{fig:network_diagram} illustrates a sample instruction from \taskunificationframework. We use special tokens \texttt{<image>}, \texttt{<audio>}, \texttt{<obj>} which we later replace with instruction-guided image, audio and object categories respectively to generate prefix-based prompting.

% Moreover, it has been demonstrated \cite{llava, minigpt4, m3it} that refining dialogue instruction data can boost multi-modal LLMs' capabilities in comprehending human instructions thereby producing realistic and accurate responses.   

\vspace{0.1in}
\section{Experiments and Results}\label{sec:experiments_and_results}

% We now validate our approach and compare to existing generalist and specialist baselines in the literature.

% \RGnote{Better to talk about Implementation Details as Sec. 5.1, then baselines, then metrics.}

\customsubsection{Baselines} To the best of our knowledge, \modelname\ is the first MLLM that unifies audio-visual spatial and temporal grounding, alongside possessing strong reasoning capabilities. We carefully choose the closest baseline for each task and suitably adapt them for fair comparisons. Owing to BuboGPT's \cite{bubogpt} spatial localization ability, we select it as our baseline for the audio referred image grounding task. Most similar in spirit to our image guided audio-temporal localization task is TimeChat \cite{timechat}. It leverages the pre-trained VideoLlama model and suitably instruction-tune it to tackle temporal grounding tasks. Due to their audio-visual comprehension abilities, we resort to X-InstructBLIP \cite{xinstructblip}, Macaw-LLM \cite{macawllm}, PandaGPT \cite{pandagpt}, and VideoLlama \cite{videollama} as baselines for audio-visual fact-checking, AV question answering, and AV captioning tasks respectively.
% are multimodal models capable of handling both audio and visual inputs, however lacking fine-grained understanding. 
% Finally, we also provide a baseline model which we call $\text{\modelname}_{\text{baseline}}$. Specifically, the difference between this baseline model and our final model is that we do not introduce the OT-based alignment and \strongalignmentmodule losses during the training phase (although the cross attention exists between the modality-specific adapter layers in the \strongalignmentmodule module). 
Please refer to Tab. \ref{tab:comparison_overview} for an overview of the characteristics of the generalist baselines. For specialist baselines, refer to the corresponding task tables. We finetune all baselines on our datasets except for using Openimages-AudioSet and Openimages-VGGSound train splits from the audio-referred visual grounding task. 

% \RGnote{And maybe it's better to name this subsection Baselines, and in the beginning briefly set the stage that we compare to both generalist and specialist baselines. This sub-section mainly discusses generalist baselines. But in the end, comment that for specialist baselines, please refer to the corresponding tables for the tasks.}

\vspace{0.05in}
\customsubsection{Main Results}
\vspace{0.05in}

\noindent{\textbf{Audio Referred Image Grounding (ARIG)}} 
% We categorize this task into the following two related yet slightly different categories: 
% \noindent{\textbf{Bounding Box Prediction.}} 
This task involves visual grounding by predicting the coordinates of a bounding box around the object of interest guided by the input audio. We prepare 1.2M image-audio-instruction pairs using steps explained in Sec. \ref{data_curation}. We add details of the input instruction format and model output in the appendix. \modelname achieves superior performance in sounding object localization task, setting a new benchmark as shown in Tab. \ref{tab:spatial_grounding_boundingbox}.
% \noindent{\textbf{Sounding Object Segmentation.}} We equip our model to perform segmentation by extending the object localization pipeline. For this task, we employ SAM \cite{sam} which when fed with the coordinates of the bounding box encode them via positional encoding to produce the segmentation map. Fig. \ref{} demonstrates strong spatial understanding and commonsense reasoning capability. Tab. \ref{tab:spatial_grounding_segmentation} 
\begin{table}[t]
    \centering
    \renewcommand{\arraystretch}{0.8}
    \resizebox{\columnwidth}{!}{%
    \begin{tabular}{lc|cc|cc|cc|cc}
\toprule
\multicolumn{1}{c}{} &
   &
  \multicolumn{2}{c|}{\textbf{VGG-SS}} &
  \multicolumn{2}{c|}{\textbf{Flickr-SoundNet}} &
  \multicolumn{2}{c|}{\textbf{PascalSound}} &
  \multicolumn{2}{c}{\textbf{AVSBench}} \\ 
\multicolumn{1}{l}{\multirow{-2}{*}{\textbf{Models}}} &
  \multirow{-2}{*}{\textbf{Generalist?}} &
  \textbf{cIoU $\uparrow$} &
  \textbf{AUC $\uparrow$} &
  \textbf{cIoU $\uparrow$} &
  \textbf{AUC $\uparrow$} &
  \textbf{cIoU $\uparrow$} &
  \textbf{AUC $\uparrow$} &
  \multicolumn{1}{c}{\textbf{cIoU $\uparrow$}} &
  \multicolumn{1}{c}{\textbf{AUC $\uparrow$}} \\ \midrule
SSPL \cite{song2022self}  &
  \textcolor{OrangeRed}{\ding{55}} &
  33.90 &
  38.00 &
  76.70 &
  60.50 &
  51.72 &
  39.79 &
  61.32 &
  48.44 \\
EZ-VSL \cite{mo2022localizing}  &
  \textcolor{OrangeRed}{\ding{55}} &
  38.85 &
  39.54 &
  83.94 &
  63.60 &
  51.90 &
  40.25 &
  60.06 &
  49.64 \\
SSL-TIE \cite{liu2022exploiting}  &
  \textcolor{OrangeRed}{\ding{55}} &
  38.63 &
  39.65 &
  79.50 &
  61.20 &
  52.14 &
  40.44 &
  62.88 &
  51.28 \\
SLAVC \cite{mo2022closer}  &
  \textcolor{OrangeRed}{\ding{55}} &
  39.80 &
  -- &
  86.00 &
  -- &
  52.29 &
  42.19 &
  63.39 &
  51.07 \\
MarginNCE \cite{park2023marginnce}  &
  \textcolor{OrangeRed}{\ding{55}} &
  39.78 &
  40.01 &
  85.14 &
  64.55 &
  53.61 &
  45.52 &
  65.85 &
  52.92 \\
HearTheFlow \cite{fedorishin2023hear}  &
  \textcolor{OrangeRed}{\ding{55}} &
  39.40 &
  40.00 &
  84.80 &
  64.00 &
  55.48 &
  47.40 &
  67.49 &
  54.39 \\
FNAC \cite{sun2023learning}  &
  \textcolor{OrangeRed}{\ding{55}} &
  41.85 &
  40.80 &
  85.14 &
  64.30 &
  57.38 &
  48.03 &
  68.78 &
  56.19 \\
Alignment \cite{senocak2023sound}  &
  \textcolor{OrangeRed}{\ding{55}} &
  42.64 &
  41.48 &
  82.40 &
  64.60 &
  58.34 &
  49.86 &
  71.57 &
  57.52 \\ 
% [0.5ex]\cdashline{1-10}\\[-1.0ex]
% \customdottedlinetablespatial
\midrule
BuboGPT \cite{bubogpt} &
  \textcolor{ForestGreen}{\ding{51}} &
  40.31 &
  39.68 &
  81.17 &
  62.29 &
  58.52 &
  51.63 &
  74.33 &
  59.49 \\ \midrule  
% $\text{Meerkat}_{\text{baseline}}$ &
%   \textcolor{ForestGreen}{\ding{51}} &
%   42.93 &
%   41.14&
%   84.26 &
%   64.29&
%   61.82 &
%   52.65 &
%   75.62 &
%   62.19 \\
\rowcolor{ThemeColor} 
% $\text{Meerkat}_{\text{final}}$  
\textbf{\modelname (ours)} &
  \textcolor{ForestGreen}{\ding{51}} &
  \cellcolor{ThemeColor}\textbf{48.51} &
  \cellcolor{ThemeColor}\textbf{45.62} &
  \cellcolor{ThemeColor}\textbf{88.35} &
  \cellcolor{ThemeColor}\textbf{67.88} &
  \cellcolor{ThemeColor}\textbf{65.23} &
  \cellcolor{ThemeColor}\textbf{56.10}&
  \cellcolor{ThemeColor}\textbf{79.82}&
  \cellcolor{ThemeColor}\textbf{65.35} \\ \midrule
$\textcolor{blue}{\Delta_{\text{\modelname} - \text{BuboGPT}}}$ &
  \textcolor{ForestGreen}{\ding{51}} &
  \colorbox{increase}{+20.34\%} &
  \colorbox{increase}{+14.97\%} &
  \colorbox{increase}{+8.85\%} &
  \colorbox{increase}{+8.97\%} &
  \colorbox{increase}{+11.47\%} &
  \colorbox{increase}{+8.66\%} &
  \colorbox{increase}{+7.39\%} &
  \colorbox{increase}{+9.85\%} \\
  \bottomrule
\end{tabular}%
}
\vspace{0.05in}
\caption{\textbf{Audio referred image grounding results.} For AVSBench we follow the same train/test splits for all methods. We use the VGG-SS, Flickr-SoundNet, and PascalSound datasets only for evaluation. %\modelname outperforms SoTA by setting a new baseline.
}
\label{tab:spatial_grounding_boundingbox}
\vspace{-7mm}
\end{table}

\begin{table}[!t]
    \centering
    % Left table
    \begin{minipage}{0.48\textwidth}
        % \centering
        % \setlength{\tabcolsep}{3pt}
        % \fontsize{10}{9}\selectfont
        \renewcommand{\arraystretch}{0.7}
% \resizebox{\columnwidth}{!}{%
\begin{adjustbox}{width=\columnwidth,center}
\begin{tabular}{lc|c|c}
\toprule
                   &   & \textbf{LLP}          & \multicolumn{1}{c}{\textbf{AudioSet Strong}} \\ 
\multirow{-2}{*}{\textbf{Models}} &
  \multirow{-2}{*}{\textbf{Generalist?}} &
  \textbf{F1-score $\uparrow$} &
  \multicolumn{1}{c}{\textbf{F1-score $\uparrow$}} \\ \midrule
AVE \cite{tian2018audio}      & \textcolor{OrangeRed}{\ding{55}} & 35.47                  &  37.42                                            \\
AVSDN \cite{lin2019dual}     & \textcolor{OrangeRed}{\ding{55}} & 37.15                  &   41.48                                           \\
AVVP \cite{llpdataset}              & \textcolor{OrangeRed}{\ding{55}} & 48.93                  &   49.20                                           \\
% [0.5ex]\cdashline{1-4}\\[-1.0ex]
% \customdottedlinetabletemporal
\midrule
TimeChat \cite{timechat}          & \textcolor{ForestGreen}{\ding{51}} & 51.28                       &  54.66                                             \\
% VideoLlama         & \textcolor{ForestGreen}{\ding{51}} &                       &                                              \\
% X-InstructBLIP     & \textcolor{ForestGreen}{\ding{51}} & \multicolumn{1}{l|}{} &                                              \\ \hline
\midrule
% $\text{\modelname}_{\text{baseline}}$ & \textcolor{ForestGreen}{\ding{51}} & 52.13                      &  54.71                                            \\
\rowcolor{ThemeColor} 
% $\text{\modelname}_{\text{final}}$ 
\textbf{\modelname (ours)} &
  \textcolor{ForestGreen}{\ding{51}} & 
\cellcolor{ThemeColor}\textbf{54.96} & 
  \cellcolor{ThemeColor}\textbf{56.85} \\
  \midrule
  $\textcolor{blue}{\Delta_{\text{\modelname} - \text{TimeChat}}}$ &
  \textcolor{ForestGreen}{\ding{51}} &
    \colorbox{increase}{+7.18\%} &
  \colorbox{increase}{+4.01\%} \\
  \bottomrule
\end{tabular}%
% }
\end{adjustbox}
\vspace{0.05in}
\caption{\textbf{Image guided audio temporal localization results.} We report the segment level F1-scores and attribute our performance gain over specialist models to our multi-task learning strategy.}
\label{tab:audio_temporal_grounding}
    \end{minipage}
\hfill
\begin{minipage}{0.50\textwidth}
        \centering
\renewcommand{\arraystretch}{1.15}
\resizebox{\columnwidth}{!}{%
\begin{tabular}{l|c|c|c|c}
\toprule
\multicolumn{1}{c|}{} &
  \multicolumn{1}{c|}{\textbf{Type 1}} &
  \multicolumn{1}{c|}{\textbf{Type 2}} &
  \multicolumn{1}{c|}{\textbf{Type 3}} &
  \multicolumn{1}{c}{\textbf{Type 4}} \\
\multicolumn{1}{c|}{\multirow{-2}{*}{\textbf{Model}}} &
  \textbf{F1-score} $\uparrow$ &
  \textbf{F1-score} $\uparrow$ &
  \textbf{F1-score} $\uparrow$ &
  \textbf{F1-score} $\uparrow$ \\ \toprule
% AV LLM \cite{avllm}     &   0.64 &
% 0.67 &
% 0.59 &
% 0.73   \\
Macaw-LLM \cite{macawllm}  &    0.65 &
0.70 &
0.56 &
0.77  \\
PandaGPT \cite{pandagpt}      & 0.67 &
0.70 &
0.66 &
0.70 \\
VideoLlama \cite{videollama}    & 0.71 &
0.72 &
0.72 &
0.78 \\
BuboGPT \cite{bubogpt}       & 0.72 &
0.66 &
0.67 &
0.70 \\
X-InstructBLIP \cite{xinstructblip} & 0.73 &
0.72 &
0.72 &
0.80 \\
TimeChat \cite{timechat}      & 0.74 &
0.76 &
0.74 &
0.82 \\ \midrule
\rowcolor{ThemeColor} 
% $\text{\modelname} (ours) %_{\text{final}}
% $ 
\textbf{\modelname (ours)}
& 
\textbf{0.85} &
\textbf{0.83} &
\textbf{0.84} &
\textbf{0.88}  \\ 
  \midrule
   $\textcolor{blue}{\Delta_{\text{\modelname} - \text{TimeChat}}}$ &
  \colorbox{increase}{+14.86\%} &
  \colorbox{increase}{+9.21\%} &
  \colorbox{increase}{+13.51\%} &
  \colorbox{increase}{+7.32\%} \\
  \bottomrule
\end{tabular}%
}
\vspace{0.05in}
\caption{\textbf{Audio-Visual fact-checking} requires powerful reasoning capabilities across audio-visual modalities.}
% Methods such as TimeChat which is adept at temporal localization are found to perform well on tasks that require strong audio-segment level understanding. Whereas, BuboGPT demonstrates considerably well visual grounding capabilities 
% \RGnote{Same here. Maybe report average across all types, and show the separate performance only in Supp.}
\label{tab:factchecking_tasks}
% \vspace{2mm}
\end{minipage}
    % \vspace{-2mm}
    \vspace{-7mm}
\end{table}
\begin{figure}[t]
    \centering
    \includegraphics[width=\columnwidth]{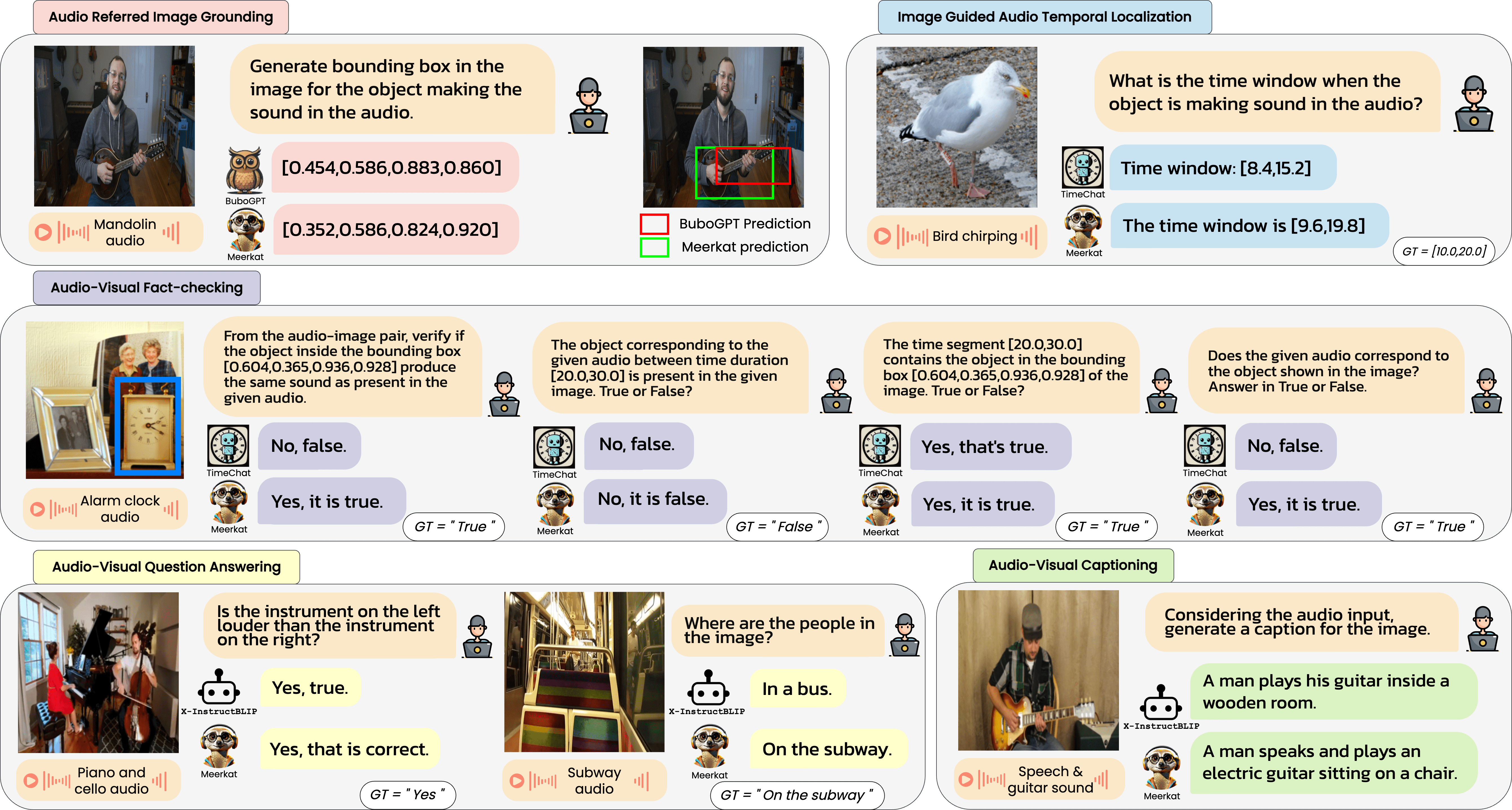}
    \vspace{-5mm}
    \caption{\textbf{Qualitative results.} We compare our method against its closest baselines on all downstream tasks. \modelname aided by our novel design approach and instruction tuning datasets achieves superior performance on spatio-temporal grounding as well as coarse-grained tasks by outperforming prior approaches.}
    \label{fig:qualitative_results}
    \vspace{-5mm}
\end{figure}
% \hspace{-1.8mm}
\vspace{1mm}
\noindent{\textbf{Image Guided Audio Temporal Localization (IGATL).}} When prompted to indicate a time interval within which a certain audio event occurs, \modelname is capable of producing accurate time bounds in the form \texttt{[tStart, tEnd]}, where \texttt{tStart} and \texttt{tEnd} are the start and end times, respectively. For all our experiments, we maintain the audio duration to be 30s. Different from prior visual grounding-based approaches \cite{ferret, vistallm, minigptv2}, we present a new audio event localization task by setting a new baseline. We attribute the superior performance of our method on fine-grained audio temporal localization task to our specially designed \weakalignmentmodule and \strongalignmentmodule modules, which ensure superior modality-specific guidance. Fig. \ref{fig:qualitative_results} demonstrates our model can locate a precise time interval associated with an audio event. Tab. \ref{tab:audio_temporal_grounding} reports the quantitative comparison of our method against other baselines.

\vspace{1mm}
\noindent{\textbf{Audio-Visual Fact-checking (AVFact).}} \label{sec:mmfact} In this section we introduce a new suite of tasks that involves a strong comprehension of the audio-visual semantic information. These tasks broadly require the model to analyze and verify whether a given statement about an audio-visual scenario holds or not. Although we do not use GT spatio-temporal annotations to train the model, we classify this task under the fine-grained category as the task requires the model to attend to a specific region/time interval as passed in the query. To alleviate inconsistencies in evaluation, we restrict the model's response to binary \textit{True/False} only. We divide these tasks into the following 4 categories:

% For example: \texttt{Verify if the object inside the <bbToken>} \texttt{produces the same sound as present in the given <audioToken>.} 

\noindent{\textit{Type 1:}} Given an audio-image pair, verify if the object within the bounding box produces sound that corresponds to the input audio.  

\noindent{\textit{Type 2:}} Given an audio snippet, verify whether its visual counterpart is present in the image or not. 

\noindent{\textit{Type 3:} Given an audio-image pair, verify if the object present within the provided bounding box produces sound that corresponds to the audio within a given time segment. 

\noindent{\textit{Type 4:}} Given an audio-image pair, verify if the supplied audio is related to the object within the provided bounding box. \\
In Tab. \ref{tab:factchecking_tasks} we contrast the performance of other baselines against \modelname on all four types of AVFact tasks.

\begin{table}[!t]
\centering
\renewcommand{\arraystretch}{0.92}
\resizebox{\columnwidth}{!}{
\begin{tabular}{lc|ccc|ccc|cccc}
\toprule
\multirow{2}{*}{\textbf{Model}} &  \multirow{2}{*}{\textbf{Generalist?}} & \multicolumn{3}{c|}{\textbf{AVQA}} & \multicolumn{3}{c|}{\textbf{MUSIC AVQA}} & \multicolumn{4}{c}{\textbf{VALOR-32K}} \\
& & \textbf{Exist $\uparrow$} &
\textbf{Localis $\uparrow$} &
\textbf{Temp $\uparrow$} &
\textbf{Exist $\uparrow$} &
\textbf{Localis $\uparrow$} &
\textbf{Temp $\uparrow$} &
\textbf{BLEU@4 $\uparrow$} &
\textbf{METEOR $\uparrow$} &
\textbf{ROUGE $\uparrow$} &
\textbf{CIDEr $\uparrow$} \\ \midrule
AVSD \cite{schwartz2019simple}      & \textcolor{OrangeRed}{\ding{55}} & 81.61 & 
58.79 & 
61.41 & 
-     &  
-      &  
-     & - & - & - & - 
\\
PanoAVQA \cite{yun2021pano} & \textcolor{OrangeRed}{\ding{55}} & 81.21 & 
59.33 & 
% 64.91 & 
% 64.22 & 
63.23 & 
% 66.64  &  
-       & 
-      &  
-     % & 
% -     & 
% -  
& - & - & - & - 
\\
ST-AVQA \cite{musicavqadataset}   & \textcolor{OrangeRed}{\ding{55}} & 81.81 & 
64.51 & 
% 70.80 & 
% 66.01 & 
63.23 & 
% 69.54  &   
-     &  
-      &  
-     % &  
% -     &  
% -
& - & - & - & - 
\\
CAD \cite{nadeem2024cad}                        & \textcolor{OrangeRed}{\ding{55}} & 
83.42 & 
73.97 & 
% 76.37 & 
% 74.88 & 
76.16 & 
% 76.96  &   
-     &  
-      & 
-      % & 
% -      &  
% -    
& - & - & - & - 
\\
AVST \cite{musicavqadataset}               & \textcolor{OrangeRed}{\ding{55}} & -      & -      &  -     % &  -      &  -     &  -     
& 
72.44 & 
65.54   & 
% 68.22 & 
% 63.31  & 
59.36 
& - & - & - & - \\
LAVISH \cite{lin2023vision}             & \textcolor{OrangeRed}{\ding{55}} & -      &  -     &  -    % &  -     &  -     &  -     
& 
73.83  & 
65.00  & 
% 73.28 & 
% 63.49  & 
60.81 
& - & - & - & - \\
LAST \cite{liu2024tackling}                       & \textcolor{OrangeRed}{\ding{55}} & -      &  -     &  -     % &  -     &  -     &   -    
&
76.21  & 
68.91  & 
% 75.23 & 
% 65.60 & 
60.60 
& - & - & - & - \\
SMPFF \cite{chen2021mm21}  & \textcolor{OrangeRed}{\ding{55}} & - & - & - & - & - & - & 7.59                      & 12.64                     & 28.69                     & 37.18                     \\
VALOR \cite{valor}                                   & \textcolor{OrangeRed}{\ding{55}} & - & - & - & - & - & -& 8.97                      & 14.88                     & 30.86                     & 55.73                     \\
% [0.5ex]\cdashline{1-12}\\[-2.0ex]
% \customdottedline
% \vspace{-0.5mm}
% \customdottedlineavqa
\midrule
Macaw-LLM \cite{macawllm}                   & \textcolor{ForestGreen}{\ding{51}} & 82.19       &    
74.86   & 
% 78.16      & 
% 77.54      & 
78.98      & 
% 78.34      &   
72.99           & 
71.28  & 
% 76.61          & 
% 67.77   &  
59.36    
& 9.36                          &  15.28                        & 33.31                         & 58.98 \\
PandaGPT \cite{pandagpt}                   & \textcolor{ForestGreen}{\ding{51}} & 83.38      & 
76.81      &  
% 78.92     & 
% 78.02      &  
79.11   & 
% 79.24  &  
78.48       &  
73.12     &  
% 79.06         & 
% 70.58  &  
65.85  
&    10.35                      & 16.92                         &  34.88                        & 61.22
\\
VideoLlama \cite{videollama}                 & \textcolor{ForestGreen}{\ding{51}} &   
84.48    & 
77.06      &  
% 79.90     & 
% 77.26      &   
81.36    & 
% 80.01      &  
81.21           & 
76.10  &  
% 82.90         &  
% 72.32   & 
67.52 
& 11.45                         & 17.39                         &   35.14                       &  63.63                
\\
X-InstructBLIP \cite{xinstructblip}              & \textcolor{ForestGreen}{\ding{51}} &   
85.53    & 
80.09     &  
% 81.14     &  
% 82.29     & 
83.91      &  
% 82.59     & 
80.28             & 
77.45 &  
% 83.89         & 
% 73.43   &  
68.83  
& 12.31                         &  18.82                        & 37.93                         & 65.73
\\
\midrule
\rowcolor{ThemeColor}
% $\text{\modelname}_{\text{final}}$ 
\textbf{\modelname (ours)} &
  \textcolor{ForestGreen}{\ding{51}} & \textbf{88.24}
   & \textbf{86.65}
   % % & \textbf{84.60}
   % % & \textbf{87.05}
   & \textbf{86.55}
   % % & \textbf{86.61}
   & \textbf{83.62}
   & \textbf{80.51}
   % % & \textbf{85.70}
   % % & \textbf{75.98}
   & \textbf{73.33} &
   \textbf{16.88} & \textbf{23.18} & \textbf{45.67} & \textbf{76.84}
   \\ 
   \midrule
   $\textcolor{blue}{\Delta_{\text{\modelname} - \text{X-InstructBLIP}}}$ &
  \textcolor{ForestGreen}{\ding{51}} &
  \colorbox{increase}{+3.17\%} &
  \colorbox{increase}{+8.19\%} &
  % \colorbox{increase}{+4.26\%} &
  % \colorbox{increase}{+5.78\%} &
  \colorbox{increase}{+3.15\%} &
  % \colorbox{increase}{+4.87\%} % &
  \colorbox{increase}{+4.16\%} &
  \colorbox{increase}{+3.95\%} &
  % \colorbox{increase}{+2.16\%} &
  % \colorbox{increase}{+3.47\%} &
  \colorbox{increase}{+6.54\%} &
  \colorbox{increase}{+37.12\%} &
  \colorbox{increase}{+23.17\%} &
  \colorbox{increase}{+20.41\%} &
  \colorbox{increase}{+16.9\%}
  \\
  \bottomrule
\end{tabular}%
}
% \end{adjustbox}
\vspace{0.05in}
\caption{\textbf{Quantitative results on AVQA and AV captioning tasks}. The reported numbers on AVQA dataset \cite{avqadataset} are on the val split. For the MUSIC-AVQA dataset \cite{musicavqadataset}, results are reported on the balanced test set. Here, Exist: Existential, Localis: Localisation, Temp: Temporal. Evaluation for AV captioning is done on VALOR-32K \cite{valor} val set. \modelname demonstrates strong coarse-grained understanding abilities.}
\label{tab:avqa_avcaptioning}
\vspace{-7mm}
\end{table}

\vspace{1mm}
\noindent{\textbf{Audio-Visual Question Answering (AVQA).}} Audio-visual question answering aims to answer questions encompassing both audio and visual modalities. We collect question-answer pairs from the AVQA \cite{avqadataset} and MusicAVQA \cite{musicavqadataset} datasets and augment them with instruction tuning templates (details in appendix) to prepare the data samples. We contrast our method against SoTA generalist and specialist models on the AVQA task in Tab. \ref{tab:avqa_avcaptioning}. We report the evaluation results on the other metrics like Count and Comp in the appendix.  

% add more details referring the table

\vspace{1mm}
\noindent{\textbf{Audio-Visual Captioning (AVC).}} This task learns how to generate text tokens conditioned on audio-visual inputs. In contrast to image/audio-only captioning methods, this requires strong multi-modal understanding and reasoning capabilities. We note that \modelname\ outperforms existing specialist and generalist models by a considerable margin and sets a new baseline on a recent benchmark dataset VALOR \cite{valor}, as shown in Tab. \ref{tab:avqa_avcaptioning}.

We argue that the seamless extension of \modelname to coarse-grained tasks is facilitated by the strong semantic understanding acquired by our model during training. This comprehension ability enables our model to effectively navigate and interpret the complexities inherent in coarse-grained tasks, showcasing the versatility and easy extensibility of our approach.

% \RGnote{I wonder do we need to add Meerkatbaseline in all the tables, given we have a dedicated table here on the different training objectives. Seem redundant and didn't add much to add it in those tables?}

\vspace{0.05in}
\customsubsection{Ablation Study} 
\label{sec:ablation}

% \RGnote{Table 7 is the most important and intuitive ablation. Other potential abaltions that can be done:

% \begin{itemize}
%     \item One thing we have emphasized is that we have proposed a unified framework for these five audio-visual tasks. One thing that people may question is the word ``unified''. Is there any benefit of training all five tasks together (namely multi-task)? What is the performance of multi-task vs. single task training?
%     \item Another question that may arise is the quality of LLM we used. How accurate and reliable are they? And how sensitive is our model performance with respect to a different LLM?
% \end{itemize}

% }

\noindent{\textbf{Weak vs. Strong Alignment.}} We ablate the quantitative effectiveness of our proposed weak and strong alignment modules in Tab. \ref{tab:ablation_alignment_module}. Without the \strongalignmentmodule module, the method's performance on the visual grounding task is considerably worse. For a similar reason, ablating this module in AVFact (Type 3), which requires region-level visual understanding, also shows inferior performance. For coarse-grained tasks (AV Captioning, AVQA), introducing $\mathcal{L}_{\text{OT}}$ boosts performance compared to the baseline. Overall, optimal performance is achieved when two objective functions work in tandem with optimal weight factors.

\begingroup
\setlength{\columnsep}{7pt}
\begin{wraptable}{r}{0.6\textwidth}
% \vspace{-12pt}
\vspace{-7.5mm}
\small
\centering
\setlength{\tabcolsep}{5pt}
\renewcommand{\arraystretch}{1.0}
\resizebox{\linewidth}{!}
{
\begin{tabular}{ccc|c|c|c|c|c}
\toprule
\multicolumn{3}{c|}{\textbf{Training Objective}} &
  \textbf{VGGSS} &
  \textbf{LLP} &
  \textbf{AVFact(T3)} &
  \textbf{AVQA}
  & \textbf{VALOR} 
  \\ 
\textbf{$\mathcal{L}_{\text{CE}}$} & \textbf{$\mathcal{L}_{\text{OT}}$} & \textbf{$\mathcal{L}_{\text{AC}}$} & \textbf{cIOU} $\uparrow$ & \textbf{F1-score} $\uparrow$ & \textbf{F1-score} $\uparrow$ 
& \textbf{Avg} $\uparrow$ 
&  \textbf{CIDEr} $\uparrow$ 
\\ \hline
\textcolor{ForestGreen}{\ding{51}} &
  \textcolor{OrangeRed}{\ding{55}} &
  \textcolor{OrangeRed}{\ding{55}} &
   42.93 &
   52.13 &
   0.76 &
   84.00 &
   71.52 
   \\
\textcolor{ForestGreen}{\ding{51}} &
  \textcolor{ForestGreen}{\ding{51}} &
  \textcolor{OrangeRed}{\ding{55}} &
   43.75 & 53.41
   & 0.78 
   & 85.91
   & 73.49
   \\
\textcolor{ForestGreen}{\ding{51}} &
  \textcolor{OrangeRed}{\ding{55}} &
  \textcolor{ForestGreen}{\ding{51}} & 46.83
   & 52.57
   & 0.81
   & 85.82
   & 73.14
   \\
\rowcolor{ThemeColor}\textcolor{ForestGreen}{\ding{51}} &
  \textcolor{ForestGreen}{\ding{51}} &
  \textcolor{ForestGreen}{\ding{51}} &
   \textbf{48.51} &
   \textbf{54.96} &
   \textbf{0.84} &
   \textbf{87.14} &
   \textbf{76.84} 
   \\ \bottomrule
\end{tabular}%
}
\vspace{-6pt}
\caption{\textbf{Ablation on different combinations of $\mathcal{L}_{\text{OT}}$ and $\mathcal{L}_{\text{AC}}$}. \modelname achieves optimal performance with a weighted linear combination of the 3 objective functions on all tasks. AVQA avg is calculated over Exist, Localis, and Temp.}
% We systematically analyze our design choice by ablating the critical components. We experiment with different combinations of $\mathcal{L}_{OT}$ and $\mathcal{L}_{AC}$. \modelname achieves optimal performance with a weighted linear combination of the 3 alignment modules on all downstream tasks.
\label{tab:ablation_alignment_module}
\vspace{-5mm}
\end{wraptable}

\noindent{\textbf{Evaluation on Pre-training Tasks.}} To study the effect of \textit{unified} pre-training, we evaluate our model under single task vs. multi-task learning setting. We gradually add datasets for each task and assess the model's performance. On quantitative evaluation, we note that our multi-task setting is indeed benefitting from each other in achieving superior performance as shown in Tab. \ref{tab:ablation_unified}. While the model trained on fine-grained tasks performs significantly well on the coarse-grained tasks, introducing the coarse-grained tasks in the training set doesn't have a considerable impact on ARIG, IGATL, and AVFact - underlining the importance of our collected fine-grained datasets. 

\noindent{\textbf{Full vs. LoRA Finetuning.}} We conduct experiments on different modes of LLM fine-tuning. As shown in Fig. \ref{fig:full_vs_lora_finetuning}, LoRA \cite{lora} based fine-tuning with r=32 achieves optimal performance. Lower values of r (4,16) performs poorly compared to 32 and we empirically find full-finetuning performs slightly worse than LoRA (r=32). We add more ablation results in the appendix.

\vspace{0.05in}
\customsubsection{Qualitative Analysis} 

Fig. \ref{fig:qualitative_results} illustrates the comparison of \modelname with its closest baseline on all downstream tasks. We observe that our model powered by the combination of \weakalignmentmodule and \strongalignmentmodule is equipped with finer region-level understanding compared to Bubo-GPT \cite{bubogpt}. Similarly, on image-guided audio temporal localization, our method outperforms TimeChat \cite{timechat}. We attribute the excellent performance of \modelname to the strong AV association learning backed by the instruction tuning data and multi-task learning set-up.   
% their model being trained for long video understanding is not able to replicate its performance at image-level supervision.
For the AVQA task, the recently proposed X-InstructBLIP \cite{xinstructblip} achieves comparable results. We argue that fuelled by a strong fine-grained understanding acquired through the pre-training stages, \modelname can extract additional contextual information from the visual modality. Our training paradigm emphasizes on both audio and visual modalities facilitating precise audio understanding by the model when compared against Video-LLaMA \cite{videollama}. Finally, on the AVFact tasks, our approach achieves superior performance due to its better multi-modal comprehension skills.      

\begin{figure}[t]
    \centering
    \begin{minipage}[b]{0.62\textwidth}
        \centering
\renewcommand{\arraystretch}{1.3}        

\resizebox{\columnwidth}{!}{%
\begin{tabular}{ccccc|c|c|c|c|c}
\toprule
\multicolumn{5}{c|}{\textbf{Pre-training Task}} &
  \multicolumn{1}{c|}{\textbf{VGG-SS}} &
  \multicolumn{1}{c|}{\textbf{LLP}} &
  \multicolumn{1}{c|}{\textbf{AVFact}} &
  \multicolumn{1}{c|}{\textbf{AVQA}} &
  \multicolumn{1}{c}{\textbf{VALOR}} \\
\multicolumn{1}{c}{\textbf{ARIG}} \hspace{1mm}&
  \multicolumn{1}{c}{\textbf{IGATL}} \hspace{1mm} &
  \multicolumn{1}{c}{\textbf{AVFC}} \hspace{1mm} &
  \multicolumn{1}{c}{\textbf{AVQA}} \hspace{1mm} &
  \multicolumn{1}{c|}{\textbf{AVC}} &
  \multicolumn{1}{c|}{\textbf{cIOU} $\uparrow$}  &
  \multicolumn{1}{c|}{\textbf{F1-score} $\uparrow$}  &
  \multicolumn{1}{c|}{\textbf{Avg F1-score} $\uparrow$}  &
  \multicolumn{1}{c|}{\textbf{Avg Acc.} $\uparrow$}  &
  \multicolumn{1}{c}{\textbf{CIDEr} $\uparrow$}  \\ \hline
\textcolor{ForestGreen}{\ding{51}} & \textcolor{OrangeRed}{\ding{55}}
   & \textcolor{OrangeRed}{\ding{55}}
   & \textcolor{OrangeRed}{\ding{55}}
   & \textcolor{OrangeRed}{\ding{55}}
   & 47.53
   & 18.73
   & 0.71
   & 77.22
   & 67.82
   \\
\textcolor{ForestGreen}{\ding{51}} &
  \textcolor{ForestGreen}{\ding{51}} & \textcolor{OrangeRed}{\ding{55}} 
   & \textcolor{OrangeRed}{\ding{55}}
   & \textcolor{OrangeRed}{\ding{55}}
   & 47.75
   & 54.26
   & 0.74
   & 79.74
   & 70.19
   \\
\textcolor{ForestGreen}{\ding{51}} &
  \textcolor{ForestGreen}{\ding{51}} &
  \textcolor{ForestGreen}{\ding{51}} & \textcolor{OrangeRed}{\ding{55}}
   & \textcolor{OrangeRed}{\ding{55}}
   & 48.17
   & 54.65
   & 0.83
   & 81.11
   & 72.13
   \\
\textcolor{ForestGreen}{\ding{51}} &
  \textcolor{ForestGreen}{\ding{51}} &
  \textcolor{ForestGreen}{\ding{51}} &
  \textcolor{ForestGreen}{\ding{51}} & \textcolor{OrangeRed}{\ding{55}}
   & 48.29
   & 54.82
   & 0.83
   & 86.68
   & 74.14
   \\
\textcolor{ForestGreen}{\ding{51}} &
  \textcolor{ForestGreen}{\ding{51}} &
  \textcolor{ForestGreen}{\ding{51}} &
  \textcolor{ForestGreen}{\ding{51}} &
  \textcolor{ForestGreen}{\ding{51}} &
  \cellcolor[HTML]{FAE8C8}{\textbf{48.51}} &
  \cellcolor[HTML]{FAE8C8}{\textbf{54.96}} &
  \cellcolor[HTML]{FAE8C8}{\textbf{0.85}} &
  \cellcolor[HTML]{FAE8C8}{\textbf{87.14}} & \cellcolor[HTML]{FAE8C8}{\textbf{76.84}}
   \\ \bottomrule
\end{tabular}%
}
\vspace{-1mm}
\captionof{table}{\textbf{We systematically analyze the effect of multi-task learning}. Here ARIG: audio referred image grounding, IGATL: image guided audio temporal localization, AVFC: audio-visual fact-checking, AVQA: audio-visual question answering, and AVC: audio-visual captioning. AVQA avg accuracy calculated over Exist, Localis, and Temp.}
\label{tab:ablation_unified}
    \end{minipage}%
    \hfill
    \begin{minipage}[b]{0.35\textwidth}
        \centering
        \resizebox{1.05\textwidth}{!}{
			\hspace{-2mm}\begin{tikzpicture}
            \begin{axis}[
                axis x line*=bottom,
    		  axis y line*=left,
                ymin=0, ymax=70,
        	xmin=0, xmax=5,
                xticklabel={\pgfmathparse{\tick}\pgfmathprintnumber{\pgfmathresult}},    
                xtick={0, 1, 3, 5},
                xtick distance=0.2, % Equally spaced tick labels
                xlabel={\tiny{\# Training Epochs}},
                ylabel={\tiny{cIOU}},
                legend style={at={(0.28,1.08)},anchor=north,font=\tiny,draw=none,fill=none},
                xlabel shift=-5pt,
                ylabel shift=-4pt,
                xticklabel style={font=\tiny},
                yticklabel style={font=\tiny},
		      label style={font=\tiny},
                width=\linewidth,
                ]
                \addplot[LinePlot1,mark=*,mark options={scale=0.5},style={thick}] coordinates {
                    (0,	6.44)
                    (1,	10.24)
                    (2,	25.21)
                    (3,	38.33)
                    (4,	41.37)
                    (5,	43.25)
                    % (0,	2.48)
                    % (1,	8.47)
                    % (2,	15.52)
                    % (3,	24.29)
                    % (4,	29.67)
                    % (5,	26.13)
                };
                \addlegendentry{\scalebox{0.5}{Full}}

                \addplot[LinePlot2,mark=*,mark options={scale=0.5},style={thick}] coordinates {
                    (0,	3.99)
                    (1,	12.84)
                    (2,	29.26)
                    (3,	41.33)
                    (4,	46.97)
                    (5,	47.51)
                };
                \addlegendentry{\scalebox{0.5}{LoRA r=32}}

                \addplot[LinePlot3,mark=*,mark options={scale=0.5},style={thick}] coordinates {
                    (0,	1.95)
                    (1,	9.82)
                    (2,	21.16)
                    (3,	32.36)
                    (4,	35.73)
                    (5,	37.19)
                    % (0,	6.44)
                    % (1,	10.24)
                    % (2,	25.21)
                    % (3,	38.33)
                    % (4,	41.37)
                    % (5,	43.25)
                };
                \addlegendentry{\scalebox{0.5}{LoRA r=16}}

                \addplot[LinePlot4,mark=*,mark options={scale=0.5},style={thick}] coordinates {
                    (0,	2.48)
                    (1,	8.47)
                    (2,	15.52)
                    (3,	24.29)
                    (4,	29.67)
                    (5,	26.13)
                    % (0,	1.95)
                    % (1,	9.82)
                    % (2,	21.16)
                    % (3,	32.36)
                    % (4,	35.73)
                    % (5,	37.19)
                };
                \addlegendentry{\scalebox{0.5}{LoRA r=4}}
            \end{axis}
        \end{tikzpicture}
        
% \begin{tikzpicture}
%         \begin{axis}[
%             axis x line*=bottom,
%             axis y line*=left,
%             xlabel={\# Iterations},
%             ylabel={Accuracy},
%             legend style={at={(0.75,0.3)},anchor=north},
%             xtick=data,
%             xticklabels={0.25, 0.45, 0.65, 0.75, 0.9, 1},
%             width=10cm,
%             height=6cm,
%             ]
%             \addplot[LinePlot1,mark=triangle,mark options={scale=0.85},style={thick}] coordinates {
%                 (0.25, 28.39)
%                 (0.45, 31.11)
%                 (0.65, 37.93)
%                 (0.75, 47.51)
%                 (0.9, 46.52)
%                 (1, 44.27)
%             };
%             \addlegendentry{Full finetuning}

%             \addplot[LinePlot2,mark=triangle,mark options={scale=0.85},style={thick}] coordinates {
%                 (0.25, 35.65)
%                 (0.45, 37.18)
%                 (0.65, 46.66)
%                 (0.75, 53.96)
%                 (0.9, 48.59)
%                 (1, 42.3)
%             };
%             \addlegendentry{LoRA finetuning}
%         \end{axis}
%     \end{tikzpicture}
%
		}
  % \vspace{-7mm}
		\captionof{figure}{\textbf{cIoU upper bound on VGG-SS} for Full vs. LoRA based finetuning.}
        \label{fig:full_vs_lora_finetuning}
    \end{minipage}
    % \label{fig:full_vs_lora_finetuning}
    \vspace{-4mm}
\end{figure}

\vspace{0.05in}
\customsubsection{Training Details} 
\vspace{0.05in}

We train the model for $5$ epochs and report results using the checkpoint with the best validation loss. We use 8 A100 GPUs for training with validation at the end of every epoch. Inspired by the recent success of Low-Rank Adaptation (LoRA) \cite{lora}, we use it to finetune the LLM. \modelname\ is trained using AdamW optimizer \cite{adamw}. We use a gradient accumulation step of $3$. Training our model takes around 52 hours for 5 epochs. We utilize DeepSpeed \cite{deepspeed} for optimization during the training process. The model is trained with a learning rate of $3 \times 10^{-5}$. The warmup ratio is $0.03$, along with a cosine learning rate scheduler. We use FP16 precision for both training and inference.

\vspace{0.1in}
\section{Conclusions and Future Works}
\label{sec:conclusion_and_future_works}
\vspace{-4mm}
We presented \modelname, a powerful multi-modal large language model adept at processing audio-visual inputs to comprehend fine-grained spatio-temporal information. Our novel audio-visual alignment strategy powered by the \weakalignmentmodule and \strongalignmentmodule modules instil strong compositional understanding into \modelname, thereby making it suitable for challenging tasks like audio-referred visual grounding, image to audio temporal localization, audio-visual fact-checking, etc. To pave the way for future research in this direction, we collect \ourdataset comprising 3M instruction tuning samples and introduce \taskunificationframework that unifies five challenging audio-visual learning tasks.
%leverages the synergy between vision and audio modalities to promote progress in AVLLMs. 
Extensive experiments demonstrate the effectiveness of our approach on a wide range of downstream tasks, consistently achieving state-of-the-art performance.

In future work, we plan to equip our model to address more challenging tasks like AV segmentation. We also plan to extend the model's capability to operate on videos and handle associated tasks such as video temporal grounding, and video summarization. Future work can also focus on collecting video-centric multi-modal training data and reasoning benchmarks for evaluation at scale. Finally, our work 
% being one of the first approaches to tackle fine-grained audio-visual tasks for LLMs 
opens up avenues to study robustness and compositional understanding of AV LLMs with fine-grained comprehension abilities. 
% such a family of

% \subsection{Language}
% All manuscripts must be in English.

\bibliographystyle{splncs04}
\bibliography{main}

\newpage

\begin{table}[!t]
\centering
\resizebox{\columnwidth}{!}{%
\begin{tabular}{c}
\toprule
\textbf{\includegraphics[height=20pt]{figures/meerkat_bg_removed.png}\modelname: Audio-Visual Large Language Model } \\
\textbf{for Grounding in Space and Time} \\
\textbf{\textcolor{blue}{Appendix}}\\
\bottomrule
\end{tabular}%
}
\end{table}

% \title{~\includegraphics[height=20pt]{figures/meerkat_bg_removed.png}\modelname: Audio-Visual Large Language Model for Grounding in Space and Time}

\noindent{In this appendix we provide additional details about: \\
% \ref{supplementary_video} Supplementary video  \\
\ref{data_preparation strategy} Data preparation strategy (referenced in Sec. 4.2 of main paper) \\
\ref{instruction_templates} Dataset instruction templates (referenced in Sec. 3.4 and Sec. 5.2) \\
\ref{dataset statistics} Dataset statistics and analysis \\
\ref{qualitative results} More qualitative results \\
\ref{more ablations} More ablations (referenced in Sec. 5.3)\\
\ref{comparison with contrastive loss} Comparison with contrastive loss (referenced in Sec. 3.2) \\
\ref{comparison with two stage training} Comparison with two-stage training (referenced in Sec. 3.2)\\
\ref{role of audio} Role of audio in AVQA task \\
% \ref{prefix vs interleaved} Prefix vs. interleaved prompting  \\
\ref{optimal transport} More on optimal transport (referenced in Sec. 3.2) \\
\ref{avsbench details} AVSBench data collection \\
% \ref{other image datasets} Other image datasets \\
\ref{comparison against imagebind} Comparison against ImageBind \\
\ref{other avqa metrics} Other Quantitative metrics on AVQA task (referenced in Sec. 5.2) \\
\ref{evaluation metrics} Evaluation metrics \\
\ref{failure_cases} Failure cases \\
\ref{ethics statement} Ethics statement

% \section{Supplementary Video}
% \label{supplementary_video}
% In this video, we provide several qualitative examples from all the downstream tasks to demonstrate the performance of \modelname. Experimental results show our approach is effective in handling both fine-grained and coarse-grained tasks. We also provide examples of failure cases and discuss them in Section \ref{failure_cases}. The use of headphones is recommended.

\section{Data Preparation Strategy} 
\label{data_preparation strategy}

\subsection{Adaptation of Public Datasets.}
To collect the image-audio pairs from video-based datasets and adapt them to our setup, we carefully choose one representative image from the video. We add task-wise dataset details in Fig. \ref{fig:dataset_division}. To this end, we design a semi-automated strategy as explained later in each task section.

% to pair with the input audio and text instruction. 

\subsection{Fine-grained Data Preparation}

\noindent{\textbf{Audio Referred Image Grounding (ARIG).}} For this task, the dataset collection consists of image-audio pairs from Openimages-AudioSet, Openimages-VGGSound, AVSBench, VGGSS, PASCAL Sound, and Flickr-Soundnet. Among these, for Openimages-AudioSet, Openimages-VGGSound and VGGSS we first obtain the top 3 image frames with the highest image-text CLIP similarity scores \cite{clip} and subsequently select the most suitable frame by manual inspection to form the image-audio pair. The frames are extracted from the video segment of interest (denoted in dataset annotation). Please refer to Tab. \ref{tab:op_audioset_vggsound_mapping} for the Openimages-AudioSet / VGGSound classwise associations. We refer to this look-up table while matching the corresponding classes.

% We next elaborate on the image-audio pair formation strategy for each dataset:

\begin{figure}[t]
    \centering
    \includegraphics[width=\textwidth]{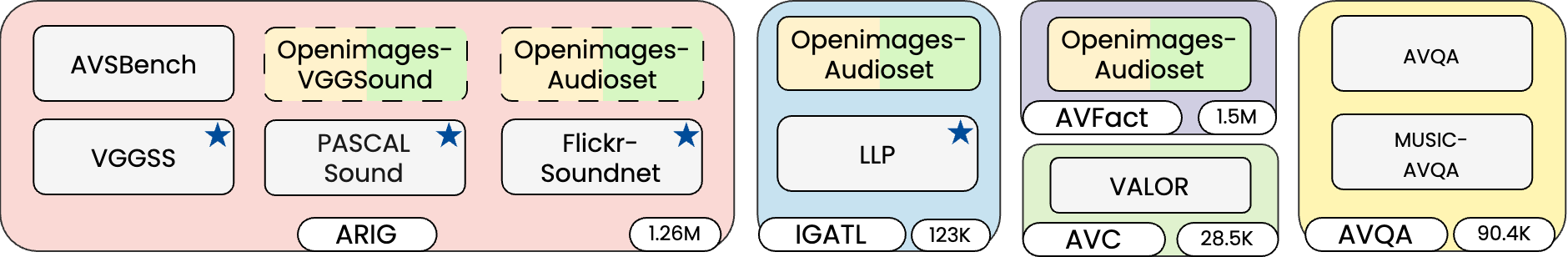}
    \vspace{-7mm}
    \caption{Task-wise dataset distribution. The bi-coloured cells denote collections of paired image-audio samples from public datasets following our data curation strategy while single-coloured cells signify direct adaptation. Datasets with dashed outlines are used only during model training while the ones with \includegraphics[width=2mm]{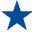} are reserved for zero-shot evaluations. Other datasets have a defined train/test split. Numbers in the bottom right represent the total \#samples present in each task.}
      \label{fig:dataset_division}
      % \vspace{-6mm}
\end{figure}

\begin{itemize}
 \item \textbf{Openimages-AudioSet:} For every sample, we obtain the [\texttt{start},\texttt{end}] time interval of the audio event of interest from the AudioSet dataset. Each sample is associated with an audio category. We use this class label while calculating the CLIP score with the image frames. We zero-pad and make length each audio piece 30s. 
 \item \textbf{Openimages-VGGSound:} We obtain the onset (\texttt{start}) of an event from the VGGSound dataset annotation and extract \textit{min}(\texttt{start} + 30, len(audio)) second snippet. If the len(audio) is less than 30s we zero pad to maintain the audio sequence length.  
 \item \textbf{AVSBench:} AVSBench comes with 5 to 6 frames along with the audio snippet. We manually choose the best frame that most closely relates to the audio event under consideration through manual inspection.  
 \item \textbf{VGGSS:} We follow a similar strategy as that of VGGSound. 
 \item \textbf{PASCAL Sound:} We choose 566 image samples from the PASCAL dataset \cite{pascal} ranging from 12 sounding classes and carefully pair them with AudioSet samples using the same protocol as Openimages-AudioSet. 
 \item \textbf{Flickr-Soundnet \cite{flickrsoundnetarda}:} Here we directly obtain the image audio pairs as released by the authors.
\end{itemize}

For all these cases we augment the image-audio pairs with our instruction tuning templates (refer to Section \ref{instruction_templates}).\\

\begin{table}[]
\centering
\resizebox*{\columnwidth}{0.98\textheight}{%
\begin{tabular}{l|l|l}
\toprule
\multicolumn{1}{c|}{\textbf{Openimages Label Name}} & \multicolumn{1}{c|}{\textbf{Audioset Label Name}} & \multicolumn{1}{c}{\textbf{VGGSound Label Name}} \\  \midrule
Aircraft & Aircraft & airplane \\  
Alarm clock & Alarm clock & alarm clock ringing \\  
Ambulance & Ambulance (siren) & ambulance siren \\  
Bicycle & Bicycle, tricycle & -- \\  
Bird & Bird & bird chirping, tweeting \\  
Blender & Blender, food processor & electric blender running \\  
Boat & Boat, Water vehicle & sailing \\  
Bus & Bus & helicopter \\  
Camera & Camera & -- \\  
Cannon & -- & firing cannon \\
Car & Car & race car, auto racing \\  
Cat & Cat & cat meowing \\  
Cattle & -- & cattle mooing \\
Ceiling fan & Mechanical fan & running electric fan \\
Cello & -- & playing cello \\
Chainsaw & Chainsaw & chainsawing trees \\  
Cheetah & Roaring cats (lions, tigers) & cheetah chirrup \\  
Chicken & Fowl & -- \\  
Chime & Chime & wind chime \\  
Clock & Clock & -- \\  
Computer keyboard & Computer keyboard & typing on computer keyboard \\  
Computer mouse & Mouse & -- \\  
Corded phone & Dial tone & cell phone buzzing \\  
Cutting board & Chopping (food) & chopping food \\ 
Dagger & Knife & -- \\  
Digital clock & -- & alarm clock ringing \\
Dog & Dog & dog baying \\  
Door & Door & -- \\  
Door handle & Doorbell & door slamming \\  
Drill (Tool) & Drill & -- \\  
Drum & -- & playing drum kit \\
Duck & Quack & duck quacking \\  
Eagle & -- & eagle screaming \\
Elephant & -- & elephant trumpeting \\
Fireplace & Fire & -- \\  
Fixed-wing aircraft & Fixed-wing aircraft, airplane & -- \\  
Fountain & Waterfall & -- \\  
Fox & Canidae, wild dogs, wolves & fox barking \\  
French horn & -- & playing french horn \\
Frog & Frog & frog croaking \\  
Girl & Female speech, woman speaking & -- \\  
Glasses & Glass & -- \\  
Goat & Goat & goat bleating \\  
Golf cart & Cart & -- \\  
Goose & Ducks, geese, waterfowl & goose honking \\  
Grinder & -- & electric grinder grinding \\
Guitar & -- & playing acoustic guitar \\
Hair dryer & Hair dryer & hair dryer drying \\  
Hammer & Hammer & -- \\  
Hand dryer & Hair dryer & -- \\  
Handgun & Gunshot, gunfire & machine gun shooting \\  
Harmonica & -- & playing harmonica \\
Harp & -- & playing harp \\
Harpsichord & -- & playing harpsichord \\
Helicopter & Helicopter & helicopter \\  
Horse & Horse & horse neighing \\  
Human face & Female speech, woman speaking|Male speech, man speaking & -- \\  
Infant bed & -- & baby crying \\
Ipod & Music & -- \\  
Jaguar (Animal) & Roaring cats (lions, tigers) & cheetah chirrup \\  
Jet ski & Jet engine & skiing \\  
Kettle & Steam whistle & -- \\  
Kitchen knife & Knife & -- \\  
Kitchen utensil & Kitchen and dining room sounds & -- \\  
Knife & Knife & -- \\  
Land vehicle & Vehicle & car passing by \\  
Laptop & Typing & typing on computer keyboard \\  
Leopard & Roar & -- \\  
Light switch & Clicking & -- \\  
Limousine & Car & -- \\  
Lion & Roar & lions roaring \\  
% \bottomrule
\end{tabular}%
}
\end{table}

\begin{table}[]
    \centering
    \resizebox*{\columnwidth}{0.94\textheight}{
    \begin{tabular}{l|l|l}
         Magpie & Bird & magpie calling \\  
Mammal & Animal & -- \\  
Man & Male speech, man speaking & -- \\  
Mechanical fan & Mechanical fan & running electric fan \\  
Microphone & Microphone & -- \\  
Microwave oven & Microwave oven & -- \\  
Missile & -- & missile launch \\
Mixer & Blender, food processor & -- \\  
Mobile phone & Telephone & cell phone buzzing \\  
Motorcycle & Motorcycle & driving motorcycle \\  
Mouse & Mouse & -- \\  
Musical instrument & Music & orchestra \\  
Musical keyboard & -- & playing piano \\
Oboe & -- & playing oboe \\
Otter & -- & otter growling \\
Owl & Owl & owl hooting \\  
Paper cutter & Scissors & ripping paper \\  
Parrot & Bird & parrot talking \\  
Person & Female speech, woman speaking|Male speech, man speaking & -- \\  
Piano & -- & playing piano \\
Pig & Pig & pig oinking \\  
Popcorn & Burst, pop & popping popcorn \\  
Power plugs and sockets & Power tool & -- \\  
Pressure cooker & Steam & -- \\  
Printer & Printer & printer printing \\  
Rabbit & Rodents, rats, mice & -- \\  
Ratchet (Device) & Ratchet, pawl & -- \\  
Raven & Crow & crow cawing \\  
Reptile & Snake & -- \\  
Rifle & Machine gun & machine gun shooting \\  
Rocket & -- & missile launch \\
Saxophone & -- & playing saxophone \\
Sea lion & -- & sea lion barking \\
Segway & Non-motorized land vehicle & -- \\  
Sewing machine & Sewing machine & using sewing machines \\  
Sheep & Sheep & -- \\  
Shotgun & Gunshot, gunfire & -- \\  
Shower & Shower & -- \\  
Skateboard & Skateboard & skateboarding \\  
Ski & -- & skiing \\
Snail & -- & hail \\
Snake & Snake & snake rattling \\  
Snowboard & Skateboard & skiing \\  
Snowmobile & Motorcycle & -- \\  
Snowplow & Lawnmower & -- \\  
Spoon & Kitchen and dining room sounds & -- \\  
Stationary bicycle & Bicycle, tricycle & driving motorcycle \\  
Swan & Quack & -- \\  
Swimming pool & Water & -- \\  
Sword & Knife & -- \\  
Table tennis racket & -- & playing table tennis \\
Tablet computer & Computer keyboard & typing on computer keyboard \\  
Tap & Tap & -- \\  
Taxi & Car & hail \\  
Telephone & Telephone & telephone bell ringing \\
Television & Television & -- \\  
Tiger & Roar & -- \\  
Toilet & Toilet flush & toilet flushing \\  
Train & Train & -- \\  
Truck & Truck & -- \\  
Trombone & -- & playing trombone \\
Trumpet & -- & playing trumpet \\
Turkey & Turkey & -- \\  
Unicycle & Bicycle bell & -- \\  
Van & Car & -- \\
Vehicle & Vehicle & vehicle horn, car horn, honking \\
Violin & -- & playing violin, fiddle \\
Wall clock & Clock & alarm clock ringing \\
Washing machine & Washing machine & -- \\
Watch & Clock & -- \\
Wine glass & Glass & -- \\
Whale & -- & whale calling \\
Woman & Female speech, woman speaking|Male speech, man speaking & -- \\
Woodpecker & Wood & woodpecker pecking tree \\
\bottomrule
    \end{tabular}%
    }
    \caption{\textbf{Image audio class mapping.} We associate the image and audio classes from the Openimages and the AudioSet / VGGSound datasets and prepare a lookup table through careful manual inspection.}
    \label{tab:op_audioset_vggsound_mapping}
\end{table}

\noindent{\textbf{Image Guided Audio Temporal Localization (IGATL).}}
\begin{itemize}
\item \textbf{Openimages-AudioSet (Strong):} While curating the image samples we follow a similar strategy as before. To ensure a fair assessment we choose audio snippets that are considerably longer than the event of interest (EoI). However, through manual inspections, we ensure that the EoI lies within the extracted audio piece.

% with the exception that in AudioSet Strong (\textcolor{red}{also write about Audioset}) we manually inspect and collect 30s within which the audio event occurs. 

% \textcolor{magenta}{In terms of audio collection, to adapt AudioSet Strong into our temporal grounding dataset, we manually ensure that samples chosen have the respective labels only in the provided ground truth time-window. } \textcolor{red}{is this required to be mentioned? in case someone finds otherwise later on after dataset release, it might be an issue.}

% \section{}

\begin{table}[]
\centering
\resizebox*{\columnwidth}{0.88\textheight}{%
\begin{tabular}{c | p{15cm}}
\toprule
\multicolumn{1}{c|}{\textbf{Task}} &
  \multicolumn{1}{c}{\textbf{Example Instruction}} \\ \midrule
\multicolumn{1}{c|}{\cellcolor[HTML]{FAD9D5}} &
  $\bullet$ Given the audio and image pair, identify the object category of the audio. Now, provide a bounding box for that object in the image. The answer should be in the form [<obj>,xLeft,yTop,xRight,yBottom]. <obj> represents the object category. xLeft,yTop are coordinates of the top-left corner and xRight,yBottom are coordinates of the bottom-right corner of the bounding box. The coordinates should be within the range 0 to 1. \\
\multicolumn{1}{c|}{\cellcolor[HTML]{FAD9D5}} &
  $\bullet$ From the given audio and image pair first identify the object category of the audio. Then localize the corresponding object in the image by providing a bounding box. The answer should be in the form [<obj>,xLeft,yTop,xRight,yBottom]. <obj> represents the object category. xLeft,yTop are coordinates of the top-left corner and xRight,yBottom are coordinates of the bottom-right corner of the bounding box. The coordinates should be within the range 0 to 1. \\
\multicolumn{1}{c|}{\cellcolor[HTML]{FAD9D5}} &
  $\bullet$ Given the audio and image pair, identify the object category of the audio. Now, localize the object in the image by providing a bounding box. The answer should be in the form [<obj>,xLeft,yTop,xRight,yBottom]. <obj> represents the object category. xLeft,yTop are coordinates of the top-left corner and xRight,yBottom are coordinates of the bottom-right corner of the bounding box. The coordinates should be within the range 0 to 1. \\
\multicolumn{1}{c|}{\cellcolor[HTML]{FAD9D5}ARIG} &
  $\bullet$ Considering the audio and image pair, determine the object class of the audio. Next, localize the same object in the image by providing a bounding box. The answer should be in the form [<obj>,xLeft,yTop,xRight,yBottom]. <obj> represents the class of the object. xLeft,yTop are coordinates of the top-left corner and xRight,yBottom are coordinates of the bottom-right corner of the bounding box. The coordinates should be within the range 0 to 1. \\
\multicolumn{1}{c|}{\cellcolor[HTML]{FAD9D5}} &
  $\bullet$ Considering the audio and image pair, recognize the object category of the audio. Subsequently, draw a bounding box around that object shown in the image. The answer should be in the form [<obj>,xLeft,yTop,xRight,yBottom]. <obj> represents the category of the object. xLeft,yTop are coordinates of the top-left corner and xRight,yBottom are coordinates of the bottom-right corner of the bounding box. The coordinates should be within the range 0 to 1. \\
\multicolumn{1}{c|}{\multirow{-6}{*}{\cellcolor[HTML]{FAD9D5}}} &
  $\bullet$ Considering the audio and image pair, recognize the object category of the audio. Next, draw a bounding box around that object in the image. The answer should be in the form [<obj>,xLeft,yTop,xRight,yBottom]. <obj> represents the category of the object. xLeft,yTop are coordinates of the top-left corner and xRight,yBottom are coordinates of the bottom-right corner of the bounding box. Ensure the bounding box is within the range 0 to 1. \\ \midrule
\cellcolor[HTML]{C7E2F0} &
  $\bullet$ Identify the object category from the image. Now, find the time duration in the audio where that object is making the sound. The output should be in the form (tStart,tEnd) where tStart and tEnd are the start and end times respectively. tStart is less than tEnd. The minimum value of tStart is 0. The maximum value of tEnd is 30. \\
\cellcolor[HTML]{C7E2F0} &
  $\bullet$ Given the image, identify the object category. Next, output the time window in the audio where that object is making the sound. The output should be in the form (tStart,tEnd) where tStart and tEnd are the start and end times respectively. tStart is less than tEnd. The minimum value of tStart is 0. The maximum value of tEnd is 30. \\
\cellcolor[HTML]{C7E2F0} &
  $\bullet$ Which object do you see in the image? Please find the time window in the audio where that object is making the sound. The output should be in the form (tStart,tEnd) where tStart and tEnd are the start and end times respectively. tStart is less than tEnd. The minimum value of tStart is 0. The maximum value of tEnd is 30. \\
\cellcolor[HTML]{C7E2F0}IGATL &
  $\bullet$ Recognise the object category from the image. Now, indicate the time duration in the audio where that object is making the sound. The output should be in the form (tStart,tEnd) where tStart and tEnd are the start and end times respectively. tStart is less than tEnd. The minimum value of tStart is 0. The maximum value of tEnd is 30. \\
\multirow{-5}{*}{\cellcolor[HTML]{C7E2F0}} &
  $\bullet$ What is the category of the object that you see in the image? Now, indicate the temporal duration in the audio where that object is making the sound. The output should be in the form (tStart,tEnd) where tStart and tEnd are the start and end times respectively. tStart is less than tEnd. The minimum value of tStart is 0. The maximum value of tEnd is 30. \\ 
  \midrule
% \cellcolor[HTML]{D0CEE2} &
%   $\bullet$ From the audio-image pair, verify if the object inside the bounding box <placeholder\_bbox> produce the same sound as present in the given audio. Answer in True or False. \\
% \cellcolor[HTML]{D0CEE2} &
%   $\bullet$ Listen to the audio in the time window <placeholder\_time>. Verify if the same object is present in the image. True or False? \\
% \cellcolor[HTML]{D0CEE2}AVFact &
%   $\bullet$ Does the object inside the bounding box <placeholder\_bbox> of the image produce the same sound as within the time duration <placeholder\_time> in the given audio? Answer in True or False. \\
% \multirow{-4}{*}{\cellcolor[HTML]{D0CEE2}} &
%   $\bullet$ Here is an audio-image pair. Does the given audio correspond to the object shown in the image? Answer in True or False. \\ \midrule
% \cellcolor[HTML]{E0F2CE} &
%   $\bullet$ How many instruments are sounding in the image? \\
% \cellcolor[HTML]{E0F2CE} &
%   $\bullet$ Which is the musical instrument that sounds at the same time as the <Object>? \\
% \cellcolor[HTML]{E0F2CE} &
%   $\bullet$ Is the <Object> on the <LR> louder than the <Object> on the <LR>? \\
% \cellcolor[HTML]{E0F2CE} &
%   $\bullet$ Is there a voiceover? \\
% \multirow{-5}{*}{\cellcolor[HTML]{E0F2CE}AVQA} &
%   $\bullet$ Is the <Object> playing longer than the <Object>? \\ \midrule
% \cellcolor[HTML]{FFF5B3}AVC &
%   $\bullet$ Considering the audio input, generate a caption for the image. \\ 
  % \bottomrule
\end{tabular}%
}
% \caption{Task instructions.}
% \label{tab:task_ins}
\end{table}

\begin{table}[t]
% \vspace{-60mm}
    \resizebox{\columnwidth}{!}{%
\begin{tabular}{c | p{15cm}}
\toprule
\cellcolor[HTML]{D0CEE2} & $\bullet$ Does the object inside the bounding box <placeholder\_bbox> of the image produce the same sound as in the given audio? Answer in True or False. \\
\cellcolor[HTML]{D0CEE2} & $\bullet$ Given the image, does the object inside the bounding box <placeholder\_bbox> produce the same sound as in the given audio? Answer in True or False. \\
\cellcolor[HTML]{D0CEE2} & $\bullet$ The object inside the bounding box <placeholder\_bbox> of the image produces the same sound as in the given audio. True or False? \\
         \cellcolor[HTML]{D0CEE2} &
  $\bullet$ From the audio-image pair, verify if the object inside the bounding box <placeholder\_bbox> produces the same sound as present in the given audio. Answer in True or False. \\
  \cellcolor[HTML]{D0CEE2} & $\bullet$ The object in the given audio between time duration <placeholder\_time> is present in the image. True or False?  \\
\cellcolor[HTML]{D0CEE2} & $\bullet$ Listen to the audio in the time window <placeholder\_time>. Does this object exist in the image? Answer in True or False. \\
\cellcolor[HTML]{D0CEE2} & 
  $\bullet$ Listen to the audio in the time window <placeholder\_time>. Verify if the same object is present in the image. True or False? \\
  \cellcolor[HTML]{D0CEE2} AVFact & 
  $\bullet$ The time segment <placeholder\_time> contains the object as present in the image. True or False? \\
  \cellcolor[HTML]{D0CEE2} &
  $\bullet$ Listen to the audio in the time window <placeholder\_time>. The same object is within the bounding box <placeholder\_bbox> in the image. True or False? \\
\cellcolor[HTML]{D0CEE2} &
  $\bullet$ Does the object inside the bounding box <placeholder\_bbox> of the image produce the same sound as within the time duration <placeholder\_time> in the given audio? Answer in True or False. \\
    \cellcolor[HTML]{D0CEE2} &
  $\bullet$ The object inside the bounding box <placeholder\_bbox> of the image produces the same sound as in the time segment <placeholder\_time> of the audio. True or False? \\
  \cellcolor[HTML]{D0CEE2} &
  $\bullet$ The time segment <placeholder\_time> contains the object in the bounding box <placeholder\_bbox> of the image. True or False? \\
\multirow{-7}{*}{\cellcolor[HTML]{D0CEE2}} &
  $\bullet$ Here is an audio-image pair. Does the given audio correspond to the object shown in the image? Answer in True or False. \\
  \cellcolor[HTML]{D0CEE2} & $\bullet$ Does the given audio correspond to the object shown in the image? Answer in True or False. \\
\cellcolor[HTML]{D0CEE2} & $\bullet$ Does the given audio associate with the object shown in the image? Answer in True or False. \\
\cellcolor[HTML]{D0CEE2} & $\bullet$ Here is an audio-image pair. Does the given image associate with the object sounding in the audio? Answer in True or False. \\
\midrule
\cellcolor[HTML]{E0F2CE} &
  $\bullet$ How many instruments are sounding in the image? \\
\cellcolor[HTML]{E0F2CE} &
  $\bullet$ Which is the musical instrument that sounds at the same time as the <Object>? \\
\cellcolor[HTML]{E0F2CE} &
  $\bullet$ Is the <Object> on the <LR> louder than the <Object> on the <LR>? \\
\cellcolor[HTML]{E0F2CE} &
  $\bullet$ Is there a voiceover? \\
\multirow{-5}{*}{\cellcolor[HTML]{E0F2CE}AVQA} &
  $\bullet$ Is the <Object> playing longer than the <Object>? \\ \midrule
\cellcolor[HTML]{FFF5B3}AVC &
  $\bullet$ Considering the audio input, generate a caption for the image. \\ \bottomrule
    \end{tabular}%
    }
    \caption{\textbf{Task wise instructions template}.}
    \label{tab:task_ins}
\end{table} 

% Similar as before, in LLP we are given the [\texttt{start}, \texttt{end}] of certain events where \texttt{end} - \texttt{start} $\leq$ 10s.

% ###### LLP dataset adaptation   ##########

% talk about time window, frame selection criteria, which samples were not considered (if there are multiple subjects with same category eg speech we consider one etc. )  write all details

\item \textbf{LLP:} The LLP dataset provides fine-grained temporal annotations of the audio events in the format [\texttt{onset}, \texttt{offset}]. One representative image is chosen from within this time segment. While preparing our test set, we restrict ourselves to one category per video and their corresponding \texttt{onset} and \texttt{offset} values to rule out overlapping events within the same time interval.

% For the visual modality, since the object is present in the \texttt{start\_time}-\textit{th} time may be different from the actual audio of the event, we take the frame that is present in the \texttt{onset}-\textit{th} time.

\end{itemize}

\noindent{\textbf{Audio-Visual Fact-checking (AVFact).}} 

\begin{itemize}
\item \textbf{Openimages-AudioSet:} For \textit{Type 1} we collect samples from the AudioSet split while for \textit{Type 2}, \textit{Type 3}, \textit{Type 4} we choose samples from AudioSet Strong split as it consists of time-sensitive grounding information which is used in these three types of queries. For the image collection, we follow the same strategy as before.

\end{itemize}

\subsection{Coarse-grained Data Preparation}
For the coarse-grained tasks, we resort to direct adaptations of publicly available datasets.

\noindent{\textbf{Audio-Visual Question Answering.}}
In the absence of audio class labels, we manually inspect the video to obtain the most suitable frame for each sample. 
\begin{itemize}
\item \textbf{AVQA:} The AVQA dataset contains the \texttt{start} time stamp which denotes the onset of the event of interest. We follow the same train/test split as proposed by the authors \cite{avqadataset}.  
\item \textbf{MUSIC-AVQA:} We crop the 30s from the original 1-minute-long video sequence within which the event of interest lies. 
\end{itemize}

\noindent{\textbf{Audio-Visual Captioning (AVC).}
\begin{itemize}
\item \textbf{VALOR-32K:} Each sample in the VALOR dataset comprises an elaborate caption of the audio-visual scene. We leverage this caption to calculate the CLIP similarity score between the image-text pair and obtain the top 3 most relevant frames from within the 10s long annotation as provided by the authors. Finally, we choose one representative frame through manual inspection.   

\end{itemize}

\section{Dataset Instruction Templates}
\label{instruction_templates}

We add task-wise sample instruction templates in Tab. \ref{tab:task_ins}. To make the instruction tuning robust and incorporate sufficient diversity, we manually write a few instructions and prompt GPT-3.5 \cite{gpt3} to generate different variants. We further refine the instruction templates using GPT-4 \cite{gpt4}. Note that for AVQA and AV captioning tasks, we restrict ourselves to the questions and captions provided by the authors.

% \section{Dataset Image Frame Selection Criteria}

\section{Dataset Statistics and Analysis}
\label{dataset statistics}

\begin{figure}[!t]
  \centering
  \begin{subfigure}{0.48\textwidth}
    \centering
    \includegraphics[width=\linewidth]{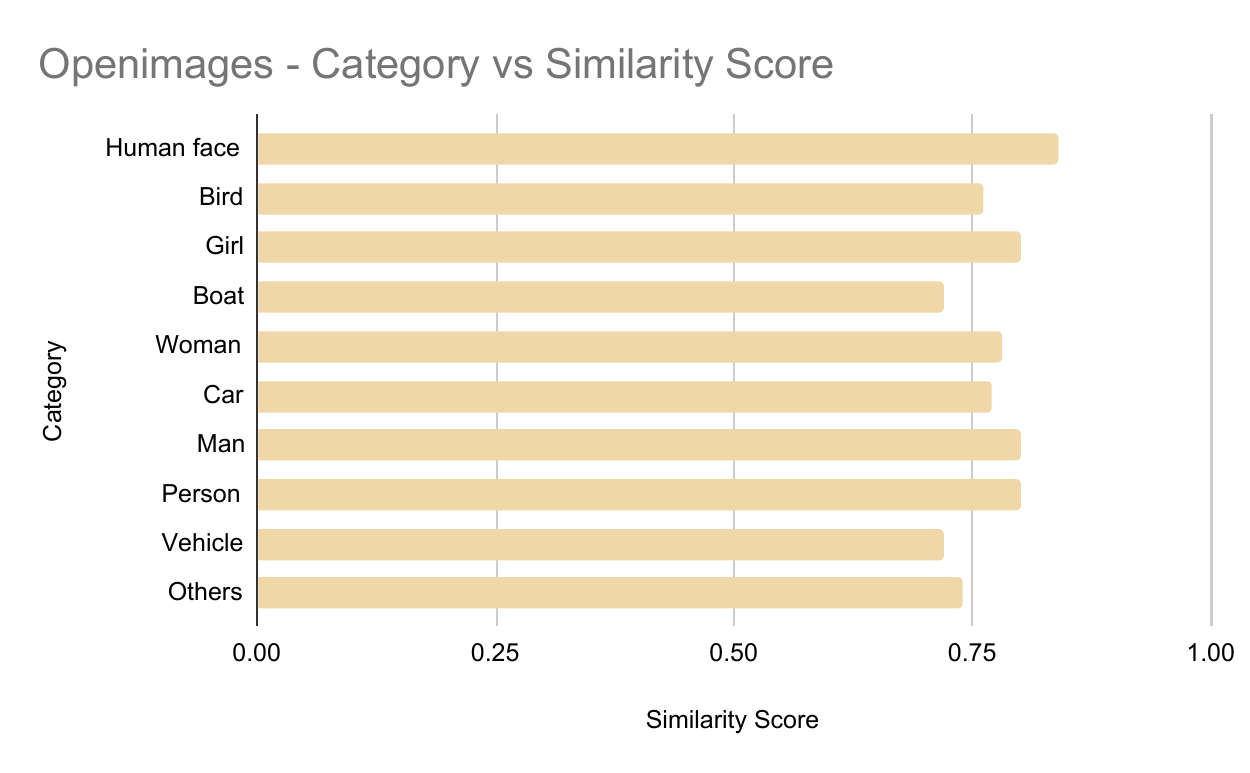}
    \caption{Image-audio similarity scores across samples in Openimages-AudioSet.}
    \label{fig:image_audio_similarity_score}
  \end{subfigure}
  \hfill
  \begin{subfigure}{0.48\textwidth}
    \centering
    \includegraphics[width=\linewidth]{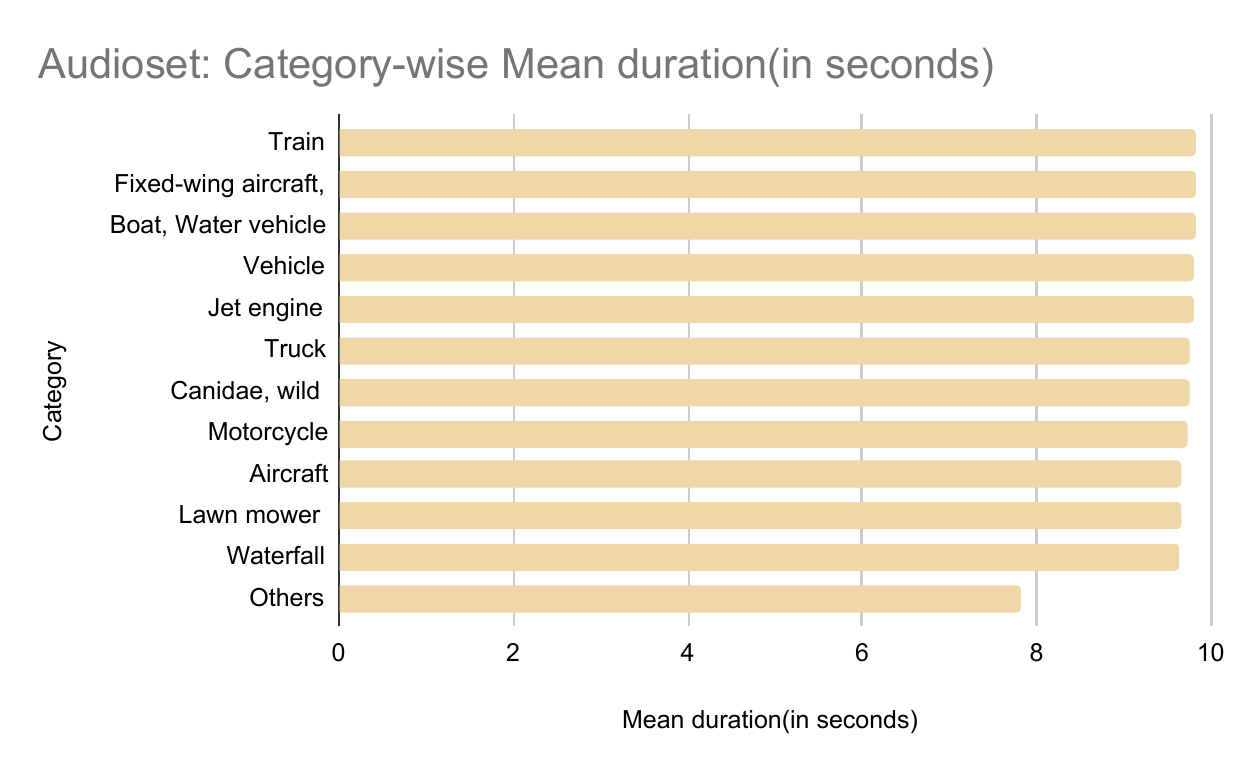}
    \caption{Category-wise average duration of chosen audio samples from AudioSet dataset.}
    \label{fig:audioset_cat_avg_duration}
  \end{subfigure}
  \\
  \begin{subfigure}{0.48\textwidth}
    \centering
    \includegraphics[width=\linewidth]{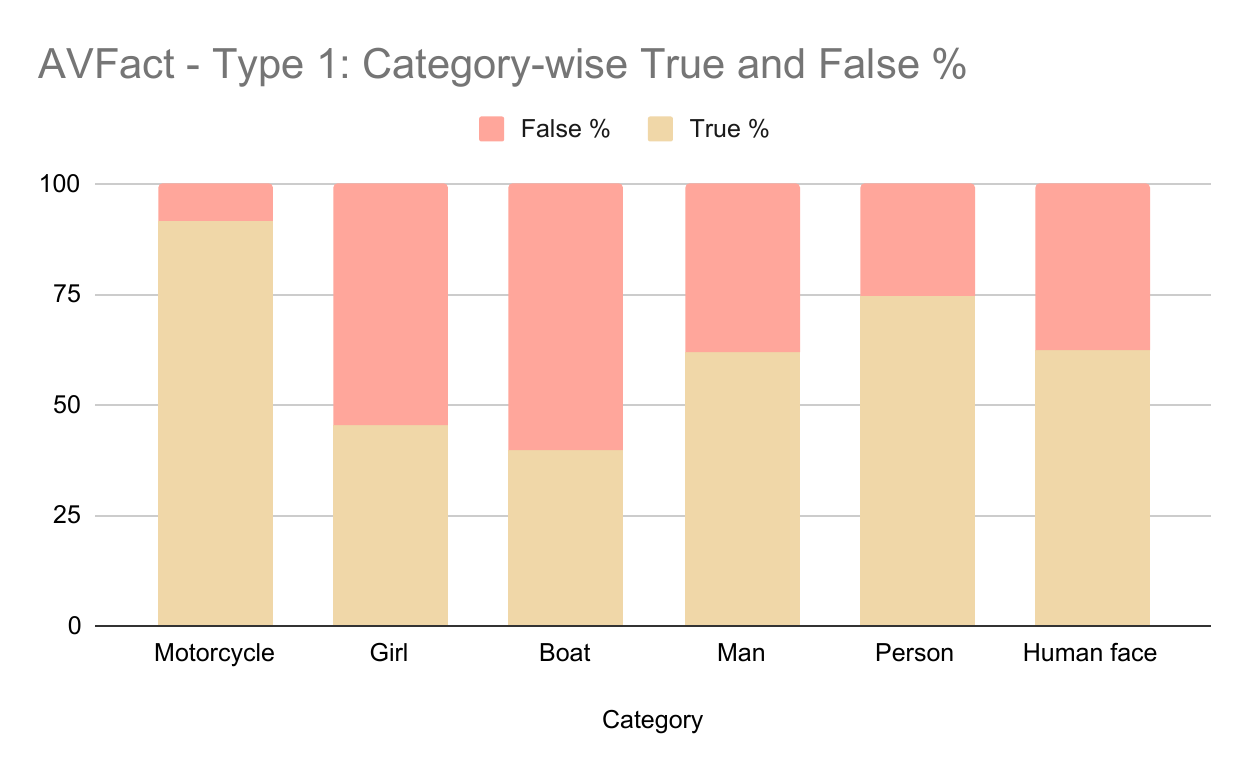}
    \caption{AVFact - Type 1: Top-6 category-wise number of True and False samples.}
    \label{fig:stat_type1}
  \end{subfigure}
  \hfill
  \begin{subfigure}{0.48\textwidth}
    \centering
    \includegraphics[width=\linewidth]{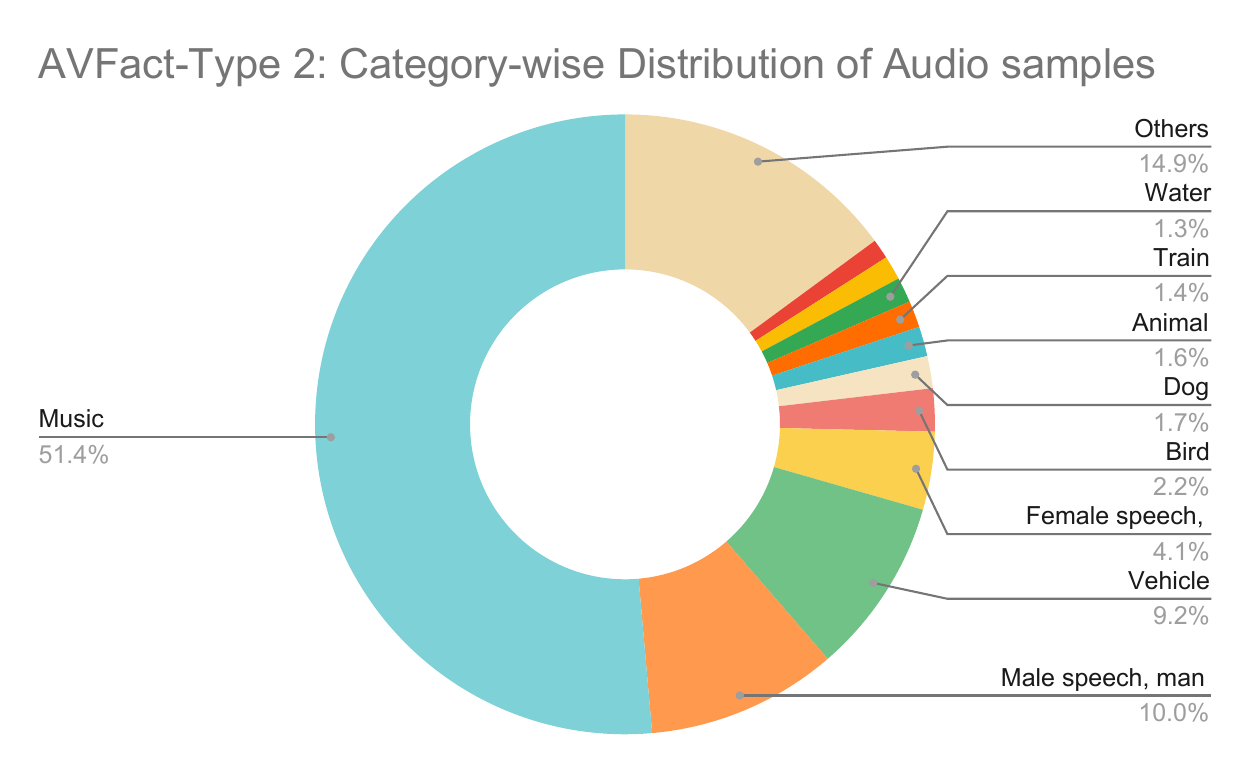}
    \caption{AVFact - Type 2: Category-wise distribution of chosen audio samples from AudioSet dataset.}
    \label{fig:stat_type2}
  \end{subfigure}
  \\
  \begin{subfigure}{0.48\textwidth}
    \centering
    \includegraphics[width=\linewidth]{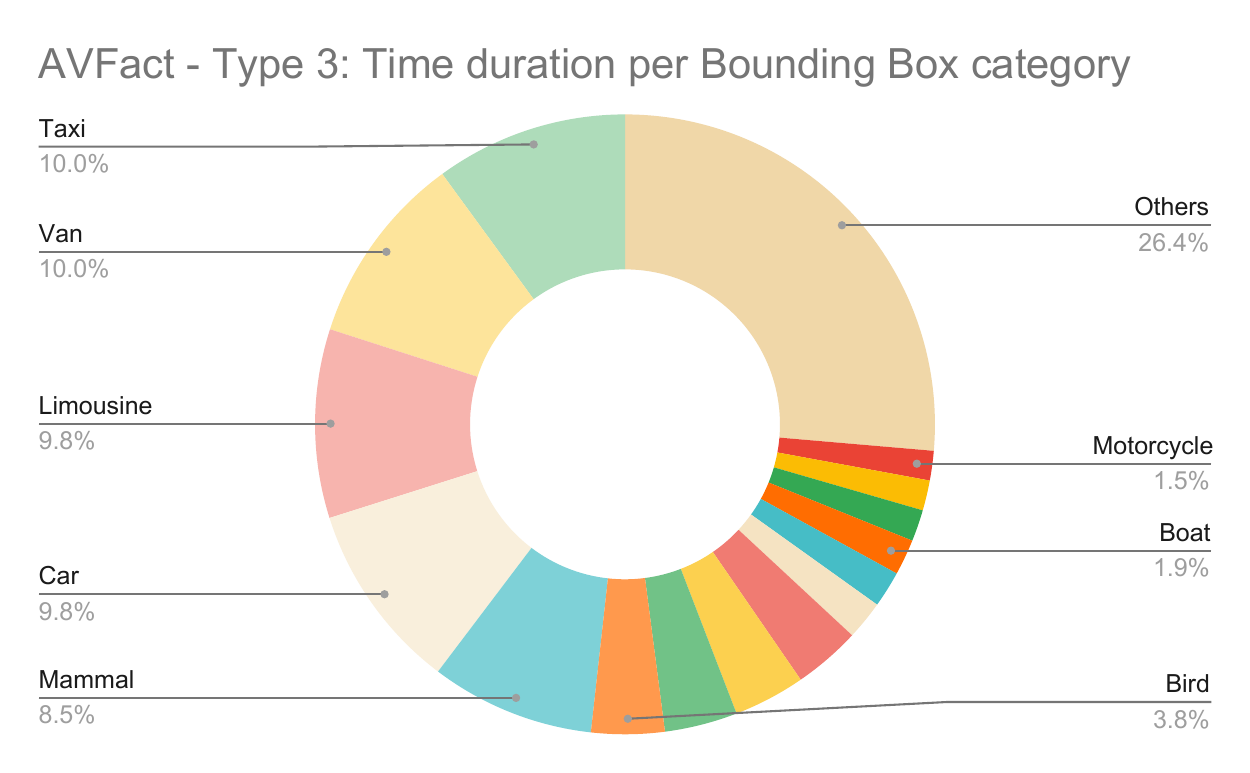}
    \caption{AVFact - Type 3: Total audio time duration per image category of chosen samples. }
    \label{fig:stat_type3}
  \end{subfigure}
  \hfill
  \begin{subfigure}{0.48\textwidth}
    \centering
    \includegraphics[width=\linewidth]{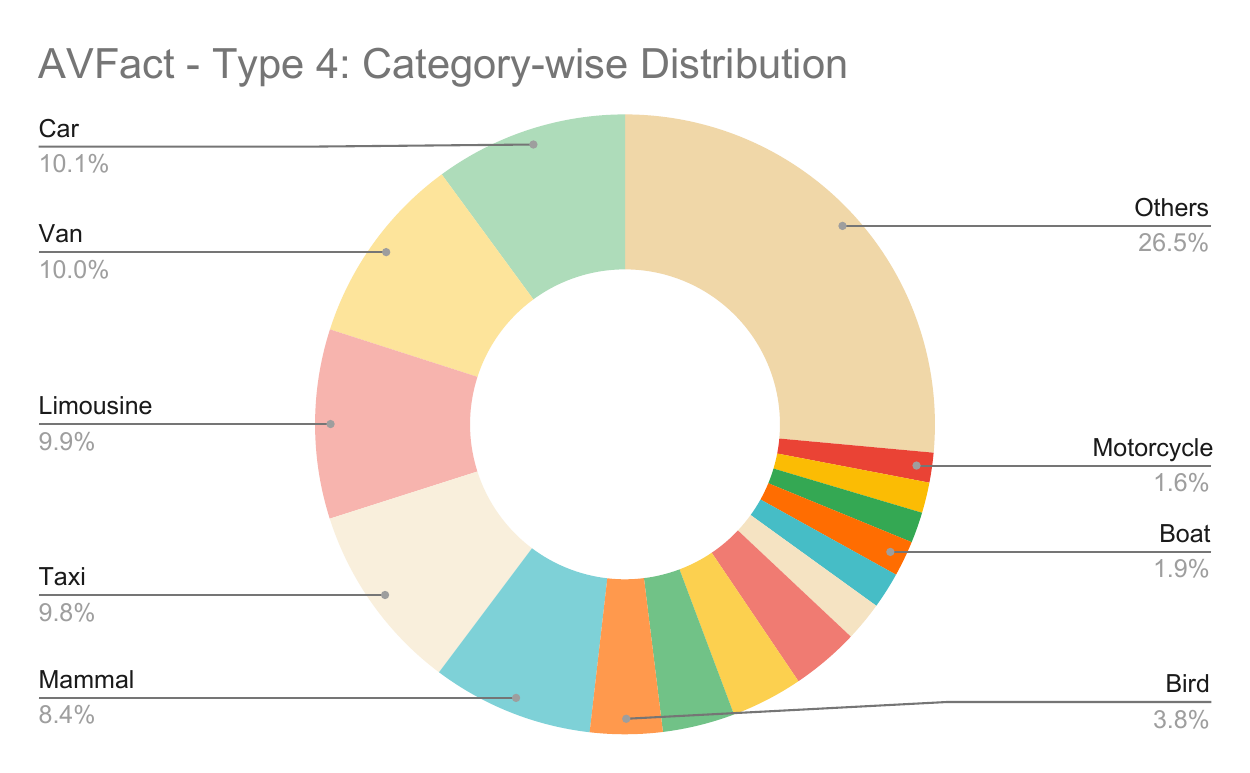}
    \caption{AVFact - Type 4: Category-wise distribution of samples from images Openimages dataset.}
    \label{fig:stat_type4}
  \end{subfigure}
  \caption{\textbf{\ourdataset statistics and analysis.}}
  \label{fig:main}
\end{figure}

% \usepackage{graphicx}
% \begin{table}[]
% \centering
% \resizebox{0.2\columnwidth}{!}{%
% \begin{tabular}{l|c}
% \toprule
% \multicolumn{1}{c|}{\textbf{Category}} & \textbf{Score} \\ \midrule
% Human face                             & 0.84           \\
% Bird                                   & 0.76           \\
% Girl                                   & 0.80           \\
% Boat                                   & 0.72           \\
% Woman                                  & 0.78           \\
% Car                                    & 0.77           \\
% Man                                    & 0.80           \\
% Person                                 & 0.80           \\
% Vehicle                                & 0.72           \\
% Others                                 & 0.74           \\ \bottomrule
% \end{tabular}%
% }
% \caption{Openimages - Category-wise Scores.}
% \label{tab:op_dataset_similarity_scores}
% \end{table}

\noindent{\textbf{Image Audio Similarity.}} To study the similarity between the image-audio pairs \cite{melfusion} from Openimages-AudioSet, we utilize the CLIP \cite{clip} and CLAP \cite{clap} scores by calculating $\mathcal{S}_{\text{CLIP}}\;\mathcal{S}_{\text{CLAP}}^{T}$, where $\mathcal{S} \in \mathbb{R}^{N \times N}$ and denotes the pairwise cross-modal similarity scores for a batch of size $N$. The CLIP similarity is calculated between the chosen image and the audio class label, similarly, the CLAP score is calculated between the audio class label and the audio snippet. The text modality acts as the bridging modality in this case. Fig. \ref{fig:image_audio_similarity_score} reports the image-audio similarity scores over the most frequent 9 categories while `others' denotes aggregation of all the remaining ones. Note the range of the scores is normalized between [0,1] with 0 being the lowest. The average score of image-audio pairs across all samples collected for the audio referred image grounding task is 0.77, supporting a strong association between the two modalities. 

\noindent{\textbf{Audio Duration.}} We report the category-wise mean duration (in sec.) of the audio samples from the AudioSet dataset in Fig. \ref{fig:audioset_cat_avg_duration} for image-guided audio temporal localization task. The `Train' class has the overall highest value with an average duration of 9.83 sec while the `Clicking' category has the lowest average duration at 0.32 sec.     

\noindent{\textbf{Class wise Robustness.}} We report class-wise (top 6 classes based on occurrence) \textit{True/False} sample count from the Openimages dataset for AVFact - Type 1 set in Fig. \ref{fig:stat_type1}. We maintain a good balance of matched and mismatched pairs to ensure our model is robust to deceptive queries.

\noindent{\textbf{AudioSet Distribution.}} Fig. \ref{fig:stat_type2} reports the class-wise distribution of samples present in the AVFact Type-2 set as collected from the AudioSet dataset.

\noindent{\textbf{Audio Duration Per Image Class.}} In Fig. \ref{fig:stat_type3} we present the duration of audio samples across various image classes from Openimages in AVFact Type-4 split. This demonstrates the overall balanced mix of image-audio distributions across different pairings.

\noindent{\textbf{Category Wise Distribution.}} Fig. \ref{fig:stat_type4} presents image category-wise distributions of samples from the Openimages dataset for AVFact Type-4 samples.

\begin{figure}[!t]
  \centering
  \begin{subfigure}{\textwidth}
    \centering
    \includegraphics[width=\columnwidth]{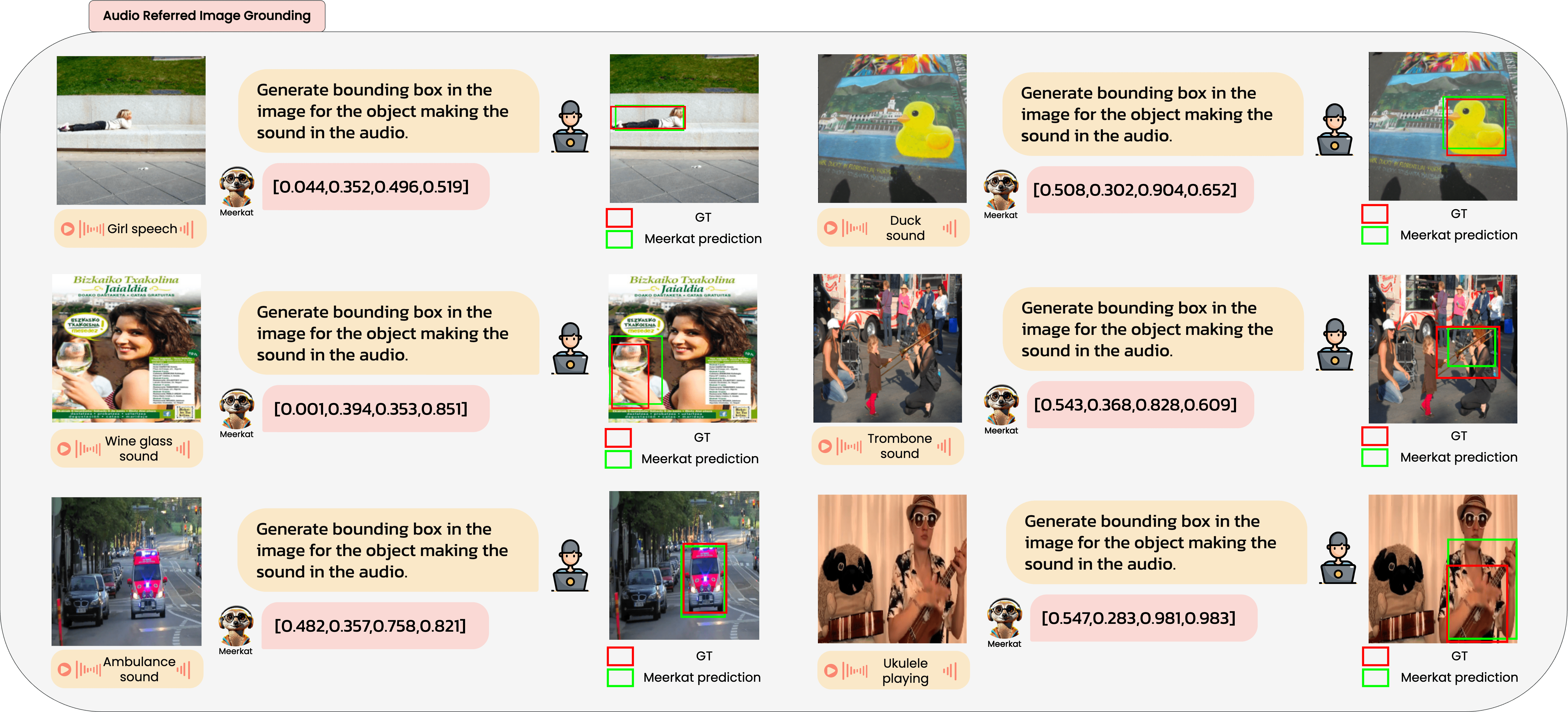}
    \caption{Qualitative samples on image grounding task.}
    \label{fig:qual_supp_spatial}
  \end{subfigure}
  \\
  \begin{subfigure}{\textwidth}
    \centering
    \includegraphics[width=\columnwidth]{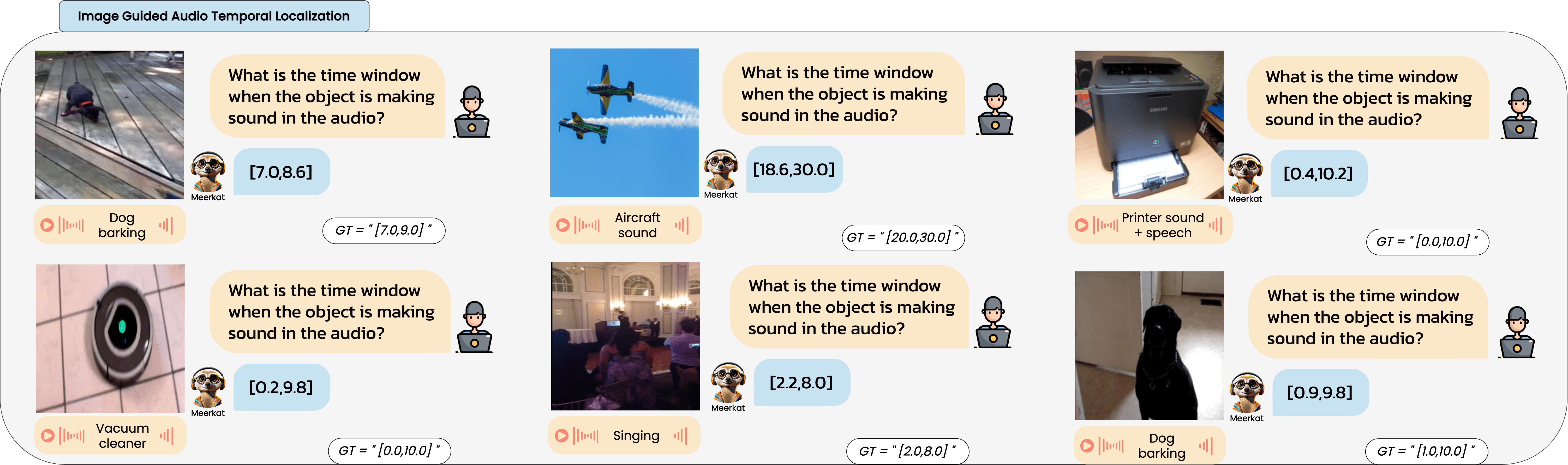}
    \caption{Qualitative results on audio temporal localization.}
    \label{fig:qual_supp_temporal}
  \end{subfigure}
  \\
  \begin{subfigure}{\textwidth}
    \centering
    \includegraphics[width=\columnwidth]{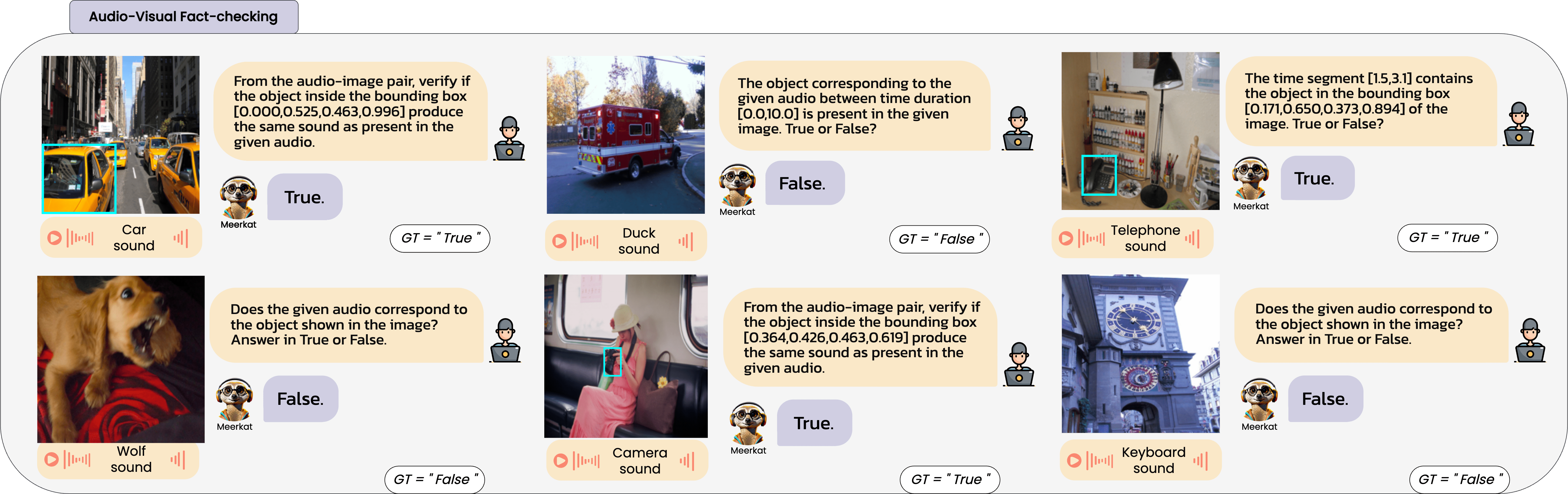}
    \caption{More results on AVFact task.}
    \label{fig:qual_supp_avfact}
  \end{subfigure}
  \caption{Qualitative results of \modelname on different fine-grained downstream tasks from the \taskunificationframework.}
  \label{fig:more_qualitative_results}
\end{figure}

\section{More Qualitative Results}
\label{qualitative results}
We provide additional qualitative results from \modelname in Fig. \ref{fig:more_qualitative_results}. In Fig. \ref{fig:qual_supp_spatial} we show excellent image grounding capabilities of our model when queried with audio inputs. We observe that even for small objects or visual scenes with complex associations among different components, \modelname can correctly identify the referred object. This underlines the fine-grained comprehension capabilities acquired by \modelname during its training phase. \modelname is equipped with strong audio temporal localization as well while prompted with an image. As evident from Fig. \ref{fig:qual_supp_temporal}, our model is capable of precisely understanding audio samples and accurately identifying the temporal onset of an event and the specific time duration of that particular event, even in the presence of other distractors and ambient sound. Fig. \ref{fig:qual_supp_avfact} depicts the fine-grained audio-visual comprehension capabilities of our method. Even when \modelname is presented with noisy audio-visual samples and scenarios that demand detailed AV association understanding, our model can produce correct results with substantially high accuracy. Our method is also adept at coarse-grained tasks like AVQA and AV captioning as demonstrated in Fig. \ref{fig:qual_supp_avqa} and \ref{fig:qual_supp_avc}.  

\begin{figure}[!t]
  \centering
  \begin{subfigure}{\textwidth}
    \centering
    \includegraphics[width=\columnwidth]{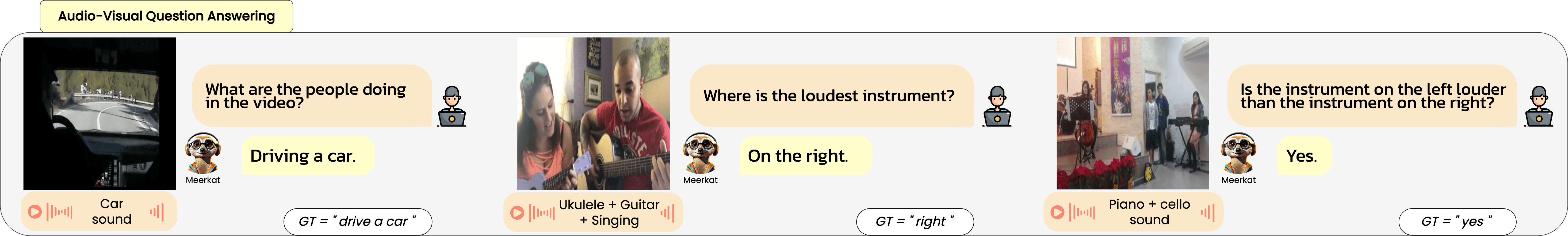}
    \caption{\modelname performance on AVQA task.}
    \label{fig:qual_supp_avqa}
  \end{subfigure}
  \\
  \begin{subfigure}{\textwidth}
    \centering
    \includegraphics[width=\columnwidth]{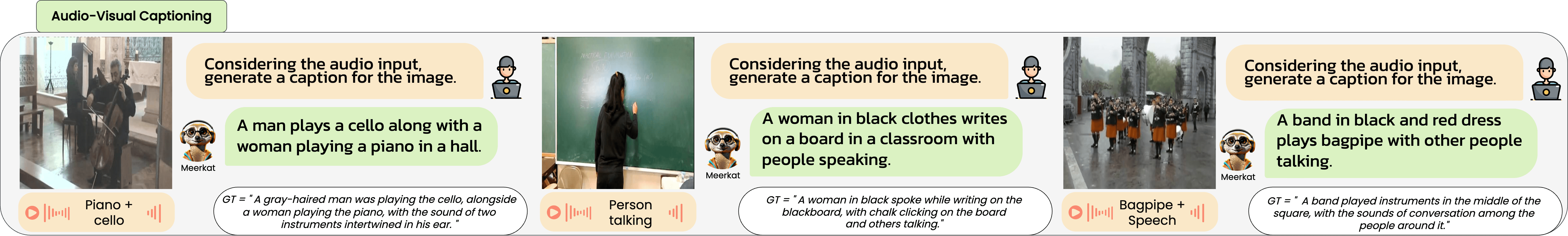}
    \caption{\modelname performance on AV Captioning task.}
    \label{fig:qual_supp_avc}
  \end{subfigure}
  \caption{Qualitative results of \modelname on different coarse-grained downstream tasks from the \taskunificationframework.}
  \label{fig:more_qualitative_results}
\end{figure}

\section{More ablations}
\label{more ablations}

% \subsection{With Different Image Encoder, Audio Encoder}

\subsection{Other Image Encoders} We compare the performance of our model on employing different image encoders as shown in Tab. \ref{tab:img_enc}. We observe the best performance with CLIP-ViT-B/16 \cite{clip} and use this as our preferred image encoder due to its compatibility with the instruction-guided image tokenizer module in our system. 

\begin{table}[h]
    \centering
    \resizebox{0.82\columnwidth}{!}{%
\begin{tabular}{c|c|c|c|c|c}
\toprule
\multirow{2}{*}{\textbf{Image Encoder}} &
  \multicolumn{1}{c|}{\textbf{VGG-SS}} &
  \multicolumn{1}{c|}{\textbf{LLP}} &
  \multicolumn{1}{c|}{\textbf{AVFact}} &
  \multicolumn{1}{c|}{\textbf{AVQA}} &
  \multicolumn{1}{c}{\textbf{VALOR}} \\
   &
  \multicolumn{1}{c|}{\textbf{cIOU} $\uparrow$}  &
  \multicolumn{1}{c|}{\textbf{F1-score} $\uparrow$}  &
  \multicolumn{1}{c|}{\textbf{Avg F1-score} $\uparrow$}  &
  \multicolumn{1}{c|}{\textbf{Avg Acc.} $\uparrow$}  &
  \multicolumn{1}{c}{\textbf{CIDEr} $\uparrow$}  \\ \hline
 
   CLIP-ViT-B/32 \cite{clip} & 42.56
   & 49.64
   & 0.78
   & 84.04
   & 73.39
   \\
   BLIP-ViT-B/16 \cite{blip} & 47.22
   & 52.83
   & 0.82
   & 85.79
   & 75.13
   \\
  \rowcolor{ThemeColor} \textbf{CLIP-ViT-B/16 (Ours)} &
  {\textbf{48.51}} &
  {\textbf{54.96}} &
  {\textbf{0.85}} &
  {\textbf{87.14}} & {\textbf{76.84}}
   \\ \bottomrule
\end{tabular}%
}
    \caption{\textbf{\modelname performance with different image encoders}}
    \label{tab:img_enc}
\end{table}

\subsection{Other Audio Encoders} 
We carry out experiments with various audio encoders in Tab. \ref{tab:aud_enc} such as Open L3 \cite{cramer2019look, arandjelovic2017look}, WAV2CLIP \cite{wu2022wav2clip}, and Wav2Vec2 \cite{wav2vec2} with the optimal performance obtained with the CLAP \cite{clap} encoder. We attribute this performance boost to its superior Swin Transformer \cite{liu2021swin} based backbone to get audio features from a log-Mel spectrogram. Owing to its large-scale contrastive language-audio pretraining to develop an audio representation by combining audio data with natural language descriptions, CLAP encoders are shown to perform exceptionally well on processing open-domain audio over speech-based encoders like Whisper \cite{whisper}.

\begin{table}[h]
    \centering
    \resizebox{0.82\columnwidth}{!}{%
\begin{tabular}{c|c|c|c|c|c}
\toprule
\multirow{2}{*}{\textbf{Audio Encoder}} &
  \multicolumn{1}{c|}{\textbf{VGG-SS}} &
  \multicolumn{1}{c|}{\textbf{LLP}} &
  \multicolumn{1}{c|}{\textbf{AVFact}} &
  \multicolumn{1}{c|}{\textbf{AVQA}} &
  \multicolumn{1}{c}{\textbf{VALOR}} \\
   &
  \multicolumn{1}{c|}{\textbf{cIOU} $\uparrow$}  &
  \multicolumn{1}{c|}{\textbf{F1-score} $\uparrow$}  &
  \multicolumn{1}{c|}{\textbf{Avg F1-score} $\uparrow$}  &
  \multicolumn{1}{c|}{\textbf{Avg Acc.} $\uparrow$}  &
  \multicolumn{1}{c}{\textbf{CIDEr} $\uparrow$}  \\ \hline
 
   Open L3 \cite{cramer2019look, arandjelovic2017look} & 44.52
   & 51.28
   & 0.76
   & 83.29
   & 72.38
   \\
   WAV2CLIP \cite{wu2022wav2clip} & 45.34
   & 51.94
   & 0.78
   & 84.46
   & 73.77
   \\
   Wav2Vec2 \cite{wav2vec2} & 46.91
   & 53.07
   & 0.81
   & 85.88
   & 75.80
   \\
  \rowcolor{ThemeColor} \textbf{CLAP audio encoder} &
  {\textbf{48.51}} &
  {\textbf{54.96}} &
  {\textbf{0.85}} &
  {\textbf{87.14}} & {\textbf{76.84}}
   \\ \bottomrule
\end{tabular}%
}
    \caption{\textbf{\modelname performance with different audio encoders}}
    \label{tab:aud_enc}
\end{table}

% \vspace{-0.6in}

\subsection{With Different LLM}

We ablate our model and replace the LLM with other recent language models such as T5 \cite{T5}, Vicuna \cite{vicuna}, and Alpaca \cite{alpaca}. We observe a noticeable drop in performance when the LLM is not instruction-tuned compared to its instruction-tuned counterpart. This demonstrates the importance of leveraging instruction-tuned LLMs under a multi-modal instruction comprehension setup. We note instruction tuning allows equipping the LLM with a customized instructions template which results in improved performance under a multi-task setting, as demonstrated in Tab. \ref{tab:llm_ablation}.

\begin{table}[]
    \centering
    \resizebox{0.82\columnwidth}{!}{%
\begin{tabular}{c|c|c|c|c|c}
\toprule
\multirow{2}{*}{\textbf{Model}} &
  \multicolumn{1}{c|}{\textbf{VGG-SS}} &
  \multicolumn{1}{c|}{\textbf{LLP}} &
  \multicolumn{1}{c|}{\textbf{AVFact}} &
  \multicolumn{1}{c|}{\textbf{AVQA}} &
  \multicolumn{1}{c}{\textbf{VALOR}} \\
   &
  \multicolumn{1}{c|}{\textbf{cIOU} $\uparrow$}  &
  \multicolumn{1}{c|}{\textbf{F1-score} $\uparrow$}  &
  \multicolumn{1}{c|}{\textbf{Avg F1-score} $\uparrow$}  &
  \multicolumn{1}{c|}{\textbf{Avg Acc.} $\uparrow$}  &
  \multicolumn{1}{c}{\textbf{CIDEr} $\uparrow$}  \\ \midrule
 
   T5 & 41.49
   & 48.50
   & 0.78
   & 82.49
   & 72.56
   \\
   Alpaca & 42.74
   & 49.98
   & 0.80
   & 83.75
   & 74.84
   \\
   Vicuna &  47.06
   &  53.68
   & 0.83
   &  86.38
   &  75.88
   \\
  \rowcolor{ThemeColor} \textbf{Llama-2} &
  \cellcolor[HTML]{FAE8C8}{\textbf{48.51}} &
  \cellcolor[HTML]{FAE8C8}{\textbf{54.96}} &
  \cellcolor[HTML]{FAE8C8}{\textbf{0.85}} &
  \cellcolor[HTML]{FAE8C8}{\textbf{87.14}} & \cellcolor[HTML]{FAE8C8}{\textbf{76.84}}
   \\ \bottomrule
\end{tabular}%
}
    \caption{\textbf{Ablative study under various LLMs.} }
    \label{tab:llm_ablation}
\end{table}

\subsection{Effect of $\mathcal{\lambda}_{OT}$ and $\mathcal{\lambda}_{AC}$}

We ablate loss hyperparameters $\lambda_{\text{OT}}$ and $\lambda_{\text{AC}}$ and compare performance of \modelname on ARIG and IGATL tasks in Fig. \ref{fig:plot_1} and Fig. \ref{fig:plot_2}, respectively. Experimental results suggest that best metrics are obtained with $\lambda_{\text{AC}}$ = 0.35 and $\lambda_{\text{OT}}$ = 0.75, respectively. 
% underlying optimal weighting of the two modules is key in obtaining the best results.  

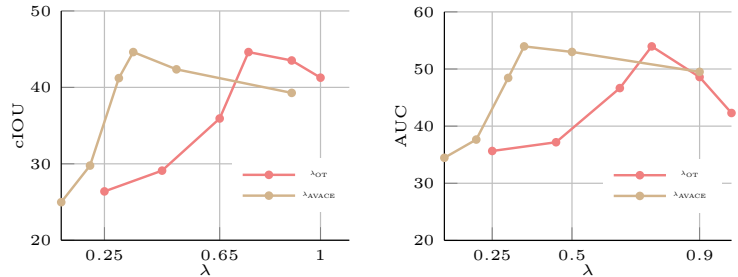
\begin{figure}[]
\vspace{-5mm}
	\centering
    \hspace{-0.55em}
	\begin{subfigure}[b]{0.4\linewidth}
		\resizebox{1.0\textwidth}{!}{
			\begin{tikzpicture}
            \begin{axis}[
                axis x line*=bottom,
    		  axis y line*=left,
                ymin=20, ymax=50,
        	xmin=0.1, xmax=1.1,
                xticklabel={\pgfmathparse{\tick}\pgfmathprintnumber{\pgfmathresult}},    
                xtick={0.25, 0.65, 1},
                xtick distance=0.2, % Equally spaced tick labels
                xlabel={\tiny{$\lambda$}},
                ylabel={\tiny{cIOU}},
                legend style={at={(0.8,0.36)},anchor=north,font=\tiny,draw=none},
                xlabel shift=-5pt,
                ylabel shift=-4pt,
                xticklabel style={font=\tiny},
                yticklabel style={font=\tiny},
		      % label style={font=\tiny},
                width=\linewidth,
                grid=major,
                ]
                \addplot[LinePlot1,mark=*,mark options={scale=0.5},style={thick}] coordinates {
                    (0.25, 26.39)
                    (0.45, 29.11)
                    (0.65, 35.93)
                    (0.75, 44.62)
                    (0.9, 43.52)
                    (1, 41.27)
                };
                \addlegendentry{\scalebox{.4}{$\lambda_{\text{OT}}$}}

                \addplot[LinePlot2,mark=*,mark options={scale=0.5},style={thick}] coordinates {
                    (0.1, 24.98)
                    (0.2, 29.77)
                    (0.3, 41.21)
                    (0.35, 44.62)
                    (0.5, 42.36)
                    (0.9, 39.28)
                };
                \addlegendentry{\scalebox{.4}{$\lambda_{\text{AVACE}}$}}
            \end{axis}
        \end{tikzpicture}%
		}
		\caption{\textbf{cIoU upper bound on VGG-SS} with varying weightage of $\lambda_{\text{OT}}$ and $\lambda_{\text{AC}}$.}
        \label{fig:plot_1}
	\end{subfigure}
     \hspace{-0em}
	\begin{subfigure}[b]{0.4\linewidth}
		\resizebox{1.0\textwidth}{!}{
			\begin{tikzpicture}
            \begin{axis}[
                axis x line*=bottom,
    		  axis y line*=left,
                ymin=20, ymax=60,
        	xmin=0.1, xmax=1.0,
                xticklabel={\pgfmathparse{\tick}\pgfmathprintnumber{\pgfmathresult}},    
                xtick={0.25, 0.5, 0.9},
                xtick distance=0.2, % Equally spaced tick labels
                xlabel={\tiny{$\lambda$}},
                ylabel={\tiny{AUC}},
                legend style={at={(0.76,0.36)},anchor=north,font=\tiny,draw=none},
                xlabel shift=-5pt,
                ylabel shift=-4pt,
                xticklabel style={font=\tiny},
                yticklabel style={font=\tiny},
		      label style={font=\tiny},
                width=\linewidth,
                grid=major,
                ]
                \addplot[LinePlot1,mark=*,mark options={scale=0.5},style={thick}] coordinates {
                    (0.25, 35.65)
                    (0.45, 37.18)
                    (0.65, 46.66)
                    (0.75, 53.96)
                    (0.9, 48.59)
                    (1, 42.3)
                };
                \addlegendentry{\scalebox{0.4}{$\lambda_{\text{OT}}$}}

                \addplot[LinePlot2,mark=*,mark options={scale=0.5},style={thick}] coordinates {
                    (0.1, 34.46)
                    (0.2, 37.65)
                    (0.3, 48.44)
                    (0.35, 53.96)
                    (0.5, 52.99)
                    (0.9,49.5)
                };
                \addlegendentry{\scalebox{0.4}{$\lambda_{\text{AVACE}}$}}
            \end{axis}
        \end{tikzpicture}
        
        %
		}
		\caption{\textbf{AUC upper bound on LLP} with varying weightage of $\lambda_{\text{OT}}$ and $\lambda_{\text{AC}}$.}
        \label{fig:plot_2}
	\end{subfigure}
 \hspace{-0em}
	% \begin{subfigure}[b]{0.32\linewidth}
	% 	\resizebox{1.0\textwidth}{!}{
	% 		\input{graphs/line_plot_3}%
	% 	}
	% 	\caption{\textbf{cIoU upper bound on VGG-SS} for Full vs. LoRA based finetuning.}
 %        \label{fig:plot_3}
	% \end{subfigure}
	\caption{\textbf{Ablative experiments} on (\textbf{a}) \textbf{spatial}  and (\textbf{b}) \textbf{temporal} localization tasks. In \textbf{(a)} and \textbf{(b)} we keep $\lambda_{\text{AC}}$ and $\lambda_{\text{OT}}$ fixed at 0.35 and 0.75 respectively while varying the other $\lambda$.}
\vspace{-5mm}
\end{figure}

% \section{Empirical results with \strongalignmentmodule improving performance over \weakalignmentmodule}

\section{Comparison with Contrastive Loss}
\label{comparison with contrastive loss}

We compare the optimal transport \cite{deepemd} based objective ($\mathcal{L}_{\text{OT}}$) with the contrastive loss-based approach \cite{nce, infonce, clip} to facilitate weak alignment in \modelname. Contrastive approaches operate on the level of global features and therefore only capture class-level information. Although such an alignment strategy may be beneficial in coarse-grained tasks, they are not suitable for tasks which require fine-grained understanding. Conversely, as employed in \weakalignmentmodule, OT-based alignment operates on the level of patches in a weakly-supervised manner. Such a form of guidance is interpretable since a transport plan is optimized which dictates the relationships between the cross-modal patch embeddings, and therefore, is more suitable for fine-grained downstream tasks. Even though OT-based alignment strategies have been employed earlier for word-region level alignment \cite{graphoptimaltransport,volta,uniter}, we are the first to introduce it under audio-visual setting. We empirically find that using $\mathcal{L}_{\text{OT}}$ is superior in all the downstream tasks (refer to Tab. \ref{tab:ablation_contrastive}). Note that in both cases $\mathcal{L}_{\text{AC}}$ is employed to add strong supervision. Based on our results, we hypothesize that initial patch-level alignment with \weakalignmentmodule\ yields high-quality representations which substantially assist \strongalignmentmodule\ to attend to the regions of interest, thereby improving localization performance, as opposed to using contrastive loss with \strongalignmentmodule.

% \textcolor{red}{comparison against nce, infonce}

\begin{table}[]
    \centering
    \resizebox{0.82\columnwidth}{!}{%
\begin{tabular}{c|c|c|c|c|c}
\toprule
\multirow{2}{*}{\textbf{Loss}} &
  \multicolumn{1}{c|}{\textbf{VGG-SS}} &
  \multicolumn{1}{c|}{\textbf{LLP}} &
  \multicolumn{1}{c|}{\textbf{AVFact}} &
  \multicolumn{1}{c|}{\textbf{AVQA}} &
  \multicolumn{1}{c}{\textbf{VALOR}} \\
   &
  \multicolumn{1}{c|}{\textbf{cIOU} $\uparrow$}  &
  \multicolumn{1}{c|}{\textbf{F1-score} $\uparrow$}  &
  \multicolumn{1}{c|}{\textbf{Avg F1-score} $\uparrow$}  &
  \multicolumn{1}{c|}{\textbf{Avg Acc.} $\uparrow$}  &
  \multicolumn{1}{c}{\textbf{CIDEr} $\uparrow$}  \\ \hline
 
   $\mathcal{L}_{\text{Contrastive}}$ &  46.95
   & 52.28
   &  0.81
   &  86.31
   &  74.98
   \\
  \rowcolor{ThemeColor} \textbf{$\mathcal{L}_{\text{OT}}$} &
  \cellcolor[HTML]{FAE8C8}{\textbf{48.51}} &
  \cellcolor[HTML]{FAE8C8}{\textbf{54.96}} &
  \cellcolor[HTML]{FAE8C8}{\textbf{0.85}} &
  \cellcolor[HTML]{FAE8C8}{\textbf{87.14}} & \cellcolor[HTML]{FAE8C8}{\textbf{76.84}}
   \\ \bottomrule
\end{tabular}%
}
    \caption{\textbf{Comparison against contrastive loss based weak alignment strategy}}
    \label{tab:ablation_contrastive}
\end{table}

\section{Comparison with Two-stage Training}
\label{comparison with two stage training}

We systematically study the effect of the two-stage vs. single-stage training paradigm. Inspired by recent works \cite{fiber, glip} on fine-grained image understanding tasks, we design a two-stage experimental set-up. In stage I, we perform modality alignment among the image and audio encoders through weak supervision, by employing \weakalignmentmodule module. We do not use LLM in this stage I and therefore the only objective we optimize is $\mathcal{L}_{\text{OT}}$. Stage I training is followed by stage II training involving the \strongalignmentmodule module to provide strong supervision. In stage II, we fine-tune LLM using LoRA. Experimental results show comparable performance in both cases as depicted in Tab. \ref{tab:ablation_stage}. We opt for single-stage training because not only it is superior (in terms of performance, see Tab. \ref{tab:ablation_stage}), but it is also computationally efficient and less resource intensive.

\begin{table}[]
    \centering
    \resizebox{0.82\columnwidth}{!}{%
\begin{tabular}{c|c|c|c|c|c}
\toprule
\multirow{2}{*}{\textbf{Model}} &
  \multicolumn{1}{c|}{\textbf{VGG-SS}} &
  \multicolumn{1}{c|}{\textbf{LLP}} &
  \multicolumn{1}{c|}{\textbf{AVFact}} &
  \multicolumn{1}{c|}{\textbf{AVQA}} &
  \multicolumn{1}{c}{\textbf{VALOR}} \\
   &
  \multicolumn{1}{c|}{\textbf{cIOU} $\uparrow$}  &
  \multicolumn{1}{c|}{\textbf{F1-score} $\uparrow$}  &
  \multicolumn{1}{c|}{\textbf{Avg F1-score} $\uparrow$}  &
  \multicolumn{1}{c|}{\textbf{Avg Acc.} $\uparrow$}  &
  \multicolumn{1}{c}{\textbf{CIDEr} $\uparrow$}  \\ \hline
 
   Two-stage & 48.43
   & 54.81
   & 0.85
   & 87.11
   & 76.59
   \\
  \rowcolor{ThemeColor} \textbf{Single-stage} &
  \cellcolor[HTML]{FAE8C8}{\textbf{48.51}} &
  \cellcolor[HTML]{FAE8C8}{\textbf{54.96}} &
  \cellcolor[HTML]{FAE8C8}{\textbf{0.85}} &
  \cellcolor[HTML]{FAE8C8}{\textbf{87.14}} & \cellcolor[HTML]{FAE8C8}{\textbf{76.84}}
   \\ \bottomrule
\end{tabular}%
}
    \caption{\textbf{Comparison against two stage training}}
    \label{tab:ablation_stage}
\end{table}

% \section{Contrastive Based Weak Alignment vs OT Based Weak Alignment }

% \textcolor{red}{qual samples for each task}

% \section{Dataset Curation Details}

\section{Role of Audio in AVQA Task}
\label{role of audio}

To study the role of the audio modality and how effectively our model can encode audio information, we perform an ablation study by removing the audio information altogether and performing visual-only question answering.  We note the performance of our method drops significantly when only the visual modality is used to answer the same set of questions underlying the role of the audio modality. Tab. \ref{tab:with_wo_audio} demonstrates the quantitative results.

\begin{table}[]
    \centering
    \resizebox{0.82\columnwidth}{!}{%
\begin{tabular}{c|c|c|c|c|c|c}
\toprule
\multicolumn{1}{c|}{\textbf{Model}} &
  \multicolumn{1}{c|}{\textbf{Exist} $\uparrow$}  &
  \multicolumn{1}{c|}{\textbf{Localis} $\uparrow$}  &
  \multicolumn{1}{c|}{\textbf{Count} $\uparrow$}  &
  \multicolumn{1}{c|}{\textbf{World K} $\uparrow$}  &
  \multicolumn{1}{c|}{\textbf{Temp} $\uparrow$}  &
  \multicolumn{1}{c}{\textbf{Avg} $\uparrow$}  \\ \midrule
 
   Without audio & 83.62
   & 79.28
   & 80.46
   & 78.49
   & 69.26
   & 78.22
   \\
  \rowcolor{ThemeColor} \textbf{With audio} &
  \cellcolor[HTML]{FAE8C8}{\textbf{88.24}} &
  \cellcolor[HTML]{FAE8C8}{\textbf{86.65}} &
  \cellcolor[HTML]{FAE8C8}{\textbf{84.60}} &
  \cellcolor[HTML]{FAE8C8}{\textbf{87.05}} & \cellcolor[HTML]{FAE8C8}{\textbf{86.55}}
  & \textbf{86.61}
   \\ \bottomrule
\end{tabular}%
}
    \caption{\textbf{Role of audio modality.} Quantitative results on AVQA dataset when the model is presented with and without audio.}
    \label{tab:with_wo_audio}
\end{table}

\section{More on Optimal Transport}
\label{optimal transport}

Our \weakalignmentmodule is responsible for cross-modal alignment of image and audio feature representations in a weakly-supervised manner. This is enabled by minimizing the Wasserstein distance ($\mathcal{D}_{\text{Wasserstein}}$) between the image and audio (spectrogram) patches and subsequently learning an optimal transport plan $\mathbf{\Omega}$. The detailed steps of Optimal Transport-based Wasserstein Distance ($\mathcal{L}_{\text{Wasserstein}}$) computation are outlined in Algorithm \ref{algo:wass_dist}.

\begin{algorithm}[]
\small
\caption{\modelname: Wasserstein Distance Computation in \weakalignmentmodule}
\label{algo:wass_dist}
% \begin{abox}
\begin{algorithmic}[1]
\Require{Images: $\{I_i\}_{i=1}^{k}$; Audios: $\{A_j\}_{j=1}^{k}$; Total Optimal Transport steps: $\mathcal{S}_{\mathbf{\Omega}}$; Initial scaled unity matrix: $\boldsymbol{\sigma} = \frac{1}{k}\mathbf{1_k}$; Initial Transport Plan: $\mathbf{\Omega}^{(1)} = \mathbf{1} \mathbf{1}^\top$; Cosine similarity matrix: $\mathbf{C}_{ij} = c(I_i, A_j)$; similarity matrix decay factor: $\beta$; Scaled similarity matrix: $\mathbf{\Upsilon}_{ij} = {\rm e}^{-\frac{\mathbf{C}_{ij}}{\beta}}$.}
\Ensure{Learned Optimal Transport Plan: $\mathbf{\Omega}$; Wasserstein Distance: $\mathcal{D}_{\text{Wasserstein}}$.}
\For{$t \in \{1,2,3, \cdots \mathcal{S}_{\mathbf{\Omega}}\}$}
\State $\mathbf{Q} \gets \mathbf{\Upsilon} \odot \mathbf{\Omega}^{(t)}$ \Comment{$\odot$ is Hadamard product}
    \For{$l \in \{1,2,3,\cdots, L\}$}
        \State $\boldsymbol{\delta} \gets \frac{1}{k\mathbf{Q}{\boldsymbol{\sigma}}}$, $\boldsymbol{\sigma} \gets \frac{1}{k\mathbf{Q}^\top\boldsymbol{\delta}}$
    \EndFor
    \State $\mathbf{\Omega}^{(t+1)} \gets \text{diag}(\boldsymbol{\delta})\mathbf{Q}\text{diag}(\boldsymbol{\sigma})$
\EndFor
\State $\mathcal{D}_{\text{Wasserstein}} \gets \langle \mathbf{C}^{\top}, \mathbf{\Omega}\rangle$ \Comment{$\langle \cdot, \cdot \rangle$ is the Frobenius dot-product} \\
\Return $\mathbf{\Omega}$, $\mathcal{D}_{\text{Wasserstein}}$
\end{algorithmic}
% \end{abox}
\end{algorithm}

\section{AVSBench Data Collection}
\label{avsbench details}

Given the segmentation mask of an object, we consider the top-most, left-most, bottom-most and right-most points and draw horizontal and vertical projection lines as shown in Fig. \ref{fig:seg_to_bb}. These lines intersect each other at four points which when connected gives us the desired bounding box that completely encloses the object of interest. For each such bounding box we consider the coordinates $(x_{\text{Left}},y_{\text{Top}})$ and $(x_{\text{Right}},y_{\text{Bottom}})$ as GT labels, as shown in Fig. \ref{fig:seg_to_bb}.
\begin{figure}[]
    \centering
    \includegraphics[width=0.8\columnwidth]{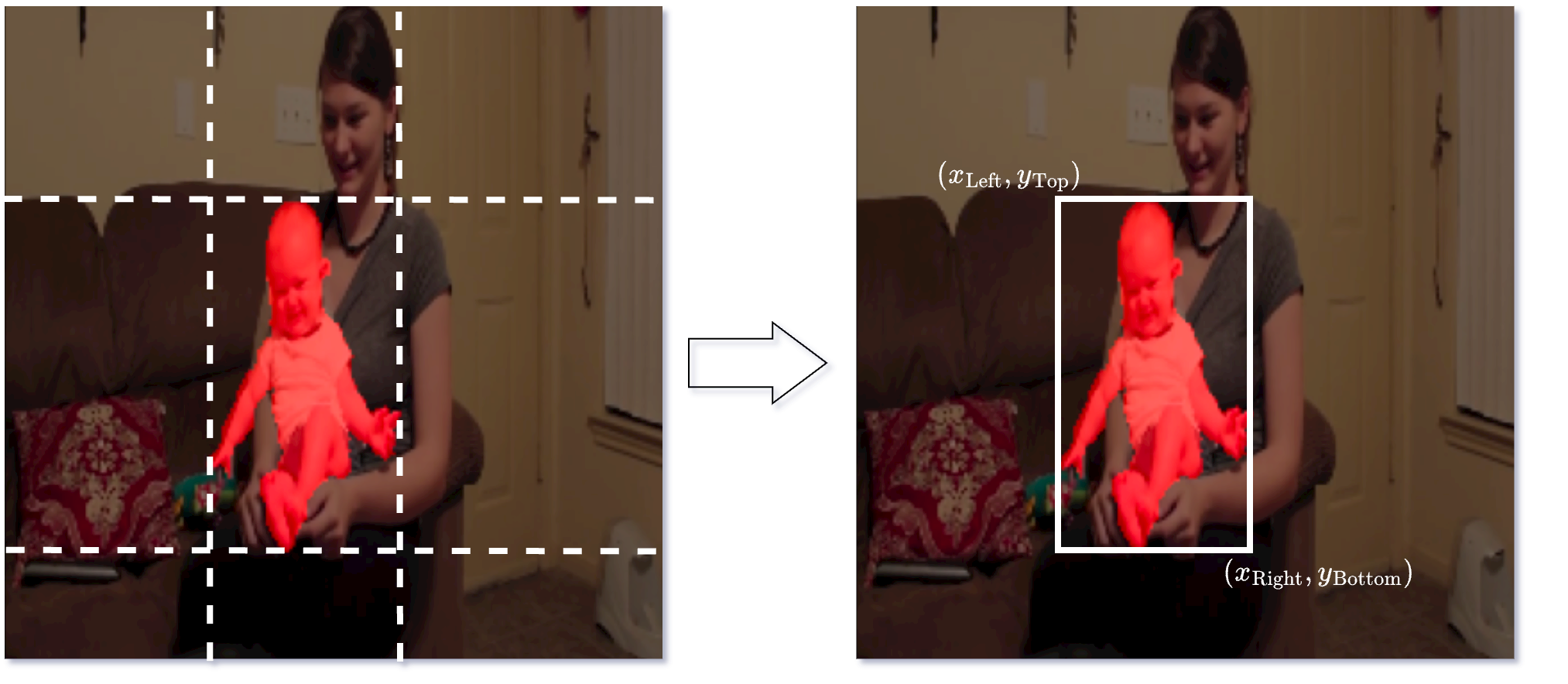}
    \caption{\textbf{Transformation from segmentation mask to bounding box.}}
    \label{fig:seg_to_bb}
\end{figure}

% \section{Details on openimages-audioset, openimages-vggsound, pascal sound preparation}

% \textcolor{red}{add lookup table between openimages audioset / openimages vggsound}

% \section{Other Image Datasets}
% \label{other image datasets}
% Why We Don't Choose COCO, LVIS Datasets

\section{Comparison against ImageBind}
\label{comparison against imagebind}

% car horn example

% \textcolor{red}{we may not add this car horn example as it might create more confusion. This can't be a claim of our paper because we are not doing proper study. just adding empirical results would be a good idea.}

We employ modality-specific encoders from ImageBind \cite{imagebind} and compare them with CLIP-CLAP combination as used in \modelname (Tab. \ref{tab:imagebind_vs_meerkat}). Empirical results suggest that our encoder combination performs slightly superior compared to ImageBind. A more theoretical insight in this regard can be considered as a future work. However, this is beyond the scope of the current study.

\begin{table}[]
    \centering
    \resizebox{0.82\columnwidth}{!}{%
\begin{tabular}{c|c|c|c|c|c}
\toprule
\multirow{2}{*}{\textbf{Image and Audio Encoders}} &
  \multicolumn{1}{c|}{\textbf{VGG-SS}} &
  \multicolumn{1}{c|}{\textbf{LLP}} &
  \multicolumn{1}{c|}{\textbf{AVFact}} &
  \multicolumn{1}{c|}{\textbf{AVQA}} &
  \multicolumn{1}{c}{\textbf{VALOR}} \\
   &
  \multicolumn{1}{c|}{\textbf{cIOU} $\uparrow$}  &
  \multicolumn{1}{c|}{\textbf{F1-score} $\uparrow$}  &
  \multicolumn{1}{c|}{\textbf{Avg F1-score} $\uparrow$}  &
  \multicolumn{1}{c|}{\textbf{Avg Acc.} $\uparrow$}  &
  \multicolumn{1}{c}{\textbf{CIDEr} $\uparrow$}  \\ \hline
 
   ImageBind Encoders & 47.71
   & 54.03
   & 0.84   & 86.30
   & 75.58
   \\
  \textbf{CLIP-CLAP (ours)} &
  \cellcolor[HTML]{FAE8C8}{\textbf{48.51}} &
  \cellcolor[HTML]{FAE8C8}{\textbf{54.96}} &
  \cellcolor[HTML]{FAE8C8}{\textbf{0.85}} &
  \cellcolor[HTML]{FAE8C8}{\textbf{87.14}} & \cellcolor[HTML]{FAE8C8}{\textbf{76.84}}
   \\ \bottomrule
\end{tabular}%
}
    \caption{\textbf{Comparison against ImageBind.}}
    \label{tab:imagebind_vs_meerkat}
\end{table}

\section{Other Quantitative Metrics on AVQA task}
\label{other avqa metrics}
We evaluate the performance of our method on two additional metrics from the AVQA task namely \textit{Counting (Count)} and \textit{Comparative (Comp)} and report the performance in Tab. \ref{tab:avqa_count_comp}. These two metrics along with the three other metrics (\textit{Existential}, \textit{Localization}, and \textit{Temporal} reported in the main paper) complete the evaluation suite for the AVQA and MUSIC-AVQA tasks. We observe an overall steady performance of our method across these categorizations, by virtue of the excellent generalizability of \modelname to coarse-grained tasks.

% \textcolor{red}{table}
\begin{table}[]
\centering
\renewcommand{\arraystretch}{0.92}
\resizebox{0.82\columnwidth}{!}{
\begin{tabular}{lc|cc|cc}
\toprule
\multirow{2}{*}{\textbf{Model}} &  \multirow{2}{*}{\textbf{Generalist?}} & \multicolumn{2}{c|}{\textbf{AVQA}} & \multicolumn{2}{c}{\textbf{MUSIC-AVQA}} 
% & \multicolumn{4}{c}{\textbf{VALOR-32K}} 
\\
&  
& 
% \textbf{Exist $\uparrow$} &
% \textbf{Localis $\uparrow$} &
\textbf{Count $\uparrow$} &
\textbf{World K $\uparrow$} &
% \textbf{Temp $\uparrow$} &
% \textbf{Avg $\uparrow$} &
% \textbf{Exist $\uparrow$} &
% \textbf{Localis $\uparrow$} &
\textbf{Count $\uparrow$} &
\textbf{Comp $\uparrow$} 
% &
% \textbf{Temp $\uparrow$} % &
% \textbf{BLEU@4 $\uparrow$} &
% \textbf{METEOR $\uparrow$} &
% \textbf{ROUGE $\uparrow$} &
% \textbf{CIDEr $\uparrow$} 
\\ 
\midrule
AVSD \cite{schwartz2019simple}      & \textcolor{OrangeRed}{\ding{55}} & 
% 81.61 & 
% 58.79 & 
 63.89 &
 61.52 &
% 61.41 & 
%  65.49 &
-     &  
-      
% &  
% -     &
% -     &
% -
% & - & - & - & - 
\\
PanoAVQA \cite{yun2021pano} & \textcolor{OrangeRed}{\ding{55}} & 
% 81.21 & 
% 59.33 & 
64.91 & 
64.22 
& 
% 63.23 & 
% 66.64  &  
-       & 
-      
% &  
% -      & 
% -     & 
% -  
% & - & - & - & - 
\\
ST-AVQA \cite{musicavqadataset}   & \textcolor{OrangeRed}{\ding{55}} & 
% 81.81 & 
% 64.51 & 
70.80 & 
66.01 & 
% 63.23 & 
% 69.54  &   
-     &  
-      
% &  
% -      &  
% -     &  
% -
% & - & - & - & - 
\\
CAD \cite{nadeem2024cad}                        & \textcolor{OrangeRed}{\ding{55}} & 
% 83.42 & 
% 73.97 & 
76.37 & 
74.88 & 
% 76.16 & 
% 76.96  &   
-     &  
-      
% & 
% -       & 
% -      &  
% -    
% & - & - & - & - 
\\
AVST \cite{musicavqadataset}               & \textcolor{OrangeRed}{\ding{55}} & -      & -      
% &  -      &  -      &  -     &  -     
& 
% 72.44 & 
% 65.54   & 
68.22 & 
63.31  
% & 
% 59.36 
% & - & - & - & - 
\\
LAVISH \cite{lin2023vision}             & \textcolor{OrangeRed}{\ding{55}} & -      &  -     
% &  -     &  -     &  -     &  -     
& 
% 73.83  & 
% 65.00  & 
73.28 & 
63.49  
% & 
% 60.81 
% & - & - & - & - 
\\
LAST \cite{liu2024tackling}                       & \textcolor{OrangeRed}{\ding{55}} & -      &  -     
% &  -      &  -     &  -     &   -    
&
% 76.21  & 
% 68.91  & 
75.23 & 
65.60 
% & 
% 60.60 
% & - & - & - & - 
\\
% SMPFF \cite{chen2021mm21}  & \textcolor{OrangeRed}{\ding{55}} & - & - & - & - & - & - & 7.59                      & 12.64                     & 28.69                     & 37.18                     \\
% VALOR \cite{valor}                                   & \textcolor{OrangeRed}{\ding{55}} & - & - & - & - & - & -& 8.97                      & 14.88                     & 30.86                     & 55.73                     \\
% [0.5ex]\cdashline{1-12}\\[-2.0ex]
% \customdottedline
% \vspace{-0.5mm}
% \customdottedlineavqa
\midrule
Macaw-LLM \cite{macawllm}                   & \textcolor{ForestGreen}{\ding{51}} & 
% 82.19       &    
% 74.86   & 
78.16      & 
77.54      & 
% 78.98      & 
% 78.34      &   
% 72.99           & 
% 71.28  & 
76.61          & 
67.77   
% &  
% 59.36    
% & 9.36                          &  15.28                        & 33.31                         & 58.98 
\\
PandaGPT \cite{pandagpt}                   & \textcolor{ForestGreen}{\ding{51}} & 
% 83.38      & 
% 76.81      &  
78.92     & 
78.02      &  
% 79.11   & 
% 79.24  &  
% 78.48       &  
% 73.12     &  
79.06         & 
70.58  
% &  
% 65.85  
% &    10.35                      & 16.92                         &  34.88                        & 61.22
\\
VideoLlama \cite{videollama}                 & \textcolor{ForestGreen}{\ding{51}} &   
% 84.48    & 
% 77.06      &  
79.90     & 
77.26      &   
% 81.36    & 
% 80.01      &  
% 81.21           & 
% 76.10  &  
82.90         &  
72.32   
% & 
% 67.52 
% & 11.45                         & 17.39                         &   35.14                       &  63.63                
\\
X-InstructBLIP \cite{xinstructblip}              & \textcolor{ForestGreen}{\ding{51}} &   
% 85.53    & 
% 80.09     &  
81.14     &  
82.29     & 
% 83.91      &  
% 82.59     & 
% 80.28             & 
% 77.45 &  
83.89         & 
73.43   
% &  
% 68.83  
% & 12.31                         &  18.82                        & 37.93                         & 65.73
\\
\midrule
\rowcolor{ThemeColor}
% $\text{\modelname}_{\text{final}}$ 
\textbf{\modelname (ours)} &
  \textcolor{ForestGreen}{\ding{51}} 
  % & \textbf{88.24}
   % & \textbf{86.65}
   & \textbf{84.60}
   & \textbf{87.05}
   % & \textbf{86.55}
   % & \textbf{86.61}
   % & \textbf{83.62}
   % & \textbf{80.51}
   & \textbf{85.70}
   & \textbf{75.98}
   % & \textbf{73.33} 
   % \textbf{16.88} & \textbf{23.18} & \textbf{45.67} & \textbf{76.84}
   \\ 
   \midrule
   $\textcolor{blue}{\Delta_{\text{\modelname} - \text{X-InstructBLIP}}}$ &
  \textcolor{ForestGreen}{\ding{51}} &
  % \colorbox{increase}{+3.17\%} &
  % \colorbox{increase}{+8.19\%} &
  \colorbox{increase}{+4.26\%} &
  \colorbox{increase}{+5.78\%} &
  % \colorbox{increase}{+3.15\%} &
  % \colorbox{increase}{+4.87\%}  &
  % \colorbox{increase}{+4.16\%} &
  % \colorbox{increase}{+3.95\%} &
  \colorbox{increase}{+2.16\%} &
  \colorbox{increase}{+3.47\%} 
  % &
  % \colorbox{increase}{+6.54\%} 
  % \colorbox{increase}{+37.12\%} &
  % \colorbox{increase}{+23.17\%} &
  % \colorbox{increase}{+20.41\%} &
  % \colorbox{increase}{+16.9\%}
  \\
  \bottomrule
\end{tabular}%
}
% \end{adjustbox}
\vspace{0.05in}
\caption{\textbf{Quantitative results on AVQA task}. The reported numbers on AVQA dataset \cite{avqadataset} are on the val split. For the MUSIC-AVQA dataset \cite{musicavqadataset}, results are reported on the balanced test set. Here, Count: Counting, Comp: Comparative.}
\label{tab:avqa_count_comp}
\vspace{-7mm}
\end{table}

\section{Evaluation Metrics}
\label{evaluation metrics}

For the visual grounding task, we evaluate our model against other baselines 
% comparative baseline models across four benchmark datasets: VGG-SS \cite{vggss}, SoundNet-Ficker \cite{flickrsoundnetarda}, PascalSound \cite{pascal} and AVSBench \cite{avsbench}. We employ 
two key metrics to assess visual grounding effectiveness: Intersection over Union (IoU) and Area Under Curve (AUC). These metrics provide a comprehensive measure of our model's ability to accurately localize visual elements in correlation with auditory cues.
% In addressing the temporal audio event localization task, we assess \modelname alongside baseline models using the LLP \cite{llpdataset} and AudioSet Strong \cite{audioset} benchmarks. We compute the segment-level F-score to measure the accuracy.
For the image-guided audio temporal localization task, we report the segment-level F-score.
For the Audio-Visual Fact-checking (AVFact) task, 
% we collect a dataset containing  to evaluate the measure the accuracy of low-level audio-visual semantic understanding (see more data details in the Appendix). 
we split the evaluation tasks into four different categories, each with its unique dimension of assessment. We report the Precision and Recall scores for each category.
% Regarding the audiovisual captioning task, our evaluation includes the VALOR-32K \cite{valor} benchmarks. 
We report the performance of audio-visual captioning task on several established metrics, including BLEU@4 \cite{bleu}, METEOR \cite{meteor}, ROUGE \cite{rouge} and CIDEr \cite{cider}. 
Lastly, for the audiovisual visual question answering, we follow \cite{liu2024tackling, nadeem2024cad} and report
% model evaluations are conducted on the AVQA \cite{avqadataset} and MUSIC-AVQA \cite{musicavqadataset} benchmarks. For MUSIC-AVQA, we test the models on the balanced test set. each benchmark includes 
5 different types of audio-visual relationships, including \textit{Existential}, \textit{Location}, \textit{Counting}, \textit{Temporal}, and \textit{Comparative}.

% \textcolor{red}{Report Comp, Count metrics}

\section{Failure Cases}
\label{failure_cases}

Although \modelname demonstrates impressive reasoning and grounding capabilities under various audio-visual settings, there are still some cases where the model fails to comprehend complex and obscured references, especially in cluttered environments or audios with multiple overlapping sounds. Fig. \ref{fig:failure_cases} demonstrates a few cases where our method produces suboptimal or sometimes incorrect inference results. In Fig. \ref{fig:failure_cases_spatial} due to the lack of visibility of the object of interest (Chainsaw), our model couldn't correctly identify the spatial region pertaining to it. Similarly, as the facial region of the speaker is not evident, \modelname fails to correctly locate the active speaker. In Fig. \ref{fig:failure_cases_temporal} due to the overlapping audio of multiple instruments and the presence of ambient sound, our method could partially capture the duration through which the guitar makes sound (refer to supplementary video). The same happens with the other temporal audio localization example where the audio starts with a loud baby laughter sound which gradually fades with the adult person's voice taking over. Our model could only identify the initial part of the baby's sound. For the AV Fact task (in Fig. \ref{fig:failure_cases_avfact}), in the first example, due to occluded facial region, our model produces the wrong output, whereas in the second example, due to the indistinguishable, cluttered and blurry background, \modelname fails to correctly identify the flying bird.

\begin{figure}
  \centering
  \begin{subfigure}{0.98\textwidth}
    \centering
    \includegraphics[width=\columnwidth]{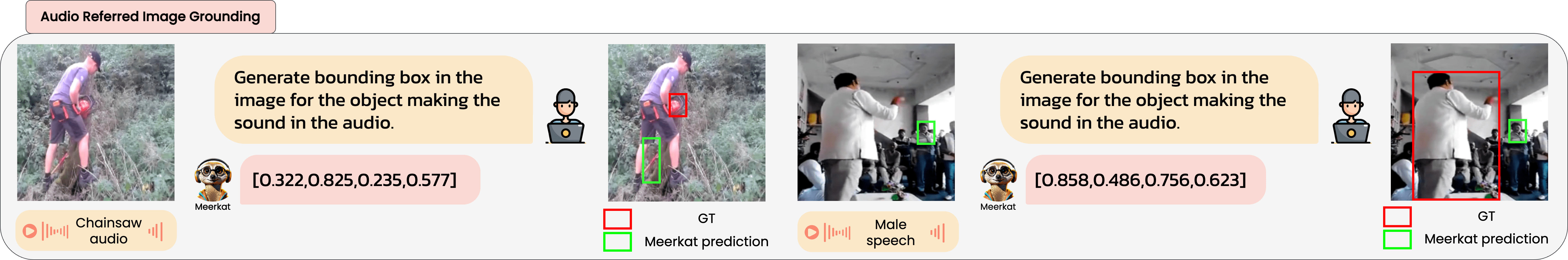}
    \caption{Failure cases on image grounding task.}
    \label{fig:failure_cases_spatial}
  \end{subfigure}
  \\
  \begin{subfigure}{0.98\textwidth}
    \centering
    \includegraphics[width=\columnwidth]{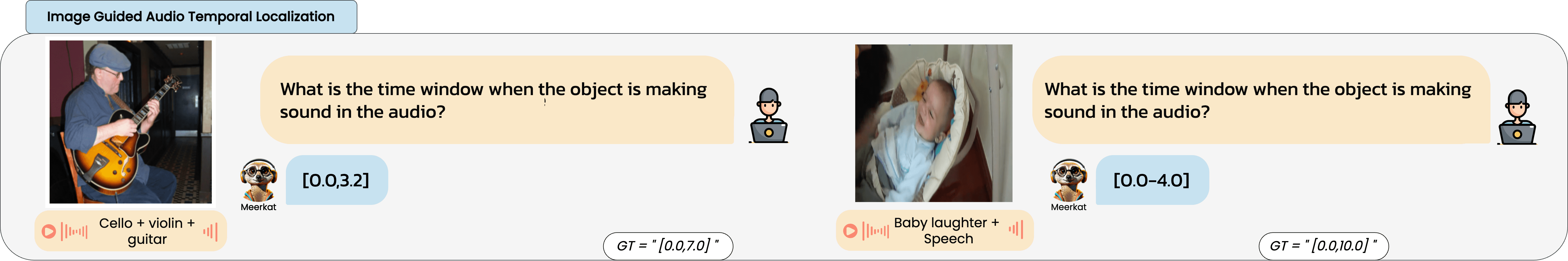}
    \caption{Failure cases on audio temporal localization.}
    \label{fig:failure_cases_temporal}
  \end{subfigure}
  \\
  \begin{subfigure}{0.98\textwidth}
    \centering
    \includegraphics[width=\columnwidth]{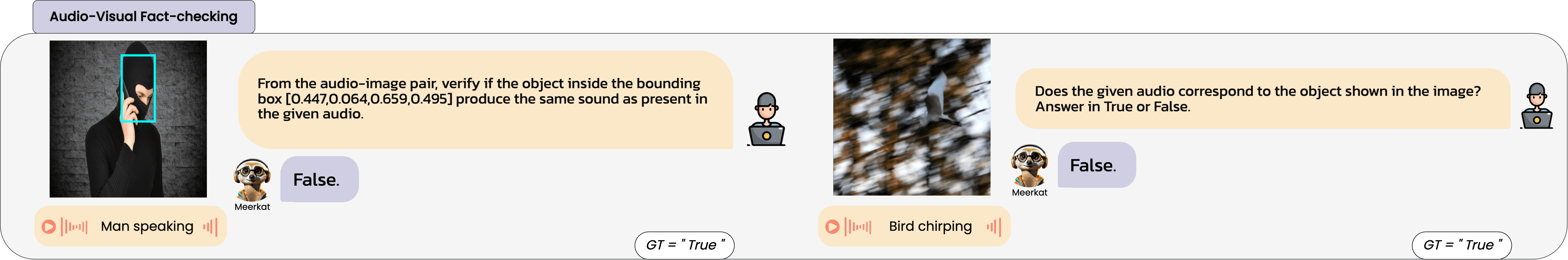}
    \caption{Failure cases on AVFact task.}
    \label{fig:failure_cases_avfact}
  \end{subfigure}
  \caption{Failure cases of \modelname on different fine-grained tasks.}
  \label{fig:failure_cases}
\end{figure}

\section{Ethics Statement}
\label{ethics statement}

In this paper, we propose a novel framework for multi-modal LLM by combining the audio and visual modalities. For all the tasks we leverage publicly available datasets and do not engage in collecting any private data. However, we acknowledge that the public datasets may have implicit bias \cite{fabbrizzi2022survey}. While LLMs being pre-trained on web-scale data inherently contain extensive knowledge about the real world, we recognize its potential learning bias as well. Moreover, these methods are prone to mistakes and might generate wrong or misleading results. The existing tools to measure various aspects of the LLM-generated outputs (e.g., toxicity \cite{helm}) are predominantly restricted to the language modality and not applicable across other modalities. 

It's important for the users to recognize these limitations and proceed with caution, especially in scenarios where the precision and neutrality of results hold significant importance. Users are encouraged to thoroughly scrutinize and validate the outputs of the model to avoid the possibility of disseminating inaccurate information. We will publicly release the codebase and curated datasets to ensure reproducibility and encourage future research. Finally, during our data preparation stage, we don't collect or use any personal/human subject data.

\end{document}